\pdfoutput=1
\documentclass{article}
\pdfsuppressptexinfo=-1 \pdfinfoomitdate=1 \pdftrailerid{}
\PassOptionsToPackage{nonamebreak}{natbib}
\usepackage{iclr2027_conference,times}
\usepackage[T1]{fontenc}
\usepackage[utf8]{inputenc}
\usepackage{amsmath,amssymb,booktabs,multirow,graphicx,array,xcolor}
\usepackage{colortbl}
\usepackage{subcaption}
\usepackage{capt-of}
\usepackage{placeins}
\usepackage{hyperref}
\usepackage{url}
\makeatletter
\g@addto@macro\UrlBreaks{\do-}
\makeatother
\hypersetup{colorlinks=false,pdfborder={0 0 1},citebordercolor={0 1 0},linkbordercolor={1 0 0},urlbordercolor={0 1 1},
  pdftitle={Accuracy Is Not Evidence: Measuring and Training Evidence Dependence in Video Temporal Grounding},pdfauthor={Zijie Jin, Shuai Zhao, Haoran Tao, Chaomin Huang, Liwen Luo}}
\usepackage{fontawesome5}
\iclrfinalcopy
\definecolor{brand}{RGB}{0,0,0}
\newcommand{\urlhome}{https://jinzijie-bupt.github.io/Accuracy-Is-Not-Evidence/}
\newcommand{\urlcode}{https://github.com/jinzijie-bupt/Accuracy-Is-Not-Evidence}
\newcommand{\urlmodel}{https://huggingface.co/jinzijie-bupt}
\newcommand{\urldata}{https://huggingface.co/datasets/jinzijie-bupt/Accuracy-Is-Not-Evidence}

\renewcommand{\floatpagefraction}{0.8}

\newcommand{\papertablestyle}{\footnotesize\setlength{\tabcolsep}{3pt}\renewcommand{\arraystretch}{1.06}}
\newcommand{\NA}{---}
\definecolor{pcrow}{RGB}{233,245,237}                         %
\newcommand{\ci}[2]{\textcolor{black!55}{[#1,\,#2]}}          %

\DeclareMathOperator{\softplus}{softplus}
\DeclareMathOperator{\relu}{relu}
\newcommand{\Gm}{\Gamma\,\text{(mask)}}
\newcommand{\Gs}{\Gamma\,\text{(swap)}}
\newcommand{\dE}{d_{E}}
\newcommand{\dO}{d_{O}}
\newcommand{\Rfive}{R@0.5}
\newcommand{\grouped}{placebo-controlled objective}
\newcommand{\audit}{fixed-answer audit}

\title{Accuracy Is Not Evidence:\\ Measuring and Training Evidence Dependence\\ in Video Temporal Grounding}
\author{Zijie Jin$^{1}$\quad Shuai Zhao$^{1,\dagger}$\quad Haoran Tao$^{2}$\quad Chaomin Huang$^{3}$\quad Liwen Luo$^{4}$\\[2pt]
\small $^{1}$Beijing University of Posts and Telecommunications\\
\small $^{2}$Institute of Automation, Chinese Academy of Sciences\quad $^{3}$Peking University\quad $^{4}$Independent Researcher\\[2pt]
\small\texttt{\{\href{mailto:jinzijie@bupt.edu.cn}{jinzijie},\,\href{mailto:zhaoshuaiby@bupt.edu.cn}{zhaoshuaiby}\}@bupt.edu.cn}}

\begin{document}
\maketitle
\lhead{Preprint}
\vspace{-18pt}
{\centering
\hypersetup{pdfborder={0 0 0}}%
\href{\urlhome}{\textcolor{brand}{\faGlobe\ Homepage}}\hspace{2em}%
\href{\urlcode}{\textcolor{brand}{\faGithub\ Code}}\hspace{2em}%
\href{\urlmodel}{\textcolor{brand}{\faCube\ Model}}\hspace{2em}%
\href{\urldata}{\textcolor{brand}{\faDatabase\ Data}}\par}
\vspace{2pt}
{\renewcommand{\thefootnote}{}\footnotetext{$^{\dagger}$Corresponding author.}}

\begin{abstract}
In video temporal grounding, a correct interval can come from finding the event, answering blind or fitting one annotation style, and Recall at IoU scores them alike. We measure evidence dependence, how much the answer rests on the event's frames. Our \audit{} edits either the frames inside the annotated interval or, as a placebo, as many frames elsewhere in the same video, and compares the two drops in the annotated answer's likelihood. Supervised fine-tuning and three public checkpoints trained with IoU-reward reinforcement learning raise the accuracy of Qwen2.5-VL-7B by 9 to 20 points, partly by answering blind and by fitting one annotation. Yet their answers rest on the event no more than the base model's, and the same holds for IoU-reward training on our data at the audit's frame rate. This comparison is a difference of the model's own likelihoods, so we can train on it. A placebo contrast and a bound on the placebo drop rule out learning to react to any edit. With them, evidence dependence rises in all five base models we train, and the weight on clean supervision controls how much. Added to IoU-reward training, the objective raises evidence dependence while keeping the accuracy gain. Masking the event's frames makes the trained models decline or miss the event more often than their base; masking as many other frames does not. The dependence moves to another event when the query names it, and holds on a second dataset and under patch scrambling, an edit never used in training.
\end{abstract}

\section{Introduction}
\label{sec:intro}

Video temporal grounding asks a model when a described event happens. Given a video and the query ``another person is fixing a doorknob'', the model returns an interval. The benchmark counts it correct when its IoU with the annotated interval is at least 0.5 (Recall at IoU 0.5, \Rfive{}; \citealp{charadessta2017,lei2021qvhighlights}), and we give changes in it in percentage points. Post-training raises this accuracy by many points, but a model can earn a correct interval in three ways. It can find the event. It can answer blind: with every frame replaced by black, Qwen2.5-VL-7B still returns an interval for 12\% of Charades-STA test questions; fine-tuned on 2,400 annotated timestamps, it does so for 78\%. And it can fit one annotation style: a fixed window over the first ten seconds reaches two thirds of the base model's accuracy, because benchmarks carry shortcuts \citep{geirhos2020shortcut} in the length and position of their intervals \citep{otani2020hidden,yuan2021closer}.

\begin{figure}[t]
\centering
\begin{minipage}[c]{0.42\linewidth}
\centering
\includegraphics[width=\linewidth]{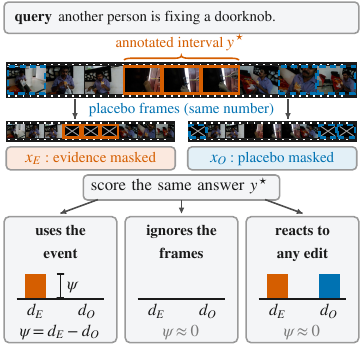}
\end{minipage}\hfill
\begin{minipage}[c]{0.56\linewidth}
\centering
\includegraphics[width=\linewidth]{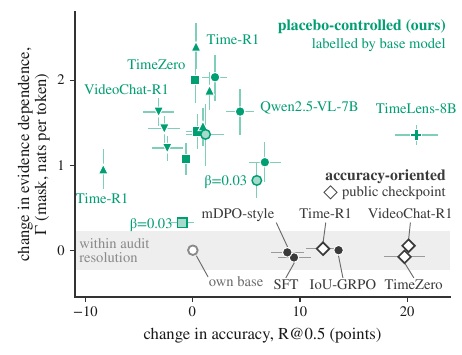}
\end{minipage}
\caption{\textbf{Accuracy and evidence dependence are separate axes.} Left: the audit scores the same correct answer, the annotated interval $y^\star$, with the event's frames masked ($x_E$) and with as many frames masked elsewhere, the placebo ($x_O$). The schematic bars show the two drops in the answer's log-likelihood, $\dE$ and $\dO$, and their difference $\psi$ for three kinds of model. Right: each point places one post-trained model by its change from its own base (the origin) in \Rfive{} (horizontal) and in evidence dependence $\Gamma$, the change in $\psi$ (vertical), with \Rfive{} on the full test split and $\Gamma$ on the first 200 test questions; lines are 95\% intervals. The grey band is the audit's resolution, the farthest from zero that the 95\% interval of $\Gamma$ reaches between two seeds of one recipe (Section~\ref{sec:audit}). Accuracy-oriented training (dark grey) moves models right but not out of the band; SFT, IoU-GRPO and mDPO-style are trained from Qwen2.5-VL-7B, and the public checkpoints (open diamonds) are compared with it. The \grouped{} (green) moves them up. Marker shape gives the base model, its several points differ in $\alpha$ or $\beta$, and light-filled points use the smaller $\beta{=}0.03$ (Section~\ref{sec:objective}).}
\label{fig:teaser}
\end{figure}

The quantity we measure is \emph{evidence dependence}: how much a model's confidence in its answer rests on the frames that contain the event. Finding the event means its frames are enough to place the answer; evidence dependence means they are needed. A model can have the first without the second. Current audits alter the video and regenerate the answer \citep{otani2020hidden,tvbench2025,leemvbench2026,tempcore2026,carve2026}. When the regenerated interval changes, the score cannot tell lost evidence from a different guess. A score of how much worse the model does without the evidence also rewards one that does worse whenever anything is removed, a known confound in attribution \citep{hooker2019roar}.

Our \audit{} (Section~\ref{sec:audit}; Figure~\ref{fig:teaser}, left) scores the annotated interval by teacher forcing, so every readout is the likelihood of the same answer. For the doorknob query, we black out the doorknob-fixing frames or, as a placebo, as many frames elsewhere in the same video. The difference between the two drops in likelihood is zero for a model that reacts equally to every edit, however much each edit hurts it. Our measure, $\Gamma$, is how much training changes this difference from the base model, on the questions both models answer correctly. We report it next to the trained model's own placebo drop. Pairs of independently trained seeds of the same recipe, audited against each other, show how far from zero $\Gamma$ can reach by chance alone. This sets the audit's \emph{resolution}, and only a larger $\Gamma$ counts as an effect.

The post-training for accuracy that we audit takes all three routes and does not raise evidence dependence (Section~\ref{sec:findings}; Figure~\ref{fig:teaser}, right). We audit Qwen2.5-VL-7B after supervised fine-tuning (SFT) and IoU-reward GRPO (IoU-GRPO), both trained by us, and three public checkpoints trained from it by reinforcement learning (TimeZero, Time-R1 and VideoChat-R1). All are more accurate than the base. The public checkpoints' predictions follow the event more often when its frames are moved, so part of their gain comes from finding it. SFT, TimeZero and Time-R1 answer most all-black videos, and against a second annotation the public checkpoints lose their lead over the base. Yet none raises $\Gamma$ beyond the audit's resolution. For SFT, IoU-GRPO, TimeZero and Time-R1, even the answers they newly get right rest on the event no more than the base's own correct answers.

Because the audit's difference comes from the model's own log-likelihoods, we can train on it (Section~\ref{sec:objective}), and the placebo is needed there too. Rewarding the evidence drop alone teaches the model to react to any edit: its own placebo drop rises far above the base's. A contrast against the matched placebo and a bound on the placebo drop remove this solution. We call the result the \emph{\grouped{}}. With it, $\Gamma$ rises in all five base models we train. The weight $\alpha$ on clean supervision controls the amount: in Qwen2.5-VL-7B, raising it from 0.25 to 1 halves $\Gamma$ and triples the accuracy gain. Used alone, the objective gains fewer points than SFT. Added to IoU-GRPO, it keeps IoU-GRPO's accuracy gain and still raises $\Gamma$.

We then test the trained models on quantities outside the loss (Section~\ref{sec:readouts}). When the event's frames are masked, the placebo-controlled Qwen2.5-VL-7B model declines or misses the event more often than its base, and with decoding forced to give an interval it still loses 15 points of \Rfive{} more to this mask than to the placebo mask, relative to its base. Unlike SFT, the placebo-controlled Qwen2.5-VL-7B and Time-R1 models decline queries about another video at least as selectively as their bases. In most trained models the dependence also moves with the query, and it holds on QVHighlights and under patch scrambling, an edit never seen in training.

\paragraph{Contributions.}
(i) We introduce the \audit{}, which measures evidence dependence against a matched placebo in the same video (Section~\ref{sec:audit}). (ii) We show that the accuracy-oriented post-training we audit takes all three routes without raising evidence dependence (Section~\ref{sec:findings}). (iii) We introduce the \grouped{}, which raises evidence dependence in all five base models we train, with the weight on clean supervision setting how much (Section~\ref{sec:objective}). (iv) We show that the trained models decline or miss the event more often than their base when its frames are masked, and that their dependence moves with the query (Section~\ref{sec:readouts}).

\section{Measuring evidence dependence}
\label{sec:audit}

A grounding benchmark checks whether a prediction agrees with the annotation. The \audit{} asks instead how much of the model's confidence in a correct answer disappears when the event's frames are edited, compared with editing as many other frames. For the query ``another person is fixing a doorknob'', we score the annotated interval on the original video, with the doorknob-fixing frames blacked out, and with as many frames blacked out elsewhere in the same video. A model that uses the event loses confidence under the first edit only; one that reacts to any edit loses it under both (Figure~\ref{fig:teaser}, left). Throughout, the \emph{evidence} of a question means the frames inside its annotated interval, and evidence dependence means dependence on these frames. The audit does not measure other visual information a model may use, and Appendices~\ref{app:add-prior} to~\ref{app:add-cd} read $\Gamma$ against the annotation prior, the frames next to the event and the training split.

\paragraph{The evidence drop, the placebo drop and $\Gamma$.}
For a question with video $x$, query $q$ and annotated interval $y^\star$, let $\ell(x)=\log p(y^\star\mid x,q)$ be the model's log-likelihood of the annotated answer, in its template's format and averaged over the answer tokens. Editing the frames inside $y^\star$, the \emph{evidence frames}, gives $x_E$. Editing the same number of frames drawn at random outside $y^\star$, with the same operator, gives the \emph{placebo} $x_O$. The evidence drop, the placebo drop and their difference are
\begin{equation}
\dE=\ell(x)-\ell(x_E),\qquad \dO=\ell(x)-\ell(x_O),\qquad \psi=\dE-\dO,
\end{equation}
in nats per token, with the placebo drop averaged over three draws. The difference $\psi$ says how much more the evidence edit hurts than the placebo edit. We measure the effect of training as the change in $\psi$ from the base model,
\begin{equation}
\Gamma \;=\; \frac{1}{|H_\cap|}\sum_{i\in H_\cap}\big(\psi^{\mathrm{trained}}_i-\psi^{\mathrm{base}}_i\big),
\end{equation}
where $H_\cap$ holds the questions that both models localise correctly (IoU $\ge0.5$ on the clean video). $\Gamma$ thus compares two ways of being right on the same questions. In the doorknob question the event spans 13.2 to 23.7\,s, so $x_E$ blacks out the frames sampled in that span and $x_O$ as many frames sampled elsewhere in the video. Its $\psi$ is positive when losing the doorknob-fixing frames hurts the annotated answer more than losing other frames, and the question enters $\Gamma$ when both the trained model and its base localise it. The table of notation on page~\pageref{tab:notation-main} collects the symbols.

\paragraph{How large a $\Gamma$ counts.}
Before any training, the evidence edit already costs a little more than the placebo edit: on the questions it answers correctly, the Qwen2.5-VL-7B base has $\psi$ between $+0.18$ and $+0.23$ nats per token. A $\Gamma$ of zero means that training left this unchanged. Two seeds of one recipe differ only by chance, so we audit two pairs of seeds of the placebo-controlled Qwen model (Section~\ref{sec:objective}) against each other. Their 95\% intervals of $\Gamma$ reach at most $0.23$ nats per token from zero under mask and $0.31$ under swap. We call each bound the audit's \emph{resolution} and read a $\Gamma$ as an effect only when its estimate lies beyond it. For scale, the placebo-controlled models of Section~\ref{sec:objective} reach a $\Gamma$ of about $+1$ to $+2$. The likelihood audit uses the first 200 test questions with three placebo draws each, and Appendix~\ref{app:added} repeats it on 600 questions from other videos and under occlusion, new seeds and other prompts. Accuracy (\Rfive{}) is measured on the full 3{,}720-question Charades-STA test split, and intervals are bootstrapped over videos (Appendix~\ref{app:estimands}).

\begin{table}[t]
\centering\scriptsize\setlength{\tabcolsep}{4pt}\renewcommand{\arraystretch}{1.08}
\caption*{\textbf{Notation used in the main text.} Likelihood quantities are token means in nats, and accuracy changes are in percentage points of \Rfive{}. Table~\ref{tab:notation} gives the full notation of the appendix.}
\label{tab:notation-main}
\begin{tabular}{@{}lp{10.6cm}@{}}
\toprule
Symbol & Meaning \\
\midrule
$y^\star$, $x$, $q$ & annotated interval, video, query; the \emph{evidence frames} are the frames inside $y^\star$ \\
$x_E$, $x_O$ & video with the evidence frames edited; with as many frames outside $y^\star$ edited (placebo) \\
$\dE$, $\dO$, $\psi$ & drops of the answer's log-likelihood under $x_E$ and under $x_O$ (nats per token); $\psi=\dE-\dO$ \\
$H_\cap$ & questions that both the trained model and its base localise (IoU $\ge0.5$ on the clean video) \\
$\Gamma$, $\Gs$ & mean trained-minus-base change in $\psi$ on $H_\cap$, under mask and under swap \\
resolution & farthest from zero that the 95\% interval of $\Gamma$ reaches when two seeds of one recipe are audited against each other ($0.23$ under mask, $0.31$ under swap) \\
\Rfive{}, points & share of questions whose predicted interval has IoU $\ge0.5$ with $y^\star$; changes in percentage points \\
evidence share, follows & readouts on generated answers (end of this section) \\
$\alpha$, $\beta$ & weight on clean supervision and scale of the margins in the objective (Section~\ref{sec:objective}) \\
16 frames, 2 fps & 16 frames sampled uniformly from each video; 2 frames sampled per second \\
\bottomrule
\end{tabular}
\end{table}

\paragraph{What is edited and what is scored.}
We use two edit operators. \emph{Mask} sets frames to black; \emph{swap} replaces them with frames from another video. Mask is the primary operator: $\Gamma$ without a label refers to it, and under swap we write $\Gs$. The audit teacher-forces each model's answer string directly after the question, without the reasoning trace that TimeZero and Time-R1 write, so the scored answer is the same under every edit; generated predictions include the trace. Each model and its base share one prompt template and sample frames at 2 per second.

\paragraph{Three signatures of evidence dependence.}
Each signature rules out another cause of a drop under the evidence edit. First, the trained model's own placebo drop must stay near its base's. A model whose placebo drop rises reacts to any edit, and training on the evidence drop alone produces one (Section~\ref{sec:objective}); we therefore report every trained model's $\Gamma$ with its own placebo drop. Second, the dependence should move with the query: when the query names another event in the same video, it should move to that event's frames, which Section~\ref{sec:readouts} tests. Third, the rise in $\psi$ should not come from a sharper likelihood on the clean video. SFT sharpens it as much as the placebo-controlled models do, yet its $\Gamma$ stays at or below zero, so sharpening alone does not raise $\Gamma$.

\paragraph{Two readouts on generated answers.}
The likelihood audit scores a fixed answer, so we also check what the models generate. We regenerate each model's predictions on the first 1{,}000 test questions with the evidence frames masked and with as many placebo frames masked (accuracies $a_E$ and $a_O$). The \emph{evidence share} $1-a_E/a_O$, the counterpart of $\psi$ for generated answers, is the share of the placebo-masked accuracy lost to the evidence mask. For \emph{follows}, we exchange the event's frames with an equally long block elsewhere in the video and regenerate on $H_\cap$; a prediction follows if its IoU with the annotated interval moved to that block is at least $0.5$. Follows tests whether the event's frames are enough to place the answer; the evidence share tests whether they are needed.

\section{What post-training accuracy is made of}
\label{sec:findings}

Every accuracy-oriented model in Table~\ref{tab:audit} is 9 to 20 points more accurate than its base, Qwen2.5-VL-7B-Instruct \citep{qwen25vl2025}, and none raises $\Gamma$ beyond the audit's resolution. Three of them are public checkpoints trained for this task with IoU-reward reinforcement learning: TimeZero and Time-R1, two releases of one method \citep{wang2025timer1}, and VideoChat-R1 \citep{videochatr1_2025}. We train SFT and an mDPO-style preference for clean over corrupted video \citep{wang2024mdpo} on the same 2,400 Charades-STA pairs as our objective (Section~\ref{sec:objective}), and IoU-GRPO \citep{deepseekmath2024} on the full Charades-STA training split at 16 frames per video. Trained instead at the audit's 2 frames per second on the same 2,400 pairs as our objective, IoU-GRPO gains about 16 points on the full test split, and on 600 further test questions its $\Gamma$ stays within about $0.07$ of zero at both seeds and at a checkpoint matched to our objective's training time. Adding our objective to its loss keeps the gain and raises $\Gamma$ to about $+0.9$ (Appendix~\ref{app:add-matched}). Appendix~\ref{app:extended} gives the per-model tables.

\providecommand{\est}[3]{\begin{tabular}[t]{@{}c@{}}#1\\[-1.5pt]{\fontsize{6}{7}\selectfont\color{black!55}[#2,\,#3]}\end{tabular}}
\begin{table}[tb]
\centering\footnotesize\setlength{\tabcolsep}{4pt}\renewcommand{\arraystretch}{1.12}
\caption{\textbf{Training for accuracy adds correct answers but not evidence dependence; the placebo-controlled objective adds both.} All models start from Qwen2.5-VL-7B-Instruct. The base row gives levels. Every other entry is a change from the base on the same questions under the same template, except each model's own placebo drop $\dO$. Accuracy is measured on all 3{,}720 Charades-STA test questions, the likelihood audit ($\Gamma$ and $\dO$, in nats per token) on the first 200 test questions, the evidence share on the first 1{,}000, and follows on $H_\cap$ of the first 200. Accuracy changes are in points, and brackets give 95\% intervals. Bold marks a change in points whose interval excludes zero or a $\Gamma$ beyond the audit's resolution, and grey a $\Gamma$ within it.}
\label{tab:audit}
\begin{tabular}{@{}lcccccc@{}}
\toprule
 & Accuracy & \multicolumn{3}{c}{\shortstack{Evidence dependence\\(nats per token)}} & \multicolumn{2}{c@{}}{\shortstack{Regenerated answers\\(points)}} \\
\cmidrule(lr){2-2}\cmidrule(lr){3-5}\cmidrule(l){6-7}
Model & \Rfive{} & \shortstack{$\Gamma$\\mask} & \shortstack{$\Gamma$\\swap} & $\dO$ & \shortstack{evidence\\share} & \shortstack{follows\\moved event} \\
\midrule
Qwen2.5-VL-7B (base) & 0.480 & \NA & \NA & $-$0.06 & 69\% & \NA \\
\addlinespace[2pt]\multicolumn{7}{@{}l}{\emph{Trained for accuracy}} \\
\quad SFT & \est{\textbf{+9.4}}{7.9}{11.0} & \est{\textcolor{black!45}{$-$0.09}}{$-$0.15}{$-$0.02} & \textcolor{black!45}{$-$0.03} & 0.00 & \est{\textbf{\boldmath $-$14.7}}{$-$19.8}{$-$9.4} & \est{+7.0}{$-$2.0}{16.5} \\
\quad IoU-GRPO, 16 frames & \est{\textbf{+13.6}}{11.9}{15.4} & \est{\textcolor{black!45}{0.00}}{$-$0.03}{0.04} & \textcolor{black!45}{+0.02} & $-$0.05 & \est{\textbf{\boldmath $-$6.1}}{$-$11.0}{$-$1.2} & \est{+8.7}{$-$1.1}{16.8} \\
\quad mDPO-style & \est{\textbf{+8.8}}{7.3}{10.4} & \est{\textcolor{black!45}{$-$0.03}}{$-$0.09}{0.04} & \textcolor{black!45}{+0.01} & 0.00 & \est{+1.1}{$-$3.3}{5.7} & \est{+5.1}{$-$4.5}{14.1} \\
\quad TimeZero (public) & \est{\textbf{+19.8}}{17.9}{21.7} & \est{\textcolor{black!45}{$-$0.08}}{$-$0.16}{0.01} & \textcolor{black!45}{$-$0.05} & $-$0.03 & \est{+1.3}{$-$4.2}{6.9} & \est{\textbf{+16.2}}{6.4}{26.9} \\
\quad Time-R1 (public) & \est{\textbf{+12.2}}{10.6}{13.8} & \est{\textcolor{black!45}{+0.02}}{$-$0.02}{0.06} & \textcolor{black!45}{+0.05} & $-$0.07 & \est{$-$4.4}{$-$9.5}{0.8} & \est{\textbf{+15.3}}{4.6}{26.7} \\
\quad VideoChat-R1 (public) & \est{\textbf{+20.1}}{18.3}{22.1} & \est{\textcolor{black!45}{+0.05}}{0.02}{0.09} & \textcolor{black!45}{+0.03} & 0.00 & \est{+8.2}{$-$0.2}{17.7} & \est{\textbf{+20.8}}{10.8}{31.5} \\
\addlinespace[2pt]\multicolumn{7}{@{}l}{\emph{Placebo-controlled objective (ours)}} \\
\rowcolor{pcrow}\quad $\alpha{=}0.5$ & \est{\textbf{+4.4}}{3.1}{5.7} & \est{\textbf{+1.63}}{1.36}{1.89} & \textbf{+1.41} & 0.01 & \est{\textbf{+15.7}}{11.4}{20.3} & \est{+6.1}{$-$4.3}{16.1} \\
\bottomrule
\end{tabular}
\end{table}

\paragraph{The public checkpoints end up where our SFT model does.}
Other groups trained them on their own data and rewards, yet in $\Gamma$ they sit next to SFT (Figure~\ref{fig:teaser}, right). Our objective, trained on the same base and pairs as SFT, raises $\Gamma$ instead. On a second backbone, the accuracy-trained TimeLens-8B has $\Gamma=-0.16$ against Qwen3-VL-8B, within the resolution: its evidence drop falls, while its placebo drop stays where it was.

\paragraph{Finding the event.}
Part of the added accuracy uses the video. When the event's frames are moved, the public checkpoints' predictions follow them 15 to 21 points more often than the base's; for our SFT, IoU-GRPO and mDPO-style models the interval of the change includes zero (Table~\ref{tab:audit}). The public checkpoints' answers thus follow the event's frames more often without needing them more.

\begin{figure}[tb]
\centering
\includegraphics[width=\linewidth]{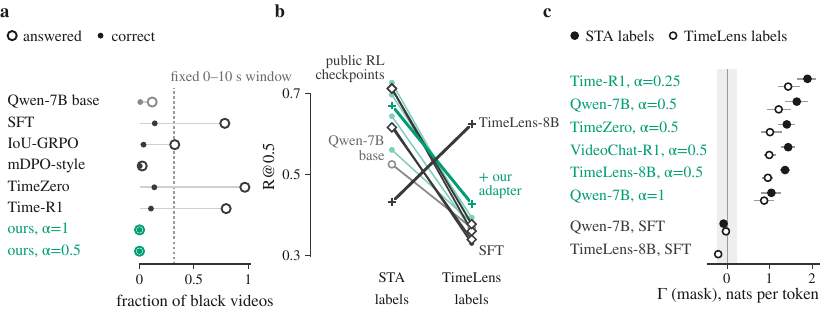}
\caption{\textbf{Part of the added accuracy comes from answering blind or fitting one annotation; $\Gamma$ keeps its sign under both label sets.} (a) shows, for all-black videos under Time-R1's prompt template, the fraction of questions still given an interval (open circles) and answered correctly (dots); the dashed line is a fixed $(0,10)$\,s window on the real videos, and green rows are our placebo-controlled Qwen models at weight $\alpha$. (b) scores each model on the questions annotated in both Charades-STA and TimeLens against each label set; the two rankings of the nine 7B configurations barely agree (Spearman $0.15$). Open diamonds are public checkpoints; open grey circles are the Qwen2.5-VL-7B base (Qwen-7B); thin lines are our 7B models (green) and SFT (dark grey); thick lines are TimeLens-8B (dark) and our adapter of it trained on the original labels (green). (c) gives $\Gamma$ (mask, nats per token, first 200 test questions) with 95\% intervals for each label set as the fixed answer (filled: Charades-STA; open: TimeLens), with the audit's resolution as a grey band; the TimeLens-8B SFT row has only the TimeLens value.}
\label{fig:findings}
\end{figure}

\paragraph{Answering blind.}
Several accuracy-trained models also answer when there is nothing to see. With every frame black, SFT still returns an interval for most questions (Section~\ref{sec:intro}). Under Time-R1's prompt template, Time-R1 answers 80\% of black videos and TimeZero 97\%. IoU-GRPO answers 32\% of black videos, while the placebo-controlled Qwen models almost never answer one (Figure~\ref{fig:findings}a). Declining black videos is not evidence dependence: the mDPO-style model almost never answers one, and its $\Gamma$ stays within the resolution.

\paragraph{Fitting one annotation.}
Part of the public checkpoints' lead is fitted to the conventions of one label set. TimeLens re-annotated Charades-STA and reported that model rankings change \citep{zhang2026timelens}; 870 test questions appear verbatim in both label sets, so we score the same predictions against each (Figure~\ref{fig:findings}b). Under the original labels the public checkpoints lead the Qwen base, and against the TimeLens labels they fall to its level. TimeLens-8B, trained on TimeLens-100K, does the reverse: it trails its base under the original labels and leads it under the TimeLens labels. Much of this difference is one convention, the length of the interval: keeping TimeLens-8B's predicted centres but setting every length to 8\,s raises its \Rfive{} under the original labels from $0.417$ to $0.678$. Our models' $\Gamma$ keeps its sign and its separation from SFT under either label set (Figure~\ref{fig:findings}c).

\paragraph{The newly correct answers rest on the event no more than the base's.}
For SFT, IoU-GRPO, TimeZero and Time-R1, $\psi$ on a model's newly correct questions differs from the base's $\psi$ on its own correct answers by $-0.08$ to $+0.03$ nats per token, and for the placebo-controlled models by $+0.9$ to $+2.0$. The new answers do use the video: regenerated with the evidence frames masked, 24\% of IoU-GRPO's new answers stay correct, against 88\% with as many placebo frames masked.
\section{Training evidence dependence}
\label{sec:objective}

\begin{figure}[tb]
\begin{minipage}[c]{0.46\linewidth}
\centering
\includegraphics[width=\linewidth]{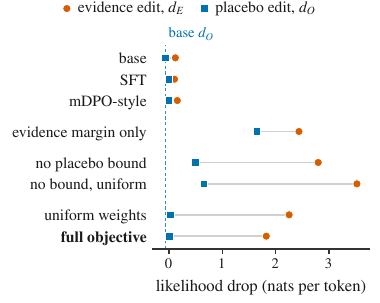}
\end{minipage}\hfill
\begin{minipage}[c]{0.51\linewidth}
\providecommand{\est}[3]{\begin{tabular}[t]{@{}c@{}}#1\\[-1.5pt]{\fontsize{6}{7}\selectfont\color{black!55}[#2,\,#3]}\end{tabular}}
\centering\footnotesize\setlength{\tabcolsep}{3pt}\renewcommand{\arraystretch}{1.12}
\captionof{table}{\textbf{With the evidence margin alone, $\Gamma$ is a third of what the full objective reaches.} Variants of the full objective (last row of Table~\ref{tab:audit}) are trained from Qwen2.5-VL-7B at $\alpha{=}0.5$; \emph{uniform} variants drop the Group-DRO weights. Samples, units, $\dO$ and bold follow Table~\ref{tab:audit}.}
\label{tab:objectives}
\begin{tabular}{@{}lcccc@{}}
\toprule
 & points & \multicolumn{3}{c@{}}{nats per token} \\
\cmidrule(lr){2-2}\cmidrule(l){3-5}
Objective & $\Delta$\Rfive{} & \shortstack{$\Gamma$\\mask} & \shortstack{$\Gamma$\\swap} & $\dO$ \\
\midrule
uniform weights & \est{\textbf{+5.4}}{4.1}{6.7} & \est{\textbf{+1.99}}{1.72}{2.24} & \textbf{+1.89} & 0.02 \\
no placebo bound & \est{\textbf{\boldmath $-$3.8}}{$-$5.4}{$-$2.1} & \est{\textbf{+2.05}}{1.75}{2.34} & \textbf{+2.34} & 0.49 \\
no bound, uniform & \est{\textbf{+3.8}}{2.4}{5.2} & \est{\textbf{+2.62}}{2.34}{2.93} & \textbf{+2.60} & 0.66 \\
evidence margin only & \est{\textbf{+6.7}}{5.2}{8.3} & \est{\textbf{+0.54}}{0.34}{0.71} & \textbf{+1.09} & 1.65 \\
\bottomrule
\end{tabular}

\end{minipage}
\caption{\textbf{The placebo contrast, and then the bound, bring the trained model's own placebo drop back to the base's.} The left panel plots, for objectives trained from Qwen2.5-VL-7B on the same pairs and audited on the same questions, the trained model's evidence drop $\dE$ and own placebo drop $\dO$ under masking (nats per token), from the base (top) to the full objective at $\alpha{=}0.5$ (bottom); the dashed line is the base's placebo drop.}
\label{fig:mechanism}
\end{figure}

The objectives of Section~\ref{sec:findings} raise accuracy but not $\Gamma$. Since $\psi$ is computed from the model's own log-likelihoods, we can write it into the loss.
\paragraph{Objective.}
In words, the objective asks that editing the event's frames lower the likelihood of the annotated answer more than editing as many other frames does, measured against the base model, while the trained model's own placebo drop stays near the base's and clean supervision keeps the answer accurate. Training scores the annotated answer by how much more likely the model being trained finds it than the frozen base model does,
$r_\theta(x)=\log p_\theta(y^\star\mid x,q)-\log p_{\mathrm{ref}}(y^\star\mid x,q)$, summed over the answer tokens. The two margins below sum over tokens, whereas the audit of Section~\ref{sec:audit} averages over them. As in preference optimisation \citep{dpo2023}, measuring against the base means that an edit which already lowers the base model's likelihood earns nothing by itself. The edits $x_E$ and $x_O$ are those of Section~\ref{sec:audit}. Training uses five edit operators: the audit's mask and swap, pixelation, temporal shift and temporal shuffle. For one training pair and one operator, the loss is
\begin{equation}
\label{eq:objective}
\begin{aligned}
\mathcal{L}_{\mathrm{op}} \;=\; &\underbrace{\softplus\!\big(-\beta\,[\,r_\theta(x)-r_\theta(x_E)\,]\big)}_{\text{evidence margin}}
\;+\; \underbrace{\softplus\!\big(-\beta\,[\,r_\theta(x_O)-r_\theta(x_E)\,]\big)}_{\text{placebo contrast}}\\
&+\; \underbrace{\relu\!\big(\dO^{\theta}-1.25\,\dO^{\mathrm{ref}}-0.10\big)}_{\text{placebo bound}}
\end{aligned}
\end{equation}
The evidence margin asks that editing the event's frames lower the likelihood of the annotated answer, so that the answer needs the event. The placebo contrast asks that this edit lower it more than editing as many other frames does, which a model that reacts equally to every edit cannot satisfy. It is the change in $\psi$ from the base, summed rather than averaged over tokens, so the loss acts on $\Gamma$ directly. The bound compares the audit's token-averaged placebo drops of the trained model, $\dO^{\theta}$, and of the base, $\dO^{\mathrm{ref}}$. It stays zero while the trained model's placebo drop is at most slightly above the base's, and it keeps the placebo drop down, which the contrast alone does not (Table~\ref{tab:objectives}). Its constants come from an earlier experiment with 16 frames of 224 pixels, in which models trained without it raised their placebo drop far above this limit (Appendix~\ref{app:prereg}), and Table~\ref{tab:objectives} tests it again at 2 frames per second. The full loss adds clean timestamp supervision,
\begin{equation}
\label{eq:total}
\mathcal{L} \;=\; \alpha\,\mathrm{NLL}_\theta(y^\star\mid x,q) \;+\; \sum_{\mathrm{op}} w_{\mathrm{op}}\,\mathcal{L}_{\mathrm{op}}
\end{equation}
where $\alpha$ sets how much clean supervision is kept and $\beta$ sets the scale of both margins ($0.1$ unless stated). The operator weights $w_{\mathrm{op}}$ are Group-DRO weights \citep{groupdro2020}, which hold the trained model's placebo drop lower than uniform weights do (Table~\ref{tab:objectives}). As $\beta\to0$ without the bound, the loss reduces to timestamp supervision alone, and this is our SFT model. The placebo-controlled models are LoRA adapters \citep{hu2022lora} trained, unless stated, for one pass over the 2,400 pairs.

\paragraph{Without the placebo terms, the model learns to react to any edit.}
Adding the terms one at a time shows that the optimiser takes the shortcut the placebo was built to exclude until both placebo terms are in the loss (Figure~\ref{fig:mechanism}, top to bottom). A preference for the clean video over one whose corrupted frames are chosen without the annotation, as in mDPO \citep{wang2024mdpo}, raises accuracy but not $\Gamma$ (Table~\ref{tab:audit}). Corrupting the annotated frames instead gives the evidence margin alone, the training form of a deletion score. The model then learns to lose likelihood wherever the video is edited. Its placebo drop rises to $1.65$ nats per token, two thirds of its evidence drop (Table~\ref{tab:objectives}). Adding the placebo contrast raises $\Gamma$, but the placebo drop still sits at $0.49$, far above the base's. The bound brings it down to $0.01$, near the base's level, while the evidence drop remains far above the base's.

\begin{figure}[tb]
\centering
\includegraphics[width=\linewidth]{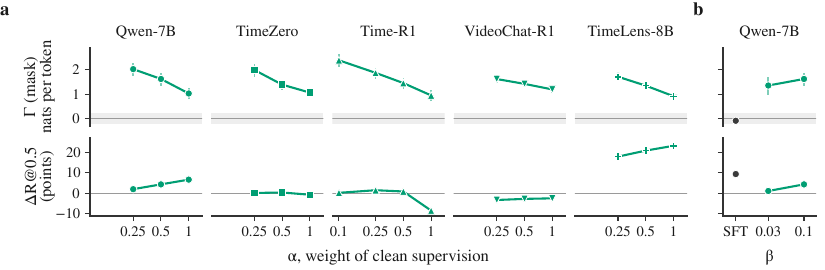}
\caption{\textbf{Evidence dependence falls with the weight on clean supervision; the accuracy change depends on the base.} The top row shows $\Gamma$ (mask, nats per token, first 200 test questions) and the bottom row the change in \Rfive{} from the model's own base (points, full test split), with 95\% intervals. (a) Each column plots one base against $\alpha$ at $\beta{=}0.1$; TimeLens-8B is built on Qwen3-VL, and the grey band is the audit's resolution. (b) plots Qwen2.5-VL-7B at $\alpha{=}0.5$ against $\beta$, with SFT at $\beta\to0$.}
\label{fig:dose}
\end{figure}

\paragraph{The weight on clean supervision controls how much is learned.}
With both placebo terms, $\Gamma$ rises in all five base models we train: Qwen2.5-VL-7B, the public TimeZero, Time-R1 and VideoChat-R1, and TimeLens-8B, built on Qwen3-VL \citep{zhang2026timelens,qwen3vl2025}. At $\beta{=}0.1$ it lies between $+0.9$ and $+2.4$, and within every base it falls as $\alpha$ grows (Figure~\ref{fig:dose}). In Qwen2.5-VL-7B, moving $\alpha$ from 0.25 to 1 lowers $\Gamma$ from $+2.03$ to $+1.03$. Over the same range, the accuracy gain rises from 2 to 7 points. The five adapters of Table~\ref{tab:families} all keep their own placebo drop within $0.10$ nats per token of their base's.

\begin{table}[tb]
\centering\scriptsize\setlength{\tabcolsep}{3pt}\renewcommand{\arraystretch}{1.05}
\caption{\textbf{The placebo-controlled objective raises evidence dependence in every base model; the change in accuracy varies with the base.} Each base model is paired with its adapter, trained at $\alpha{=}0.5$ on the same 2{,}400 Charades-STA pairs and run under the base's own template. Accuracy is measured on the full Charades-STA test split and the likelihood audit on the first 200 test questions. Changes are in points, $\Gamma$ and $\dO$ are in nats per token, and $\dO$ is given for the adapter and for its base; brackets give 95\% intervals.}
\label{tab:families}
\begin{tabular}{@{}lccccc@{}}
\toprule
Base model & base \Rfive & $\Delta$\Rfive{} & \shortstack{$\Gamma$\\mask} & \shortstack{$\Gamma$\\swap} & $\dO$ trained / base \\
\midrule
Qwen2.5-VL-7B & 0.480 & +4.4 \ci{3.1}{5.7} & +1.63 \ci{1.36}{1.89} & +1.41 & 0.01 / $-$0.06 \\
TimeZero & 0.677 & +0.5 \ci{$-$0.7}{1.5} & +1.40 \ci{1.19}{1.60} & +1.31 & 0.03 / $-$0.02 \\
Time-R1 & 0.601 & +0.9 \ci{$-$0.2}{2.0} & +1.46 \ci{1.23}{1.68} & +1.20 & 0.00 / $-$0.08 \\
VideoChat-R1 & 0.668 & $-$2.7 \ci{$-$4.2}{$-$1.2} & +1.43 \ci{1.27}{1.58} & +0.89 & $-$0.01 / 0.00 \\
TimeLens-8B & 0.417 & +20.9 \ci{18.9}{22.8} & +1.36 \ci{1.24}{1.47} & +0.88 & $-$0.03 / $-$0.03 \\
\bottomrule
\end{tabular}
\end{table}

\paragraph{The accuracy change depends on the base; more data raises $\Gamma$ further.}
At $\alpha{=}0.5$ the accuracy change runs from $-2.7$ points on VideoChat-R1 to $+20.9$ on TimeLens-8B (Table~\ref{tab:families}). Audited every 400 pairs, $\Gamma$ is still rising at the end of one pass. Doubling the training pairs to 4,800, all else fixed, takes the Qwen2.5-VL-7B model to $\Gamma=+2.78$ and a gain of 8.1 points (Appendix~\ref{app:budget}).

\paragraph{The accuracy cost depends on what the objective is combined with.}
Used alone, it gains fewer points than SFT on the same pairs and loses accuracy in domain on VideoChat-R1. On QVHighlights it adds nothing to Qwen2.5-VL-7B, where SFT gains 6.8 points, and it lowers the other placebo-controlled models by 1.5 to 4.0 points (Table~\ref{tab:s8}). Added with weight 0.01 to the loss of IoU-GRPO at 2 frames per second, it keeps IoU-GRPO's accuracy gain in domain, comes within a point of it on QVHighlights and raises $\Gamma$ to about $+0.9$, below the $\Gamma$ of the objective alone (Appendix~\ref{app:add-matched}). At the same accuracy gain, it keeps fewer of its correct answers than IoU-GRPO once the event's frames are masked (Appendix~\ref{app:add-matched-acc}). Under reworded instructions it also keeps its $\Gamma$, whereas two seeds of the objective alone then differ by more than the audit's resolution (Appendix~\ref{app:add-prompts}).

\section{What the trained models do with the evidence}
\label{sec:readouts}

Every test in this section lies outside the loss: it changes the edit operator, the query or the dataset, or it reads the generated answer and the refusal. Appendix~\ref{app:shift} gives the per-model tables.

\begin{figure}[tb]
\centering
\includegraphics[width=\linewidth]{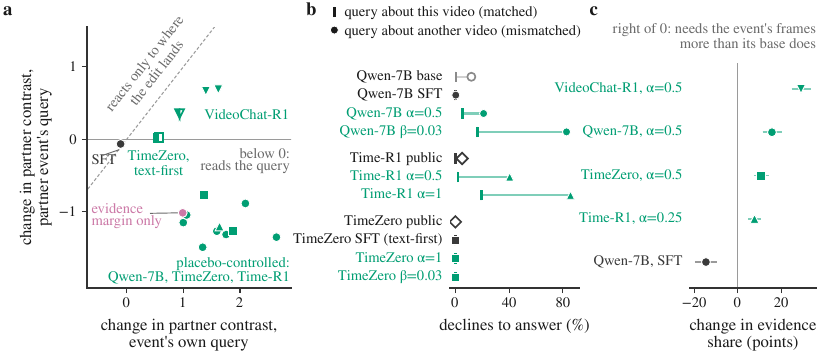}
\caption{\textbf{The placebo-controlled models respond to the query and to the evidence.} (a) Each point gives a model's change from its base (mask, nats per token) in the partner contrast, the drop when the event's frames are masked minus the drop when a partner event's frames are, with the event's own query (horizontal) and with the partner's (vertical). Marker shape gives the base (circle Qwen-7B, square TimeZero, triangle Time-R1, inverted triangle VideoChat-R1); half-filled markers were evaluated text-first, with the query before the video. (b) The tick shows how often a model answers \mbox{\texttt{There are none.}} to a query about this video, and the marker how often it does so for one about another video. (c) Each row gives the change from the base in the evidence share, the fraction of the accuracy under the placebo mask that the evidence mask removes, with its 95\% interval.}
\label{fig:readouts}
\end{figure}

\paragraph{Masking the event costs correct answers; masking elsewhere does not.}
The placebo-controlled models need the event's frames more than their bases do: every one we regenerate raises its evidence share over its base, by 6 to 29 points (Figure~\ref{fig:readouts}c; Table~\ref{tab:s6}). With the evidence frames masked, the \Rfive{} of the Qwen model at $\alpha{=}0.5$ falls from $0.523$ on the clean video to $0.087$. With as many frames masked elsewhere, every placebo-controlled model keeps its clean accuracy. With decoding forced to produce an interval, masking the event still costs the Qwen model 15 points of \Rfive{} more than the placebo mask, relative to its base, against 5 for SFT (Appendix~\ref{app:add-refusal}).

\paragraph{In tagged templates, most of the dependence sits on the framing tokens.}
Without the event's frames, a model that needs them can decline or place its interval elsewhere. In a tagged answer such as \texttt{<answer> 20.80 to 30.00 </answer>}, the two tags and the \emph{onset}, the first token after the opening tag, are the \emph{framing tokens}. The onset commits the model to a timestamp or a refusal. In the tagged templates of Qwen2.5-VL-7B, TimeZero and Time-R1, the placebo-controlled models place most of $\Gamma$ on the framing tokens. For the Qwen2.5-VL-7B model at $\alpha{=}0.5$, $\Gamma$ is $+0.79$ on the onset and $+0.86$ on the tags, against $+0.07$ on the timestamps. In generation, this shows mostly as refusals for Qwen2.5-VL-7B and as intervals moved off the event for the other two. VideoChat-R1's template fixes its tags and leaves only the timestamps free, so all of its $\Gamma$ sits on the timestamps, and with the event masked its placebo-controlled model misses it 15 points more often than its base. A Qwen2.5-VL-7B variant that applies the margins to the timestamp tokens alone reaches $\Gamma=+3.17$ on them with a similar accuracy gain (Appendix~\ref{app:add-tsonly}).

\paragraph{The dependence moves with the query.}
If the dependence belongs to the event the query names, it should move when the query names another event. On videos with two annotated events, A and B, we score A's interval and mask either A's frames or B's. The drop in the answer's log-likelihood under A's mask minus that under B's is the \emph{partner contrast}. We then ask B's query, still scoring A's interval. A model that reads the query is now hurt more by B's edit, so the change in its partner contrast from the base turns negative; one that reacts only to where the edit lands keeps it positive (Figure~\ref{fig:readouts}a). The sign reverses in ten of eleven placebo-controlled models from Qwen2.5-VL-7B, TimeZero and Time-R1 tested with both queries. The exception, near zero, is TimeZero evaluated \emph{text-first}, given the query before the video. SFT leaves the contrast unchanged, and in VideoChat-R1 the change under the partner's query shrinks by half to two thirds but keeps its sign.

\paragraph{The objective alone keeps or widens the base's gap on queries about another video.}
We also measure how often each model declines a query about another video. We pair every question with the annotation sentence of a different video. The Qwen2.5-VL-7B base declines these mismatched queries 12 points more often than matched ones, and its placebo-controlled model at $\alpha{=}0.5$ keeps the gap, at 16 points (Figure~\ref{fig:readouts}b). On Time-R1 the gap grows from 5 points for the public checkpoint to 39 at $\alpha{=}0.5$. Raising $\alpha$ to 1 on Time-R1 or lowering $\beta$ to 0.03 on Qwen2.5-VL-7B widens it further, while SFT loses the selectivity of its base and declines almost no query.

\paragraph{Unseen edits and a second dataset.}
Noise, Gaussian blur and patch scrambling never appear in training. Under patch scrambling, three placebo-controlled Qwen2.5-VL-7B and two TimeZero models reach $\Gamma$ of $+1.12$ to $+1.79$, well beyond the resolution. Under noise and blur their $\Gamma$ is small and every placebo drop stays near zero. On 600 further questions, the $\Gamma$ of the Qwen2.5-VL-7B model at $\alpha{=}0.5$ under these edits lies beyond each edit's own seed resolution of at most $0.05$ and grows with the strength of the edit (Appendix~\ref{app:add-operators}). On QVHighlights, scored against the TimeLens re-annotation \citep{zhang2026timelens}, every placebo-controlled model of Table~\ref{tab:s8} reaches $\Gamma$ of $+1.06$ to $+1.58$, which still falls with $\alpha$, and SFT stays at zero.

\section{Related work}
\label{sec:related}

\paragraph{Grounding models and what their benchmarks reward.}
The task is defined by Charades-STA \citep{charadessta2017,sigurdsson2016charades}, DiDeMo \citep{hendricks2017mcn}, ActivityNet Captions \citep{krishna2017activitynet} and QVHighlights \citep{lei2021qvhighlights}. Specialised architectures \citep{zhang2020vslnet,zhang2020tan,lin2023univtg} gave way to video LLMs \citep{lin2023videollava,ren2024timechat,huang2024vtimellm,qwen25vl2025,tpo2025} and reinforcement learning \citep{deepseekmath2024,guo2025deepseekr1} with IoU rewards \citep{wang2025timer1,videochatr1_2025,zhang2026timelens}. Blind baselines and probes show benchmarks reward priors \citep{otani2020hidden,yuan2021closer,tvbench2025,lei2022singleframe,xiao2021nextqa,nextgqa2023}, and debiasing methods reduce them \citep{lan2023debiased,yang2021dcm,nan2021ivg,jiang2024caem}. Both work at the benchmark level; we measure how much one model's correct answers rest on the event, and how training changes that.

\paragraph{Audits of evidence use.}
Deleting inputs is the basic move of perturbation attribution \citep{fong2017perturbation,petsiuk2018rise} and of rationale benchmarks, whose comprehensiveness score is a likelihood drop for a fixed answer \citep{deyoung2020eraser}. Faithfulness there means reliance rather than plausibility \citep{jacovi2020faithfulness,jain2020fresh}. Deletion without a matched control confounds such measurements \citep{hooker2019roar}. Video audits that compare answers with and without visual evidence include black-screen gaps \citep{leemvbench2026}, frame substitution \citep{tempcore2026}, dependency filters \citep{videooasis2026}, instance-level shortcut scores \citep{korkut2026traffic}, grounding stress tests \citep{reveal2026} and replay probes \citep{carve2026}. In grounding, consistency probes move the moment or rephrase the query \citep{consistency2025}. These audits ask whether a regenerated answer changes; the \audit{} asks how much the likelihood of one fixed answer rests on the evidence frames. The one matched control among them, the sham replay of \citet{carve2026}, varies whether an edit happens, whereas our placebo varies the position of an edit that always happens.

\paragraph{Rejecting irrelevant queries.}
RaTSG \citep{dong2024ratsg}, NA-VMR \citep{flanagan2025navmr} and RA-RFT \citep{lee2026rarft} train grounding models to reject a query whose event is not in the video, from irrelevant queries and rejection labels. Our audit uses only matching queries, and our objective uses no mismatched query and no rejection label. The two properties can come apart. RA-RFT leaves $\Gamma$ at zero and gives no interval for more mismatched queries than its base, while IoU-GRPO with our objective raises $\Gamma$ and, like IoU-GRPO alone, gives no interval for fewer mismatched queries than its base (Appendix~\ref{app:add-rarft}). On clip features, the released NA-VMR depends more than UniVTG on the event's clips (Appendix~\ref{app:add-navmr}).

\paragraph{Training on evidence.}
Evidence enters training through counterfactual, attribution and grounding rewards and data \citep{du2026crpo,feng2025videor1,wang2026videoktr,bao2026chartcf,luo2025ver,zheng2026f2g,vidground2026}, explanation constraints \citep{ross2017right,selvaraju2019hint}, and preference or decoding methods built on corrupted visual input \citep{leng2024vcd,favero2024m3id,xie2024vdpo,wang2024mdpo,wu2025svco,luo2025imageDPO}. Two methods add a control, random masks off the evidence \citep{defacto2025} or non-influential edits in an LLM faithfulness reward \citep{selfexpl2026}. The closest precedents put a likelihood drop inside a GRPO reward. TaRO \citep{zheng2026taro} measures it for sampled reasoning when the frames around the annotated boundaries are shuffled, and Evidence-RL \citep{evidencerl2026} for a sampled answer when an image region is neutralised, compared with area-matched other regions. Our objective applies the placebo contrast along time, scores the annotated answer rather than a sampled one, and bounds the model's own placebo drop (Section~\ref{sec:objective}). The mismatched-query test comes from selective prediction \citep{geifman2017selective,whitehead2022reliablevqa}.

\section{Conclusion}
\label{sec:conclusion}

We measured evidence dependence by scoring a fixed answer while editing either the event's frames or as many other frames of the same video. The post-training we audited adds accuracy along all three routes and does not raise evidence dependence. The evidence drop alone trains a model to react to any edit. A placebo contrast and a bound on the placebo drop make evidence dependence trainable, and the weight on clean supervision sets how much is learned. Added to IoU-reward training, the objective keeps the accuracy gain. Masking the event's frames makes models trained this way decline or miss the event more often than their base; masking as many other frames does not. The audit needs only an answer that can be scored and the location of its evidence, so evidence dependence can be reported, and trained, next to accuracy. Whether higher evidence dependence brings further practical gains, for example under distribution shift, is a question the audit now makes testable.

\FloatBarrier
\subsection*{AI use statement}
We used large language model assistants for two tasks. \emph{Retrieval and discovery:} to search for work on video temporal grounding, perturbation-based evaluation and training on visual evidence, and to summarise candidate papers; the authors read every cited paper and decided what to cite and how to position this work against it. \emph{Writing:} to edit drafts written by the authors. The paper reports many models and tests within the page limit, so most of these edits compress: restructuring sentences, merging statements that were made twice, shortening captions, and keeping terms and notation consistent across sections. The argument, the claims and the organisation of the paper are the authors'. We have reviewed all AI-assisted text and take full responsibility for the content of this paper.

\subsection*{Reproducibility statement}
Section~\ref{sec:audit} and Appendix~\ref{app:measure} define the audit: its question populations, placebo draws and intervals. Section~\ref{sec:objective} and Appendix~\ref{app:training} give the objective term by term, and Appendix~\ref{app:protocols} gives the inference protocol of each model and the training setup. The code is released at \url{https://github.com/jinzijie-bupt/Accuracy-Is-Not-Evidence}, and the question and training-pair identifiers, the per-question outputs and the LoRA adapters of the main models on Hugging Face; Appendix~\ref{app:release} lists them.

\bibliography{references}
\bibliographystyle{iclr2027_conference}

\appendix
\makeatletter
\renewcommand{\floatpagefraction}{0.86}
\setlength{\@fptop}{0pt}\setlength{\@fpsep}{14pt plus 6pt}\setlength{\@fpbot}{0pt plus 1fil}
\makeatother
\section*{Appendix overview}
The appendix follows the four steps of the main text. Appendix~\ref{app:measure} gives the audit of Section~\ref{sec:audit} in full, including what $\psi$ responds to and how $\Gamma$ relates to the scale of the likelihood. Appendix~\ref{app:extended} breaks down the accuracy that post-training adds (Section~\ref{sec:findings}). For the objective of Section~\ref{sec:objective}, Appendix~\ref{app:training} goes through each term, the earlier experiment its bound comes from, and the settings that control how much dependence is learned. The tests of Section~\ref{sec:readouts} on quantities the loss does not contain are in Appendix~\ref{app:shift}, and protocols, training setup and the release in Appendix~\ref{app:repro}. Appendix~\ref{app:added} reports additional experiments.

\FloatBarrier
\section{Measuring evidence dependence}
\label{app:measure}

\subsection{Estimand, populations and resolution}
\label{app:estimands}
\begin{figure}[t]
\centering
\includegraphics[width=\linewidth]{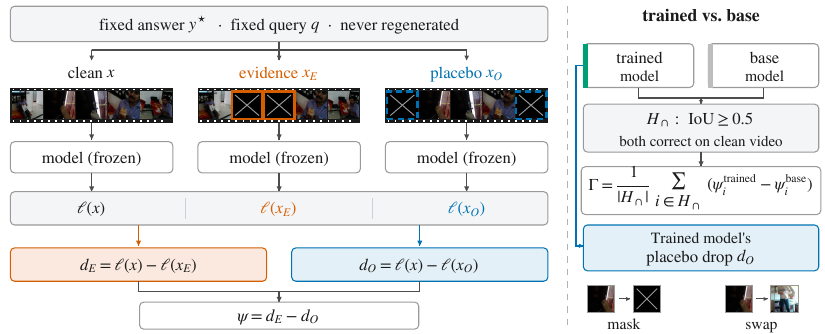}
\caption{\textbf{On each question, the audit scores one fixed answer while editing either the evidence frames or as many other frames.} The annotated answer $y^\star$ and the query $q$ are held fixed and scored by teacher forcing. Editing the frames inside the annotated interval gives $x_E$; editing as many frames outside it with the same operator gives the placebo $x_O$. The drops of the answer's log-likelihood under the two edits are $\dE$ and $\dO$, and $\psi=\dE-\dO$. The figure shows one mask draw; the audit averages $\dO$ over three draws. The right panel shows how a trained model is compared with its base: $\Gamma$ is the mean trained-minus-base change in $\psi$ over $H_\cap$, the questions both models localise on the clean video (IoU $\ge0.5$), and the trained model's placebo drop $\dO$ is reported next to it, under the mask and the swap operator.}
\label{fig:audit_pipeline}
\end{figure}

\paragraph{Readouts.}
Figure~\ref{fig:audit_pipeline} shows the audit on one question, and Table~\ref{tab:notation} collects the notation. $\dE$, $\dO$, $\psi$ and $\Gamma$ are defined in Section~\ref{sec:audit}. The generation readouts are \Rfive{}, which counts a prediction as correct when its IoU with the annotation is at least $0.5$ and counts an unparsed output as a miss; the parse rate; strict and wide abstention; and coverage, which is $1-$strict abstention, together with \Rfive{} on the answered questions. They also include the black-video answering rate and its \Rfive{}, the evidence share and follows (Appendix~\ref{app:regen}; Table~\ref{tab:s6}), and the mismatched-query abstention change (Table~\ref{tab:s4}).

\begin{table}[t]
\centering\scriptsize\setlength{\tabcolsep}{4pt}\renewcommand{\arraystretch}{1.1}
\caption{\textbf{This table collects the full notation; the table on page~\pageref{tab:notation-main} gives the symbols the main text uses.} Likelihood quantities are token means in nats, and behavioural quantities are fractions of questions. In every table, brackets give 95\% bootstrap intervals over videos unless its caption says otherwise.}
\label{tab:notation}
\begin{tabular}{@{}lp{10.2cm}@{}}
\toprule
Symbol & Meaning \\
\midrule
$y^\star$, $x$, $q$ & annotated interval, video, query \\
$x_E$, $x_O$ & video with the evidence frames edited; with as many frames outside $y^\star$ edited (placebo) \\
$\dE$, $\dO$ & drop of the answer's log-likelihood under $x_E$ and under $x_O$ (mean of three placebo draws) \\
$\psi=\dE-\dO$ & evidence dependence of one model on one question \\
$H_\cap$, $n_\cap$ & questions both the trained model and its base localise (IoU $\ge0.5$); their number \\
$\Gamma$ or $\Gm$; $\Gs$ & mean trained-minus-base change in $\psi$ on $H_\cap$ under mask (primary) and under swap \\
sign fraction & fraction of $H_\cap$ on which the trained model's $\psi$ exceeds its base's; scale-free companion of $\Gamma$ \\
$\Gamma_c$ & mean trained-minus-base change on $H_\cap$ in the bounded index $c=(\dE-\dO)/(|\dE|+|\dO|+0.05)$; scale-free companion of $\Gamma$ \\
$\pi(q)$ & partner contrast: drop in the log-likelihood of $y^\star_i$ when the question's evidence frames $E_i$ are masked minus the drop when a partner event's frames $E_j$ are masked, under query $q$ (Appendix~\ref{app:qspec}) \\
$\psi_{\mathrm{own}}$, $\psi_{\mathrm{cross}}$ & $\pi(q_i)$ and $\pi(q_j)$: the partner contrast under the question's own query and under the partner's query; $\Delta$: trained minus base, mean over the partnered questions of $H_\cap$ \\
$\Gamma_{\mathrm{spec}}$ & $\Delta\psi_{\mathrm{own}}-\Delta\psi_{\mathrm{cross}}$: the part of the change in the partner contrast that depends on which query is asked \\
$\Gamma_{\mathrm{agn}}$, $\Gamma_{\mathrm{sem}}$ & trained-minus-base change in $\psi$ under the partner's query (placebo drawn outside both events); $\Gamma-\Gamma_{\mathrm{agn}}$ \\
evidence share & $1-a_E/a_O$, with $a_E$ and $a_O$ the \Rfive{} of predictions regenerated with the evidence frames masked and with as many placebo frames masked (first 1{,}000 test questions) \\
follows & fraction of $H_\cap$ on which the prediction, regenerated after the event's frames are exchanged with an equally long block elsewhere, has IoU $\ge0.5$ with the annotated interval moved to that block \\
$\alpha$, $\beta$, $w_{\mathrm{op}}$ & objective weights: clean supervision, margin scale, per-operator weights \\
\bottomrule
\end{tabular}
\end{table}

\paragraph{Populations.}
The likelihood audit runs on the first 200 test questions (79 videos), with three placebo draws per question. Only questions a model localises correctly are scored, and $\Gamma$ uses $H_\cap$, whose size $n_\cap$ ranges from 55 to 143 on Charades-STA and from 39 to 82 on QVHighlights; Tables~\ref{tab:scale} and~\ref{tab:s8} print it per row. We varied the question set in two ways. With one question set held fixed across four placebo-controlled Qwen2.5-VL-7B models ($n{=}87$), every sign is preserved. With each contrast computed on each model's own full hit set instead of the intersection, $\Gamma$ moves by at most $0.10$ and keeps its sign in every recomputed mask contrast except IoU-GRPO's, which goes from $0.00$ to $-0.01$. On the 200 audit questions the Qwen base reaches \Rfive{} $0.560$; Table~\ref{tab:audit} gives its value on the full split.

\paragraph{Placebo construction.}
The placebo edits as many frames as the evidence set, drawn uniformly from outside the annotated interval, averaged over three draws. It is defined by position alone, so its pool includes the frames of any other annotated event in the video. Each run draws its own placebo frames; the behavioural audit uses one set of draws for all models.

\paragraph{Training seeds and resolution.}
Intervals are 95\% bootstrap intervals over videos, with 4{,}000 resamples for the accuracy changes against the base, the evidence share and follows, and 2{,}000 for every other quantity. Two independent training seeds of the Qwen model at $\alpha{=}1$ agree in $\Gamma$ ($+1.03$ $[0.80,1.27]$ and $+1.13$ $[0.92,1.34]$) and in accuracy (gains of $6.7$ and $6.5$ points in \Rfive{}), and they give the same query-crossed decomposition (Table~\ref{tab:s11}). Audited against each other, with one seed as the base, they give $\Gamma=+0.06$ $[0.02,0.10]$ under mask and $+0.15$ $[-0.01,0.31]$ under swap, and an accuracy resolution of $1.1$ points in domain and $4.9$ out of domain. At $\alpha{=}0.5$ the first seed gives $\Gamma$ $+1.63$ $[1.36,1.89]$ and a gain of $4.4$ $[3.1,5.7]$ points, and a second seed gives $+1.48$ $[1.24,1.71]$ and $5.5$ $[4.2,6.8]$. Audited against the first, the second seed gives $\Gamma=-0.15$ $[-0.23,-0.08]$ under mask and $-0.05$ $[-0.20,0.12]$ under swap. The audit's resolution (Section~\ref{sec:audit}) is the largest absolute interval end over the two pairs. The tests of the 16-frame experiment (Appendix~\ref{app:prereg}) are Holm-corrected \citep{holm1979simple}.

\subsection{What \texorpdfstring{$\psi$}{psi} responds to}
\label{app:validity}
Table~\ref{tab:battery} compares $\psi$ or $\Gamma$ with controls that each differ from the audit in one respect.

\begin{table}[ht]
\centering\papertablestyle
\caption{\textbf{$\psi$ grows with how much of the evidence is edited, and $\Gamma$ keeps its sign across placebo shapes, question sets, query wordings and unseen operators.} Each row names a property of the audit, the control that $\psi$ or $\Gamma$ is compared with, which differs from the audit in that one respect, and the result. Rows marked $^\ast$ were run in the 16-frame setup of Appendix~\ref{app:prereg}.}
\label{tab:battery}
\begin{tabular}{@{}>{\raggedright\arraybackslash}p{3.3cm}>{\raggedright\arraybackslash}p{4.9cm}>{\raggedright\arraybackslash}p{5.1cm}@{}}
\toprule
Property & Compared with & Result \\
\midrule
Ring of adjacent frames$^\ast$ & frames just outside the interval & drop lies between the evidence and placebo drops \\
Edit severity$^\ast$ & edits removing more or less of the evidence & drop rises monotonically with severity \\
Sub-step jitter$^\ast$ & each evidence frame replaced by its neighbour half a sampling step away, and likewise at the placebo & difference of the two drops $+0.004$ for the base \\
Contiguous placebo & a placebo of consecutive frames & $\Gamma$ $+1.33$ to $+1.65$ for the placebo-controlled Qwen2.5-VL-7B, TimeZero and VideoChat-R1 at $\alpha{=}0.5$ and Time-R1 at $\alpha{=}0.25$, $-0.07$ to $+0.03$ for the others \\
Selection on $H_\cap$ & a fixed cohort; each model's own hit set & signs kept (Appendix~\ref{app:estimands}) \\
Query crossed to another event & the partner event's query & Section~\ref{sec:readouts}; Table~\ref{tab:s11} \\
Query paraphrased & a second human-written query & contrast retained (Appendix~\ref{app:qspec}) \\
Operators unseen in training & noise, Gaussian blur, patch scrambling & 15/15 signs kept (Table~\ref{tab:unseen}) \\
\bottomrule
\end{tabular}
\end{table}

\subsection{\texorpdfstring{$\Gamma$}{Gamma} and the scale of the likelihood}
\label{app:scale}
$\Gamma$ does not follow the change in clean-video likelihood that training produces. Table~\ref{tab:scale} sets each trained model's $\Gamma$ next to that change and gives two versions of $\Gamma$ that do not depend on scale.

\begin{table}[tbp]
\centering\scriptsize\setlength{\tabcolsep}{3pt}\renewcommand{\arraystretch}{1.04}
\caption{\textbf{$\Gamma$ does not follow the change in clean-video NLL.} Each row pairs a model with its base: our trained models, the three public checkpoints against the Qwen2.5-VL-7B base (the pairs of Table~\ref{tab:audit}), and TimeLens-8B against Qwen3-VL-8B. All values use the mask operator on $H_\cap$, the $n_\cap$ questions both models localise. The columns give the clean-video negative log-likelihood (NLL) of the annotated answer in nats per token for the base and the trained model, then $\Gm$, its sign fraction (the fraction of $H_\cap$ on which $\psi$ rises) and $\Gamma_c$, the change in the bounded index $c=(\dE-\dO)/(|\dE|+|\dO|+0.05)$, with its 95\% interval. Text-first rows put the query before the video.}
\label{tab:scale}
\begin{tabular}{@{}>{\raggedright\arraybackslash}p{5.6cm}ccccc@{}}
\toprule
Configuration & $n_\cap$ & clean NLL base / trained & $\Gm$ & sign frac. & $\Gamma_c$ \\
\midrule
\addlinespace[1pt]\multicolumn{6}{@{}l}{\textit{Charades-STA}} \\
Qwen2.5-VL-7B, IoU-GRPO & 104 & 1.35 / 1.39 & 0.00 & 0.49 & +0.02 \ci{$-$0.09}{0.12} \\
Qwen2.5-VL-7B, SFT & 112 & 1.38 / 0.39 & $-$0.09 & 0.40 & 0.00 \ci{$-$0.14}{0.13} \\
Qwen2.5-VL-7B, evidence margin only & 104 & 1.36 / 0.40 & +0.54 & 0.74 & $-$0.34 \ci{$-$0.45}{$-$0.22} \\
Qwen2.5-VL-7B, placebo-controlled $\alpha{=}0.25$ & 101 & 1.34 / 0.59 & +2.03 & 1.00 & +0.35 \ci{0.25}{0.44} \\
Qwen2.5-VL-7B, placebo-controlled $\alpha{=}0.5$ & 112 & 1.38 / 0.46 & +1.63 & 0.97 & +0.40 \ci{0.30}{0.51} \\
Qwen2.5-VL-7B, placebo-controlled $\alpha{=}0.5$, $\beta{=}0.03$ & 89 & 1.35 / 0.39 & +1.36 & 0.73 & +0.26 \ci{0.14}{0.38} \\
Qwen2.5-VL-7B, placebo-controlled $\alpha{=}0.5$, 8,434 pairs, trained on 16 frames & 89 & 1.34 / 2.05 & +0.23 & 0.83 & +0.19 \ci{0.09}{0.29} \\
Qwen2.5-VL-7B, placebo-controlled $\alpha{=}1$ & 102 & 1.35 / 0.40 & +1.03 & 0.80 & +0.28 \ci{0.18}{0.37} \\
Qwen2.5-VL-7B, placebo-controlled $\alpha{=}1$ (second seed) & 100 & 1.35 / 0.39 & +1.13 & 0.88 & +0.28 \ci{0.20}{0.38} \\
Qwen2.5-VL-7B, placebo-controlled $\alpha{=}1$, $\beta{=}0.03$ & 105 & 1.37 / 0.38 & +0.82 & 0.83 & +0.34 \ci{0.23}{0.46} \\
Qwen2.5-VL-7B, uniform weights & 103 & 1.35 / 0.48 & +1.99 & 1.00 & +0.34 \ci{0.26}{0.45} \\
Qwen2.5-VL-7B, no placebo bound & 87 & 1.35 / 0.39 & +2.05 & 1.00 & +0.11 \ci{$-$0.02}{0.22} \\
Qwen2.5-VL-7B, no bound, uniform & 98 & 1.35 / 0.41 & +2.62 & 1.00 & +0.14 \ci{0.04}{0.26} \\
Qwen2.5-VL-7B, two-stage, $\alpha{=}0.5$ & 84 & 1.35 / 0.39 & +2.10 & 1.00 & +0.36 \ci{0.25}{0.49} \\
Qwen2.5-VL-7B, two-stage, $\beta{=}0.03$ & 86 & 1.36 / 0.39 & +2.75 & 0.98 & +0.38 \ci{0.28}{0.49} \\
TimeZero, placebo-controlled $\alpha{=}0.25$ & 133 & 1.36 / 0.54 & +2.00 & 0.99 & +0.39 \ci{0.30}{0.47} \\
TimeZero, placebo-controlled $\alpha{=}0.5$ & 135 & 1.36 / 0.49 & +1.40 & 0.99 & +0.35 \ci{0.27}{0.44} \\
TimeZero, placebo-controlled $\alpha{=}0.5$, $\beta{=}0.03$ & 131 & 1.36 / 0.39 & +0.32 & 0.91 & +0.27 \ci{0.17}{0.36} \\
TimeZero, placebo-controlled $\alpha{=}0.5$, text-first & 143 & 1.41 / 0.47 & +1.29 & 0.98 & +0.58 \ci{0.45}{0.71} \\
TimeZero, placebo-controlled $\alpha{=}1$ & 130 & 1.36 / 0.43 & +1.07 & 0.93 & +0.37 \ci{0.27}{0.47} \\
Time-R1, placebo-controlled $\alpha{=}0.1$ & 130 & 1.30 / 0.85 & +2.40 & 1.00 & +0.37 \ci{0.27}{0.46} \\
Time-R1, placebo-controlled $\alpha{=}0.25$ & 131 & 1.34 / 0.58 & +1.88 & 1.00 & +0.42 \ci{0.31}{0.54} \\
Time-R1, placebo-controlled $\alpha{=}0.5$ & 127 & 1.29 / 0.42 & +1.46 & 0.98 & +0.32 \ci{0.23}{0.42} \\
Time-R1, placebo-controlled $\alpha{=}1$ & 102 & 1.31 / 0.39 & +0.95 & 0.86 & +0.25 \ci{0.14}{0.34} \\
VideoChat-R1, placebo-controlled $\alpha{=}0.25$ & 136 & 0.58 / 0.58 & +1.63 & 0.99 & +0.42 \ci{0.33}{0.51} \\
VideoChat-R1, placebo-controlled $\alpha{=}0.5$ & 132 & 0.57 / 0.42 & +1.43 & 1.00 & +0.44 \ci{0.36}{0.53} \\
VideoChat-R1, placebo-controlled $\alpha{=}1$ & 129 & 0.58 / 0.38 & +1.20 & 1.00 & +0.40 \ci{0.32}{0.48} \\
TimeLens-8B, placebo-controlled $\alpha{=}0.5$ & 85 & 1.98 / 0.49 & +1.36 & 1.00 & +0.64 \ci{0.46}{0.83} \\
TimeZero (public) & 103 & 1.35 / 1.36 & $-$0.08 & 0.42 & $-$0.11 \ci{$-$0.25}{0.03} \\
Time-R1 (public) & 101 & 1.35 / 1.28 & +0.02 & 0.60 & $-$0.02 \ci{$-$0.10}{0.06} \\
VideoChat-R1 (public) & 101 & 0.66 / 0.56 & +0.05 & 0.71 & 0.00 \ci{$-$0.13}{0.12} \\
TimeLens-8B (public) against Qwen3-VL-8B & 85 & 0.51 / 2.00 & $-$0.16 & 0.29 & $-$0.42 \ci{$-$0.65}{$-$0.23} \\
\addlinespace[1pt]\multicolumn{6}{@{}l}{\textit{Charades, TimeLens labels}} \\
Qwen2.5-VL-7B, SFT & 55 & 1.22 / 0.40 & $-$0.03 & 0.51 & +0.05 \ci{$-$0.09}{0.18} \\
Qwen2.5-VL-7B, placebo-controlled $\alpha{=}0.5$ & 57 & 1.21 / 0.57 & +1.21 & 0.91 & +0.41 \ci{0.26}{0.57} \\
Qwen2.5-VL-7B, placebo-controlled $\alpha{=}1$ & 58 & 1.21 / 0.41 & +0.87 & 0.90 & +0.41 \ci{0.29}{0.54} \\
TimeZero, placebo-controlled $\alpha{=}0.5$ & 63 & 1.30 / 0.60 & +1.00 & 0.92 & +0.50 \ci{0.37}{0.64} \\
Time-R1, placebo-controlled $\alpha{=}0.25$ & 62 & 1.23 / 0.79 & +1.43 & 0.98 & +0.53 \ci{0.40}{0.67} \\
VideoChat-R1, placebo-controlled $\alpha{=}0.5$ & 57 & 0.54 / 0.46 & +0.98 & 0.98 & +0.48 \ci{0.36}{0.59} \\
TimeLens-8B, SFT & 75 & 0.23 / 0.44 & $-$0.21 & 0.05 & $-$0.13 \ci{$-$0.17}{$-$0.09} \\
TimeLens-8B, placebo-controlled $\alpha{=}0.5$ & 70 & 0.24 / 0.49 & +0.95 & 0.96 & +0.11 \ci{0.04}{0.17} \\
\addlinespace[1pt]\multicolumn{6}{@{}l}{\textit{QVHighlights}} \\
Qwen2.5-VL-7B, SFT & 52 & 1.18 / 0.36 & $-$0.02 & 0.46 & +0.04 \ci{$-$0.03}{0.13} \\
Qwen2.5-VL-7B, placebo-controlled $\alpha{=}0.5$ & 43 & 1.16 / 0.88 & +1.30 & 1.00 & +0.27 \ci{0.17}{0.38} \\
Qwen2.5-VL-7B, placebo-controlled $\alpha{=}1$ & 39 & 1.13 / 0.63 & +1.06 & 0.95 & +0.19 \ci{0.11}{0.28} \\
TimeZero, placebo-controlled $\alpha{=}0.5$, text-first & 51 & 1.16 / 0.80 & +0.92 & 1.00 & +0.27 \ci{0.18}{0.37} \\
\bottomrule
\end{tabular}
\end{table}

\paragraph{Units and scale.}
Likelihoods in the audit are token means in nats. In the objective, the evidence margin and the placebo contrast compare log-likelihoods summed over the answer tokens, while the placebo bound, like the audit, uses the token-mean placebo drop, and the clean-supervision term is a token-mean NLL. Under sequence sums, $\Gamma$ scales by roughly the answer length ($\approx18$ tokens).

\paragraph{Scale-free versions and clean likelihood.}
The two scale-free versions are the fraction of questions on which $\psi^{\mathrm{trained}}>\psi^{\mathrm{base}}$, and the mean change $\Gamma_c$ in a bounded contrast index $c=(\dE-\dO)/(|\dE|+|\dO|+0.05)$. Correlations are over the 32 post-trained models on Charades-STA in the table. SFT lowers the clean-video NLL as much as the placebo-controlled models do (from $1.38$ to $0.39$ nats per token) while its $\Gamma$ is at or below zero, and across the 32 models $\Gamma$ is at most weakly related to the change in clean NLL (Spearman $-0.20$). The sign fraction orders the models as $\Gamma$ does (Spearman $0.84$). It is $0.73$ or higher for every placebo-controlled model in the table, $0.60$ or lower for SFT, IoU-GRPO, TimeZero, Time-R1 and TimeLens-8B, and $0.71$ for VideoChat-R1, whose small $\Gamma$ ($+0.05$) is spread over many questions. The bounded index $\Gamma_c$ (Spearman $0.59$ with $\Gamma$) also penalises the placebo drop relative to the evidence drop, and so separates the models trained without placebo control more sharply. The evidence-margin-only model has $\Gamma=+0.54$ but $\Gamma_c=-0.34$ $[-0.45,-0.22]$, and the two models trained without the placebo bound lose most of their $\Gamma_c$. Under $\Gamma_c$, SFT's interval contains zero, and TimeLens-8B against Qwen3-VL-8B stays negative ($-0.42$ $[-0.65,-0.23]$; sign fraction $0.29$).

\subsection{Single questions, a larger base and other inputs}
\label{app:qualitative}
\begin{figure}[!htb]
\centering
\includegraphics[width=0.84\linewidth]{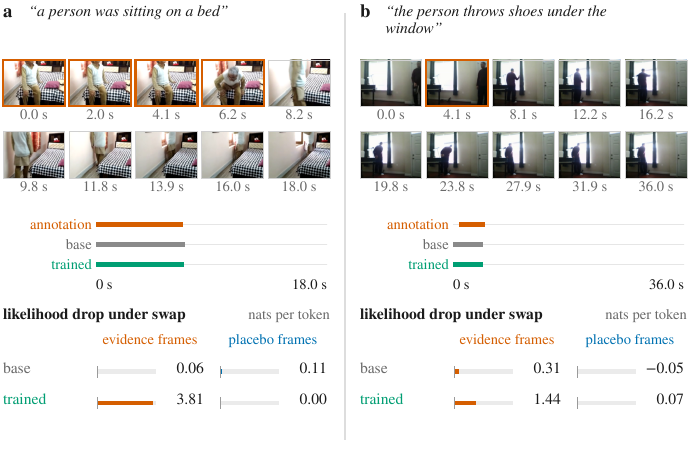}
\caption{\textbf{Two models give the same correct interval for different reasons.} Panels (a) and (b) each show a test question on which the Qwen2.5-VL-7B base and its placebo-controlled model ($\alpha{=}0.5$) predict the same interval, with IoU above $0.7$. The top rows show the native frames with the annotated ones outlined, followed by the annotated interval and both predictions. The bars below give the drop in the answer's log-likelihood (nats per token) when the evidence frames are swapped out and when as many frames outside the interval are swapped the same way. In (a), for the same correct interval, the base loses $0.06$ nats per token under the evidence swap and its trained model $3.81$, with placebo drops of $0.11$ and $0.00$.}
\label{fig:cases}
\end{figure}

Figure~\ref{fig:cases} shows two test questions on which the base and its placebo-controlled model give the same correct interval.

\paragraph{A larger base.}
At 32B, the Qwen2.5-VL-32B base localises at \Rfive{} $0.475$, answers $16\%$ of black videos and has $\psi=+0.18$ $[0.16, 0.20]$, at the level of the Qwen2.5-VL-7B base (Section~\ref{sec:audit}).

\paragraph{Beyond video.}
The audit also applies to inputs divided into other units. On document QA, with SlideVQA \citep{slidevqa2023} pages in place of frames, Qwen2.5-VL-7B and Qwen3-VL-8B lose $2.00$ and $2.61$ nats per token of the annotated answer's likelihood when the evidence pages are masked and $0.00$ and $0.06$ when as many other pages are.

\section{What post-training accuracy is made of}
\label{app:extended}
\subsection{Every configuration against its base}
\label{app:matrix}
Table~\ref{tab:s1} pairs the trained configurations of Sections~\ref{sec:findings} and~\ref{sec:objective} with the base each was trained from. It extends Table~\ref{tab:audit} with coverage, accuracy on answered questions, both edit operators and $n_\cap$, and is the reference for the trained configurations of Section~\ref{sec:readouts}. Coverage equals the parse rate in every configuration, because every output without an interval is the refusal string. At $\beta{=}0.03$ the model declines more often and, on the questions it answers, gains the most over its base among the Qwen models at $\alpha{=}0.5$. $\Gs$ tracks $\Gm$, mostly at lower magnitude.

\begin{table}[tbp]
\centering\scriptsize\setlength{\tabcolsep}{3pt}\renewcommand{\arraystretch}{1.05}
\caption{\textbf{The full audit matrix on Charades-STA compares every trained configuration with the base it was trained from.} Each block of rows belongs to one base model. A base row gives its \Rfive{} and, in the column \emph{ans.}, its \Rfive{} on the questions it answers; a trained row gives the paired changes of these two values from its base in points, the second on the questions the trained model answers. A $^\dagger$ marks a change that is not paired by question but is the difference of the two models' values. The column \emph{ord.} gives the message order, video first (v) or query text first (t); \emph{cov.} is the coverage, $1-$strict abstention; $\Gm$ and $\Gs$ are in nats per token on the $n_\cap$ questions both models localise. The last row is a constant $(0,10)$\,s window, which uses no model.}
\label{tab:s1}
\begin{tabular}{@{}>{\raggedright\arraybackslash}p{5.0cm}ccccccc@{}}
\toprule
Model / objective & ord. & \Rfive{} / $\Delta$ & cov. & ans. / $\Delta$ & $\Gm$ & $\Gs$ & $n_\cap$ \\
\midrule
\addlinespace[1pt]\multicolumn{8}{@{}l}{\textit{Qwen2.5-VL-7B}} \\
base & v & 0.480 & 0.994 & 0.482 & \NA & \NA & \NA \\
placebo-controlled $\alpha{=}1$ & v & +6.7 \ci{5.2}{8.2} & 0.937 & +9.5 & +1.03 & +1.00 & 102 \\
placebo-controlled $\alpha{=}0.5$ & v & +4.4 \ci{3.1}{5.7} & 0.957 & +6.1 & +1.63 & +1.41 & 112 \\
placebo-controlled $\alpha{=}0.5$, $\beta{=}0.03$ & v & +1.2 \ci{$-$0.6}{2.9} & 0.831 & +9.6 & +1.36 & +1.16 & 89 \\
placebo-controlled $\alpha{=}0.25$ & v & +2.1 \ci{0.8}{3.2} & 0.959 & +3.4 & +2.03 & +1.84 & 101 \\
placebo-controlled $\alpha{=}1$, $\beta{=}0.03$ & v & +6.0$^\dagger$ & 0.926 & +10.2$^\dagger$ & +0.82 & +0.72 & 105 \\
SFT & v & +9.4 \ci{7.9}{11.0} & 1.00 & +9.4 & $-$0.09 & $-$0.03 & 112 \\
IoU-GRPO & v & +13.6 \ci{11.9}{15.4} & 0.999 & +13.4$^\dagger$ & 0.00 & +0.02 & 104 \\
mDPO-style & v & +8.8 \ci{7.3}{10.4} & 0.997 & +8.9 & $-$0.03 & +0.01 & 112 \\
\addlinespace[1pt]\multicolumn{8}{@{}l}{\textit{Qwen2.5-VL-7B: objective ablation at $\alpha{=}0.5$}} \\
uniform weights & v & +5.4 \ci{4.1}{6.7} & 0.974 & +6.4 & +1.99 & +1.89 & 103 \\
no placebo bound & v & $-$3.8 \ci{$-$5.4}{$-$2.1} & 0.802 & +5.7 & +2.05 & +2.34 & 87 \\
no bound, uniform & v & +3.8 \ci{2.4}{5.2} & 0.938 & +6.6 & +2.62 & +2.60 & 98 \\
evidence margin only & v & +6.7 \ci{5.2}{8.3} & 0.959 & +8.8$^\dagger$ & +0.54 & +1.09 & 104 \\
\addlinespace[1pt]\multicolumn{8}{@{}l}{\textit{Qwen2.5-VL-7B: training budget}} \\
placebo-controlled $\alpha{=}0.5$, 4,800 pairs & v & +8.1 \ci{6.7}{9.5} & 0.981 & +8.7 & +2.78 & +2.58 & 105 \\
placebo-controlled $\alpha{=}0.5$, 2,400 pairs, trained on 16 frames & v & +2.5$^\dagger$ & 0.975 & +3.6$^\dagger$ & +0.06 & +0.04 & 103 \\
placebo-controlled $\alpha{=}0.5$, 8,434 pairs, trained on 16 frames & v & $-$5.7$^\dagger$ & 0.986 & $-$5.3$^\dagger$ & +0.23 & +0.24 & 89 \\
\addlinespace[1pt]\multicolumn{8}{@{}l}{\textit{TimeZero}} \\
base & v & 0.677 & 1.00 & 0.677 & \NA & \NA & \NA \\
base, text-first & t & 0.705 & 1.00 & 0.705 & \NA & \NA & \NA \\
placebo-controlled $\alpha{=}0.5$ & v & +0.5 \ci{$-$0.7}{1.5} & 0.999 & +0.5 & +1.40 & +1.31 & 135 \\
placebo-controlled $\alpha{=}0.25$ & v & +0.2$^\dagger$ & 1.00 & +0.2$^\dagger$ & +2.00 & +1.66 & 133 \\
placebo-controlled $\alpha{=}0.5$, text-first & t & $-$2.6$^\dagger$ & 1.00 & $-$2.6$^\dagger$ & +1.29 & +1.13 & 143 \\
placebo-controlled $\alpha{=}1$ & v & $-$0.6$^\dagger$ & 1.00 & $-$0.6$^\dagger$ & +1.07 & +0.92 & 130 \\
placebo-controlled $\alpha{=}0.5$, $\beta{=}0.03$ & v & $-$1.0 \ci{$-$2.1}{0.1} & 1.00 & $-$1.0 & +0.32 & +0.23 & 131 \\
\addlinespace[1pt]\multicolumn{8}{@{}l}{\textit{Time-R1}} \\
base & v & 0.601 & 1.00 & 0.601 & \NA & \NA & \NA \\
placebo-controlled $\alpha{=}0.25$ & v & +1.6$^\dagger$ & 0.996 & +1.8$^\dagger$ & +1.88 & +1.66 & 131 \\
placebo-controlled $\alpha{=}0.5$ & v & +0.9 \ci{$-$0.2}{2.0} & 0.982 & +2.0$^\dagger$ & +1.46 & +1.20 & 127 \\
placebo-controlled $\alpha{=}0.1$ & v & +0.3$^\dagger$ & 0.998 & +0.4$^\dagger$ & +2.40 & +2.20 & 130 \\
placebo-controlled $\alpha{=}1$ & v & $-$8.4$^\dagger$ & 0.805 & +4.2$^\dagger$ & +0.95 & +0.99 & 102 \\
\addlinespace[1pt]\multicolumn{8}{@{}l}{\textit{VideoChat-R1}} \\
base & v & 0.668 & 1.00 & 0.668 & \NA & \NA & \NA \\
placebo-controlled $\alpha{=}0.5$ & v & $-$2.7 \ci{$-$4.2}{$-$1.2} & 1.00 & $-$2.7 & +1.43 & +0.89 & 132 \\
placebo-controlled $\alpha{=}0.25$ & v & $-$3.2 \ci{$-$4.6}{$-$1.8} & 1.00 & $-$3.2 & +1.63 & +0.92 & 136 \\
placebo-controlled $\alpha{=}1$ & v & $-$2.4 \ci{$-$3.8}{$-$0.9} & 1.00 & $-$2.4 & +1.20 & +0.85 & 129 \\
\addlinespace[1pt]\multicolumn{8}{@{}l}{\textit{TimeLens-8B}} \\
base & v & 0.417 & 1.00 & 0.417 & \NA & \NA & \NA \\
placebo-controlled $\alpha{=}0.25$ & v & +17.9 \ci{15.9}{20.0} & 1.00 & +17.9 & +1.72 & +1.15 & 84 \\
placebo-controlled $\alpha{=}0.5$ & v & +20.9 \ci{18.9}{22.8} & 1.00 & +20.9 & +1.36 & +0.88 & 85 \\
placebo-controlled $\alpha{=}1$ & v & +23.2 \ci{21.4}{25.1} & 1.00 & +23.2 & +0.92 & +0.66 & 87 \\
\midrule
constant window $(0,10)$\,s, no model & \NA & 0.317 & 1.00 & 0.317 & \NA & \NA & \NA \\
\bottomrule
\end{tabular}
\end{table}

\paragraph{A second backbone.}
TimeLens-8B \citep{zhang2026timelens}, post-trained for accuracy from Qwen3-VL-8B-Instruct on TimeLens-100K, the automatically annotated training set released with the TimeLens benchmark, is audited against Qwen3-VL-8B-Instruct under the TimeLens template, with the video first, on 200 questions ($n_\cap=85$). Its $\Gm$ is $-0.16$ $[-0.26,-0.08]$ (Section~\ref{sec:findings}), and its $\Gs$ is $-0.14$ $[-0.20,-0.09]$. The change comes from the evidence side: on $H_\cap$ the evidence drop falls and the placebo drop stays where it was (Table~\ref{tab:s7}).

\subsection{The newly correct answers}
\label{app:newly}
Table~\ref{tab:s12} breaks IoU-GRPO's gain down on the \emph{newly correct} set, the questions that the trained model gets right on the clean video and its base gets wrong, with two placebo-controlled models for comparison. IoU-GRPO's newly correct questions are solvable by the constant window less often than all questions are ($26\%$ against $32\%$), and $4.5\%$ of them survive a black video. Regenerated with the evidence frames masked, IoU-GRPO's newly correct answers lose $64$ points $[54, 72]$ more than with matched placebo masking. Table~\ref{tab:newly_psi} compares, on the audited questions, a model's $\psi$ on its newly correct questions with the base model's $\psi$ on the questions the base gets right. For SFT, the newly correct questions carry less evidence dependence than the base's own correct answers ($+0.11$ against $+0.18$). For IoU-GRPO, TimeZero and Time-R1 the interval of the difference includes zero, and for the placebo-controlled models the difference is positive and larger (Section~\ref{sec:findings} gives the range); these sets are unpaired, and those of the placebo-controlled models hold 11 to 28 newly correct questions each.

\begin{table}[!htbp]
\centering\scriptsize\setlength{\tabcolsep}{3pt}\renewcommand{\arraystretch}{1.04}
\caption{\textbf{Most of IoU-GRPO's newly correct answers survive masking as many placebo frames, and most are lost when the evidence frames are masked.} $|\mathrm{NC}|$ is the number of questions a trained model localises on the clean video and its base does not. The other columns give the percentage of that set that the constant $(0,10)$\,s window also solves (on all questions it solves $32\%$), that stays correct on an all-black video, and that stays correct when regenerated with the evidence frames or as many placebo frames masked. The last two columns are given for IoU-GRPO only, on the 181 of its newly correct questions that are among the 1{,}000 questions of the behavioural audit. Brackets give 95\% intervals.}
\label{tab:s12}
\begin{tabular}{@{}lccccc@{}}
\toprule
Objective & $|\mathrm{NC}|$ & const-win (\%) & black (\%) & evid-mask (\%) & placebo-mask (\%) \\
\midrule
IoU-GRPO & 666 & 26.0 \ci{22.0}{29.7} & 4.5 \ci{2.9}{6.2} & 24.3 \ci{17.6}{31.7} & 87.8 \ci{83.0}{92.1} \\
placebo-controlled $\alpha{=}0.5$ & 273 & 19.8 \ci{14.7}{25.3} & 0.0 \ci{0.0}{0.0} & \NA & \NA \\
placebo-controlled $\alpha{=}1$ & 445 & 24.7 \ci{19.9}{29.5} & 0.0 \ci{0.0}{0.0} & \NA & \NA \\
\bottomrule
\end{tabular}
\end{table}

\begin{table}[!htbp]
\centering\scriptsize\setlength{\tabcolsep}{3pt}\renewcommand{\arraystretch}{1.04}
\caption{\textbf{$\psi$ on the newly correct questions stays within $0.1$ nats per token of the base's after accuracy training and far exceeds it after placebo-controlled training.} All values come from the likelihood audit with the mask operator on 200 questions and are in nats per token. $|\mathrm{NC}|$ is the number of questions the trained model localises and its base does not, and $\psi^{\mathrm{trained}}(\mathrm{NC})$ is the trained model's $\psi$ on them. $\psi^{\mathrm{base}}$ is the base's $\psi$ on the questions the base localises, in the audit paired with that model. The column \emph{difference} is the first minus the second; the two sets are not paired, and brackets give 95\% intervals.}
\label{tab:newly_psi}
\begin{tabular}{@{}lcccc@{}}
\toprule
Configuration & $|\mathrm{NC}|$ & $\psi^{\mathrm{trained}}(\mathrm{NC})$ & $\psi^{\mathrm{base}}$ (base hits) & difference \\
\midrule
Qwen2.5-VL-7B, SFT & 23 & +0.11 \ci{0.06}{0.15} & +0.18 \ci{0.12}{0.23} & $-$0.07 \ci{$-$0.13}{$-$0.01} \\
Qwen2.5-VL-7B, IoU-GRPO & 38 & +0.15 \ci{0.09}{0.23} & +0.23 \ci{0.16}{0.30} & $-$0.08 \ci{$-$0.17}{0.02} \\
Qwen2.5-VL-7B, placebo-controlled $\alpha{=}0.25$ & 11 & +2.22 \ci{1.63}{2.83} & +0.23 \ci{0.16}{0.30} & +1.99 \ci{1.41}{2.61} \\
Qwen2.5-VL-7B, placebo-controlled $\alpha{=}0.5$ & 14 & +1.75 \ci{1.11}{2.34} & +0.18 \ci{0.12}{0.23} & +1.57 \ci{0.93}{2.14} \\
Qwen2.5-VL-7B, placebo-controlled $\alpha{=}1$ & 26 & +1.16 \ci{0.84}{1.55} & +0.23 \ci{0.16}{0.30} & +0.93 \ci{0.60}{1.32} \\
Qwen2.5-VL-7B, placebo-controlled $\alpha{=}1$ (second seed) & 28 & +1.24 \ci{0.87}{1.66} & +0.23 \ci{0.16}{0.30} & +1.01 \ci{0.63}{1.44} \\
TimeZero (public) & 39 & +0.26 \ci{0.19}{0.32} & +0.23 \ci{0.16}{0.30} & +0.03 \ci{$-$0.08}{0.12} \\
Time-R1 (public) & 35 & +0.22 \ci{0.15}{0.30} & +0.23 \ci{0.16}{0.30} & $-$0.01 \ci{$-$0.11}{0.10} \\
\bottomrule
\end{tabular}
\end{table}

\subsection{Answering blind}
\label{app:black}
Table~\ref{tab:s3} gives the all-black readout by configuration, dataset and template. Under the VideoChat-R1 and TimeLens-8B templates no model refuses a black video. On QVHighlights the Qwen base answers no black video, while TimeZero answers $94.3\%$ of them and its placebo-controlled model $16.7\%$ (Figure~\ref{fig:ood}b).

\begin{table}[!htbp]
\centering\scriptsize\setlength{\tabcolsep}{3pt}\renewcommand{\arraystretch}{1.05}
\caption{\textbf{On an all-black video, SFT and the public checkpoints keep answering.} The placebo-controlled Qwen2.5-VL-7B models almost never answer, and those trained from TimeZero and Time-R1 answer less often than their bases. Every frame is set to zero; duration and video token count are kept. The column \emph{data} names the question set: Charades-STA (STA), QVHighlights (QVH) or Charades-STA with TimeLens labels (C-TL). \emph{ans.} is the fraction of questions still given an interval, and \emph{black \Rfive} is the \Rfive{} over all questions. \emph{abst.} is the strict (s) or wide (w) abstention rate: the strict criterion counts only the verbatim output \texttt{There are none.}, and the wide criterion counts any refusal wording (Appendix~\ref{app:protocols}). Text-first rows put the query before the video.}
\label{tab:s3}
\begin{tabular}{@{}>{\raggedright\arraybackslash}p{5.8cm}ccccc@{}}
\toprule
Model / objective & data & ans. & black \Rfive & abst.\ (s) & abst.\ (w) \\
\midrule
\addlinespace[1pt]\multicolumn{6}{@{}l}{\textit{Charades-STA}} \\
Qwen2.5-VL-7B, base, text-first & STA & 0.578 & 0.092 & 0.289 & 0.290 \\
Qwen2.5-VL-7B, base & STA & 0.116 & 0.007 & 0.348 & 0.878 \\
Qwen2.5-VL-7B, placebo-controlled $\alpha{=}0.5$, $\beta{=}0.03$ & STA & 0.002 & 0.000 & 0.498 & 0.994 \\
Qwen2.5-VL-7B, placebo-controlled $\alpha{=}0.5$ & STA & 0.000 & 0.000 & 0.065 & 0.982 \\
Qwen2.5-VL-7B, placebo-controlled $\alpha{=}1$ & STA & 0.001 & 0.000 & 0.051 & 0.994 \\
Qwen2.5-VL-7B, placebo-controlled $\alpha{=}0.5$, 8,434 pairs, trained on 16 frames & STA & 0.001 & 0.000 & 0.519 & 0.998 \\
Qwen2.5-VL-7B, uniform weights & STA & 0.000 & 0.000 & 0.042 & 0.993 \\
Qwen2.5-VL-7B, no placebo bound & STA & 0.000 & 0.000 & 0.029 & 0.994 \\
Qwen2.5-VL-7B, no bound, uniform & STA & 0.000 & 0.000 & 0.025 & 0.990 \\
Qwen2.5-VL-7B, IoU-GRPO & STA & 0.324 & 0.037 & 0.301 & 0.669 \\
Qwen2.5-VL-7B, mDPO-style & STA & 0.025 & 0.002 & 0.733 & 0.970 \\
Qwen2.5-VL-7B, SFT & STA & 0.784 & 0.140 & 0.200 & 0.216 \\
TimeZero, base, text-first & STA & 0.940 & 0.189 & 0.000 & 0.000 \\
TimeZero, base & STA & 0.968 & 0.136 & 0.004 & 0.307 \\
TimeZero, placebo-controlled $\alpha{=}0.25$ & STA & 0.474 & 0.085 & 0.180 & 0.782 \\
TimeZero, placebo-controlled $\alpha{=}0.5$, $\beta{=}0.03$ & STA & 0.530 & 0.098 & 0.127 & 0.814 \\
TimeZero, placebo-controlled $\alpha{=}0.5$, text-first & STA & 0.706 & 0.172 & 0.077 & 0.709 \\
TimeZero, placebo-controlled $\alpha{=}0.5$, trained text-first, evaluated video-first & STA & 0.591 & 0.110 & 0.074 & 0.846 \\
TimeZero, placebo-controlled $\alpha{=}0.5$, trained video-first, evaluated text-first & STA & 0.551 & 0.109 & 0.119 & 0.656 \\
TimeZero, placebo-controlled $\alpha{=}0.5$ & STA & 0.303 & 0.052 & 0.312 & 0.844 \\
TimeZero, placebo-controlled $\alpha{=}1$, text-first & STA & 0.933 & 0.233 & 0.000 & 0.000 \\
TimeZero, placebo-controlled $\alpha{=}1$ & STA & 0.439 & 0.085 & 0.178 & 0.795 \\
TimeZero, SFT, text-first & STA & 0.938 & 0.253 & 0.000 & 0.000 \\
TimeZero, two-stage (SFT, then placebo-controlled), text-first & STA & 0.924 & 0.230 & 0.000 & 0.001 \\
Time-R1, base, text-first & STA & 0.762 & 0.131 & 0.086 & 0.087 \\
Time-R1, base & STA & 0.796 & 0.105 & 0.123 & 0.272 \\
Time-R1, placebo-controlled $\alpha{=}0.1$ & STA & 0.495 & 0.062 & 0.230 & 0.513 \\
Time-R1, placebo-controlled $\alpha{=}0.25$ & STA & 0.049 & 0.011 & 0.178 & 0.959 \\
Time-R1, placebo-controlled $\alpha{=}0.5$ & STA & 0.487 & 0.042 & 0.214 & 0.547 \\
Time-R1, placebo-controlled $\alpha{=}1$ & STA & 0.184 & 0.010 & 0.463 & 0.807 \\
VideoChat-R1, base, text-first & STA & 0.948 & 0.211 & 0.000 & 0.000 \\
VideoChat-R1, base & STA & 1.00 & 0.153 & 0.000 & 0.000 \\
VideoChat-R1, placebo-controlled $\alpha{=}0.5$ & STA & 1.00 & 0.120 & 0.000 & 0.000 \\
VideoChat-R1, placebo-controlled $\alpha{=}1$ & STA & 1.00 & 0.092 & 0.000 & 0.000 \\
VideoChat-R1, SFT, text-first & STA & 0.961 & 0.262 & 0.000 & 0.000 \\
VideoChat-R1, two-stage (SFT, then placebo-controlled), text-first & STA & 0.941 & 0.239 & 0.000 & 0.000 \\
TimeLens-8B, base & STA & 1.00 & 0.152 & 0.000 & 0.000 \\
TimeLens-8B, placebo-controlled $\alpha{=}0.5$ & STA & 1.00 & 0.179 & 0.000 & 0.000 \\
\addlinespace[1pt]\multicolumn{6}{@{}l}{\textit{QVHighlights}} \\
Qwen2.5-VL-7B, base & QVH & 0.000 & 0.000 & 0.298 & 0.984 \\
Qwen2.5-VL-7B, placebo-controlled $\alpha{=}0.5$ & QVH & 0.003 & 0.000 & 0.174 & 0.987 \\
TimeZero, base & QVH & 0.943 & 0.013 & 0.008 & 0.719 \\
TimeZero, placebo-controlled $\alpha{=}0.5$ & QVH & 0.167 & 0.006 & 0.432 & 0.968 \\
\addlinespace[1pt]\multicolumn{6}{@{}l}{\textit{Charades, TimeLens re-annotation}} \\
TimeLens-8B, base & C-TL & 1.00 & 0.134 & 0.000 & 0.000 \\
TimeLens-8B, placebo-controlled $\alpha{=}0.5$ & C-TL & 1.00 & 0.133 & 0.000 & 0.000 \\
\bottomrule
\end{tabular}
\end{table}

\subsection{Accuracy and the annotation}
\label{app:labels}
Figure~\ref{fig:findings}b--c draws on the re-annotation matrix of Table~\ref{tab:s5}, which scores the same predictions under each label set on 870 questions and measures $\Gamma$ against the re-annotated intervals. On these 870 questions TimeLens-8B's \Rfive{} differs from its base Qwen3-VL-8B's by $-19.9$ $[-23.5,-16.4]$ points under the original labels and by $+8.7$ $[5.4,12.1]$ under the TimeLens labels (paired by question). On all 3{,}363 re-annotated test questions the difference is $+8.9$ $[7.3,10.6]$ points. On the 870 questions, our placebo-controlled adapter of TimeLens-8B, trained on the original labels, does the reverse of TimeLens-8B: it leads Qwen3-VL-8B under the original labels and trails it under the TimeLens labels. The three public checkpoints lead the Qwen2.5-VL-7B base by 9 to 19 points under the original labels, and the spread of the nine 7B configurations shrinks from 20 points under the original labels to 6 under the TimeLens labels (Table~\ref{tab:s5}). Against the re-annotated intervals, TimeLens-8B's $\Gm$ relative to the same base is $+0.01$ $[-0.03,0.05]$ (Table~\ref{tab:s5}). Table~\ref{tab:s9} is the window-length substitution. The constant is the typical span of the label set, so the substitution measures how much accuracy one number about the annotation carries.

\begin{table}[!htbp]
\centering\scriptsize\setlength{\tabcolsep}{3pt}\renewcommand{\arraystretch}{1.05}
\caption{\textbf{A second annotation of the same videos reorders the configurations by accuracy.} Measured against it, every placebo-controlled model keeps a positive $\Gamma$ and no accuracy-oriented model rises beyond the audit's resolution. The upper block scores one run's predictions on the 870 questions present in both label sets under each set of labels, where STA denotes the original Charades-STA labels and TL the TimeLens labels. The lower block gives $\Gm$ and $\Gs$ in nats per token, with 95\% intervals, measured with the re-annotated intervals as the fixed answer, together with the number $n_\cap$ of questions both models localise and the fraction of them on which $\psi$ rises.}
\label{tab:s5}
\begin{tabular}{@{}>{\raggedright\arraybackslash}p{5.2cm}cccc@{}}
\toprule
Model / objective & \Rfive{}, STA labels & \Rfive{}, TL labels & & \\
\midrule
Qwen2.5-VL-7B, base & 0.525 & 0.369 &  &  \\
Qwen2.5-VL-7B, placebo-controlled $\alpha{=}0.5$ & 0.561 & 0.382 &  &  \\
Qwen2.5-VL-7B, SFT & 0.622 & 0.331 &  &  \\
TimeZero, base & 0.713 & 0.361 &  &  \\
TimeZero, placebo-controlled $\alpha{=}0.5$ & 0.726 & 0.380 &  &  \\
Time-R1, base & 0.616 & 0.340 &  &  \\
Time-R1, placebo-controlled $\alpha{=}0.25$ & 0.644 & 0.364 &  &  \\
VideoChat-R1, base & 0.711 & 0.378 &  &  \\
VideoChat-R1, placebo-controlled $\alpha{=}0.5$ & 0.697 & 0.394 &  &  \\
Qwen3-VL-8B, base of TimeLens-8B & 0.631 & 0.537 &  &  \\
TimeLens-8B (public checkpoint) & 0.432 & 0.624 &  &  \\
TimeLens-8B, placebo-controlled $\alpha{=}0.5$ & 0.670 & 0.429 &  &  \\
\midrule
\multicolumn{5}{@{}l}{\textit{Evidence dependence measured against the re-annotated intervals}} \\
Model / objective & $\Gm$ & $\Gs$ & $n_\cap$ & sign frac. \\
\midrule
Qwen2.5-VL-7B, placebo-controlled $\alpha{=}0.5$ & +1.21 \ci{0.92}{1.49} & +1.06 \ci{0.77}{1.37} & 57 & 0.912 \\
Qwen2.5-VL-7B, placebo-controlled $\alpha{=}1$ & +0.87 \ci{0.63}{1.10} & +1.04 \ci{0.77}{1.33} & 58 & 0.897 \\
Qwen2.5-VL-7B, SFT & $-$0.03 \ci{$-$0.08}{0.01} & +0.04 \ci{0.00}{0.08} & 55 & 0.509 \\
TimeZero, placebo-controlled $\alpha{=}0.5$ & +1.00 \ci{0.78}{1.28} & +0.95 \ci{0.74}{1.20} & 63 & 0.921 \\
Time-R1, placebo-controlled $\alpha{=}0.25$ & +1.43 \ci{1.19}{1.70} & +1.19 \ci{0.90}{1.52} & 62 & 0.984 \\
VideoChat-R1, placebo-controlled $\alpha{=}0.5$ & +0.98 \ci{0.86}{1.14} & +0.68 \ci{0.51}{0.84} & 57 & 0.982 \\
TimeLens-8B against Qwen3-VL-8B & +0.01 \ci{$-$0.03}{0.05} & +0.03 \ci{$-$0.01}{0.07} & 78 & 0.564 \\
TimeLens-8B, placebo-controlled $\alpha{=}0.5$ & +0.95 \ci{0.82}{1.08} & +0.64 \ci{0.53}{0.75} & 70 & 0.957 \\
TimeLens-8B, SFT & $-$0.21 \ci{$-$0.24}{$-$0.18} & $-$0.17 \ci{$-$0.21}{$-$0.13} & 75 & 0.053 \\
\bottomrule
\end{tabular}
\end{table}

\begin{table}[!htbp]
\centering\scriptsize\setlength{\tabcolsep}{4pt}\renewcommand{\arraystretch}{1.05}
\caption{\textbf{Keeping each prediction's centre and replacing only its length reproduces the largest accuracy gain in the audit matrix.} The predictions come unchanged from the Charades-STA run, and only each interval's length is replaced, either by a constant 8\,s or by the length of the annotated interval. The columns give \Rfive{} and mIoU, the mean IoU with the annotation, for the predictions as made (\emph{as pred.}) and after the substitution (\emph{subst.}).}
\label{tab:s9}
\begin{tabular}{@{}>{\raggedright\arraybackslash}p{4.4cm}>{\raggedright\arraybackslash}p{3.3cm}cccc@{}}
\toprule
Configuration & length substituted & \multicolumn{2}{c}{\Rfive{}} & \multicolumn{2}{c@{}}{mIoU} \\
\cmidrule(lr){3-4}\cmidrule(l){5-6}
 & & as pred. & subst. & as pred. & subst. \\
\midrule
TimeLens-8B, placebo-controlled $\alpha{=}0.5$ & constant 8\,s & 0.626 & 0.649 & 0.540 & 0.519 \\
TimeLens-8B, placebo-controlled $\alpha{=}0.5$ & length of the annotated interval & 0.626 & 0.644 & 0.540 & 0.562 \\
TimeLens-8B, base & constant 8\,s & 0.417 & 0.678 & 0.434 & 0.542 \\
TimeLens-8B, base & length of the annotated interval & 0.417 & 0.685 & 0.434 & 0.593 \\
\bottomrule
\end{tabular}
\end{table}

\section{Training evidence dependence}
\label{app:training}
\subsection{The objective, term by term}
\label{app:terms}
\begin{figure}[b]
\centering
\includegraphics[width=\linewidth]{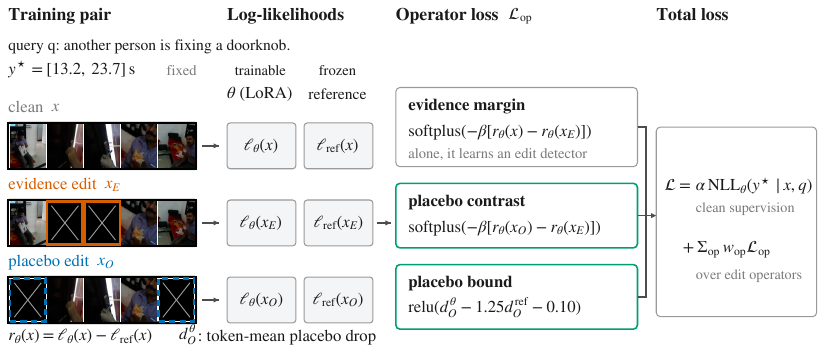}
\caption{\textbf{The placebo-controlled objective scores each training pair on the clean video and on both edits.} The trainable model and the frozen base score the fixed annotated answer on the clean video $x$, the evidence edit $x_E$ and the placebo edit $x_O$. Their log-likelihood differences $r_\theta$ enter the evidence margin and the placebo contrast, the placebo drops of both models enter the placebo bound, and the total loss adds clean supervision to the operator losses (Equations~\ref{eq:objective}--\ref{eq:total}).}
\label{fig:objective_schematic}
\end{figure}

Figure~\ref{fig:objective_schematic} traces one training pair through Equations~\ref{eq:objective}--\ref{eq:total}.

\paragraph{Components per operator.}
Table~\ref{tab:s7} decomposes the paired estimates of the trained models into $\dE$ and $\dO$ for trained model and base, per operator. The full objective moves $\dE$ and leaves $\dO$ at the base model's level; removing the bound, or the bound and the placebo contrast together, moves $\dO$ instead. The Group-DRO weights cut $\dO$ by a quarter to a third when the bound is removed. With the bound, the two weightings differ by $0.01$ in $\dO$ in the first seed: it is $0.02$ with uniform weights (Table~\ref{tab:objectives}) and $0.01$ with Group-DRO (Table~\ref{tab:audit}). In a further pair of runs sharing one seed, Group-DRO (the run of Figure~\ref{fig:dynamics}) keeps the placebo drop lower than uniform weights do, at $0.05$ against $0.12$ nats per token, with the base at $-0.07$. In that pair the Group-DRO run reaches $\Gm$ $+1.69$ and the uniform run $+1.97$ $[1.77,2.19]$, with gains of $5.8$ $[4.5,7.0]$ and $6.5$ $[5.1,7.8]$ points; audited against each other, the two runs give a $\Gm$ of $-0.31$ $[-0.38,-0.24]$.

\begin{table}[tbp]
\centering\scriptsize\setlength{\tabcolsep}{3pt}\renewcommand{\arraystretch}{1.05}
\caption{\textbf{Per edit operator, the placebo-controlled objective moves the evidence drop and keeps the placebo drop within $0.2$ nats per token of the base model's; without the placebo bound the placebo drop rises.} For the mask and the swap operator, the columns give the evidence drop $\dE$ and the placebo drop $\dO$ of the trained model (tr) and of its base on the same questions, in nats per token, followed by $\Gamma$; under mask they also give the sign fraction, the fraction of questions on which $\psi$ rises. Text-first rows put the query before the video.}
\label{tab:s7}
\begin{tabular}{@{}>{\raggedright\arraybackslash}p{4.1cm}cccccc@{\hspace{8pt}}ccccc@{}}
\toprule
 & \multicolumn{6}{c}{mask} & \multicolumn{5}{c}{swap} \\
\cmidrule(lr){2-7}\cmidrule(l){8-12}
Model / objective & $\dE^{\text{tr}}$ & $\dO^{\text{tr}}$ & $\dE^{\text{base}}$ & $\dO^{\text{base}}$ & $\Gamma$ & sign frac. & $\dE^{\text{tr}}$ & $\dO^{\text{tr}}$ & $\dE^{\text{base}}$ & $\dO^{\text{base}}$ & $\Gamma$ \\
\midrule
\addlinespace[1pt]\multicolumn{12}{@{}l}{\textit{Qwen2.5-VL-7B}} \\
placebo-controlled $\alpha{=}0.25$ & 2.29 & 0.01 & 0.18 & $-$0.07 & +2.03 & 1.00 & 1.98 & 0.03 & 0.05 & $-$0.06 & +1.84 \\
placebo-controlled $\alpha{=}0.5$, $\beta{=}0.03$ & 1.63 & 0.02 & 0.18 & $-$0.07 & +1.36 & 0.730 & 1.31 & 0.01 & 0.07 & $-$0.07 & +1.16 \\
placebo-controlled $\alpha{=}0.5$ & 1.82 & 0.01 & 0.12 & $-$0.06 & +1.63 & 0.973 & 1.55 & 0.04 & 0.06 & $-$0.04 & +1.41 \\
placebo-controlled $\alpha{=}1$, $\beta{=}0.03$ & 1.00 & $-$0.01 & 0.13 & $-$0.06 & +0.82 & 0.829 & 0.82 & 0.00 & 0.06 & $-$0.04 & +0.72 \\
placebo-controlled $\alpha{=}1$ & 1.28 & 0.01 & 0.17 & $-$0.06 & +1.03 & 0.804 & 1.17 & 0.04 & 0.07 & $-$0.06 & +1.00 \\
placebo-controlled $\alpha{=}0.5$, 4,800 pairs & 3.13 & 0.11 & 0.18 & $-$0.07 & +2.78 & 1.00 & 2.81 & 0.12 & 0.04 & $-$0.06 & +2.58 \\
placebo-controlled $\alpha{=}0.5$, 8,434 pairs, trained on 16 frames & 0.37 & $-$0.12 & 0.19 & $-$0.06 & +0.23 & 0.831 & 0.26 & $-$0.10 & 0.05 & $-$0.06 & +0.24 \\
placebo-controlled $\alpha{=}0.5$, 2,400 pairs, trained on 16 frames & 0.23 & $-$0.07 & 0.18 & $-$0.07 & +0.06 & 0.689 & 0.12 & $-$0.06 & 0.07 & $-$0.06 & +0.04 \\
uniform weights & 2.25 & 0.02 & 0.18 & $-$0.07 & +1.99 & 1.00 & 2.06 & 0.06 & 0.04 & $-$0.06 & +1.89 \\
no placebo bound & 2.80 & 0.49 & 0.19 & $-$0.07 & +2.05 & 1.00 & 2.77 & 0.30 & 0.06 & $-$0.07 & +2.34 \\
no bound, uniform & 3.52 & 0.66 & 0.18 & $-$0.07 & +2.62 & 1.00 & 3.17 & 0.45 & 0.05 & $-$0.07 & +2.60 \\
IoU-GRPO & 0.20 & $-$0.05 & 0.18 & $-$0.07 & 0.00 & 0.490 & 0.05 & $-$0.07 & 0.04 & $-$0.06 & +0.02 \\
mDPO-style & 0.16 & 0.00 & 0.12 & $-$0.06 & $-$0.03 & 0.500 & 0.11 & 0.00 & 0.06 & $-$0.04 & +0.01 \\
SFT & 0.10 & 0.00 & 0.12 & $-$0.07 & $-$0.09 & 0.402 & 0.07 & 0.00 & 0.06 & $-$0.04 & $-$0.03 \\
evidence margin only & 2.44 & 1.65 & 0.18 & $-$0.07 & +0.54 & 0.740 & 2.59 & 1.39 & 0.05 & $-$0.07 & +1.09 \\
two-stage (SFT, then placebo-controlled), $\alpha{=}0.5$ & 2.38 & 0.04 & 0.19 & $-$0.06 & +2.10 & 1.00 & 1.96 & 0.06 & 0.06 & $-$0.07 & +1.77 \\
two-stage (SFT, then placebo-controlled), $\beta{=}0.03$ & 2.99 & $-$0.01 & 0.19 & $-$0.06 & +2.75 & 0.977 & 2.27 & 0.07 & 0.06 & $-$0.06 & +2.09 \\
\addlinespace[1pt]\multicolumn{12}{@{}l}{\textit{TimeZero}} \\
placebo-controlled $\alpha{=}0.25$ & 2.22 & 0.04 & 0.16 & $-$0.02 & +2.00 & 0.992 & 1.76 & 0.03 & 0.08 & 0.00 & +1.66 \\
placebo-controlled $\alpha{=}0.5$, $\beta{=}0.03$ & 0.50 & 0.00 & 0.16 & $-$0.02 & +0.32 & 0.908 & 0.33 & 0.02 & 0.09 & 0.00 & +0.23 \\
placebo-controlled $\alpha{=}0.5$, text-first & 1.45 & 0.01 & 0.14 & $-$0.01 & +1.29 & 0.979 & 1.25 & 0.03 & 0.12 & 0.03 & +1.13 \\
placebo-controlled $\alpha{=}0.5$ & 1.61 & 0.03 & 0.16 & $-$0.02 & +1.40 & 0.985 & 1.45 & 0.06 & 0.08 & 0.00 & +1.31 \\
placebo-controlled $\alpha{=}1$ & 1.24 & $-$0.01 & 0.15 & $-$0.02 & +1.07 & 0.931 & 1.03 & 0.03 & 0.09 & 0.00 & +0.92 \\
\addlinespace[1pt]\multicolumn{12}{@{}l}{\textit{Time-R1}} \\
placebo-controlled $\alpha{=}0.1$ & 2.49 & $-$0.18 & 0.18 & $-$0.08 & +2.40 & 1.00 & 2.19 & $-$0.16 & 0.05 & $-$0.09 & +2.20 \\
placebo-controlled $\alpha{=}0.25$ & 2.07 & $-$0.02 & 0.14 & $-$0.07 & +1.88 & 1.00 & 1.82 & 0.04 & 0.08 & $-$0.04 & +1.66 \\
placebo-controlled $\alpha{=}0.5$ & 1.72 & 0.00 & 0.19 & $-$0.08 & +1.46 & 0.976 & 1.34 & 0.00 & 0.05 & $-$0.09 & +1.20 \\
placebo-controlled $\alpha{=}1$ & 1.22 & 0.02 & 0.18 & $-$0.08 & +0.95 & 0.863 & 1.15 & 0.02 & 0.05 & $-$0.09 & +0.99 \\
\addlinespace[1pt]\multicolumn{12}{@{}l}{\textit{VideoChat-R1}} \\
placebo-controlled $\alpha{=}0.25$ & 1.78 & $-$0.02 & 0.18 & 0.00 & +1.63 & 0.993 & 1.08 & 0.07 & 0.11 & 0.01 & +0.92 \\
placebo-controlled $\alpha{=}0.5$ & 1.60 & $-$0.01 & 0.18 & 0.00 & +1.43 & 1.00 & 1.01 & 0.01 & 0.12 & 0.01 & +0.89 \\
placebo-controlled $\alpha{=}1$ & 1.39 & 0.00 & 0.19 & 0.01 & +1.20 & 1.00 & 0.99 & 0.02 & 0.13 & 0.01 & +0.85 \\
\addlinespace[1pt]\multicolumn{12}{@{}l}{\textit{TimeLens-8B}} \\
public checkpoint against Qwen3-VL-8B & 0.07 & $-$0.02 & 0.22 & $-$0.03 & $-$0.16 & 0.294 & 0.01 & $-$0.03 & 0.15 & $-$0.03 & $-$0.14 \\
placebo-controlled $\alpha{=}0.25$ & 1.79 & $-$0.05 & 0.09 & $-$0.03 & +1.72 & 1.00 & 1.18 & $-$0.04 & 0.02 & $-$0.04 & +1.15 \\
placebo-controlled $\alpha{=}0.5$ & 1.43 & $-$0.03 & 0.08 & $-$0.03 & +1.36 & 1.00 & 0.90 & $-$0.03 & 0.02 & $-$0.04 & +0.88 \\
placebo-controlled $\alpha{=}1$ & 1.00 & $-$0.03 & 0.08 & $-$0.03 & +0.92 & 1.00 & 0.70 & $-$0.02 & 0.02 & $-$0.04 & +0.66 \\
\bottomrule
\end{tabular}
\end{table}

\subsection{Where the placebo bound comes from}
\label{app:prereg}
The form and constants of the bound in Equation~\ref{eq:objective} come from an earlier experiment in which every model saw 16 frames at 224\,px under one prompt (the \emph{16-frame setup}). In it, a Qwen2.5-VL-7B model trained with the two margins of Equation~\ref{eq:objective} and uniform weights but no placebo bound raised $\Gamma$ (Holm-adjusted $p\le0.0004$ over the experiment's four tests) while its placebo drop rose from $0.01$ to $3.22$ nats per token, far above $1.25$ times the base drop plus $0.10$, and its \Rfive{} fell from $0.260$ to $0.203$. Transfers of the same objective to the three public checkpoints repeated the pattern. Each raised $\Gamma$ while its placebo drop rose to $0.46$ (TimeZero), $4.37$ (VideoChat-R1) and $2.40$ (Time-R1) nats per token against limits near $0.1$, and their accuracy changed by $+2.5$, $+7.2$ and $+12.5$ points. This limit became the bound of the objective. In the same setup an IoU-reward DPO model shifted the likelihood scale itself, its clean-video NLL rising from $0.72$ to $2.27$; the objective therefore measures the evidence and placebo terms against the frozen base.

\subsection{Clean-supervision weight, base and data}
\label{app:budget}
\paragraph{Dose.}
Table~\ref{tab:families} gives one configuration per base model, and Table~\ref{tab:s2} the $\alpha$ and $\beta$ grids behind Figure~\ref{fig:dose}. Within every base, $\Gm$ falls with each step in $\alpha$; from the smallest to the largest weight it drops by $0.43$ (VideoChat-R1) to $1.45$ (Time-R1). The sign fraction is at least $0.73$ for every placebo-controlled model trained at 2 frames per second, and at least $0.97$ among those at $\beta{=}0.1$ and $\alpha\le0.5$. The two models trained on 16 frames (Training frames, below) have sign fractions of $0.69$ and $0.83$.

\begin{table}[!htbp]
\centering\scriptsize\setlength{\tabcolsep}{3pt}\renewcommand{\arraystretch}{1.05}
\caption{\textbf{Within every base, evidence dependence falls monotonically as the clean-supervision weight rises.} Each row compares one trained model with its base. $\Delta$\Rfive{} is the change in points, and $^\dagger$ marks a difference of the two accuracies that is not paired by question; $\Gm$ and $\Gs$ are in nats per token on the $n_\cap$ questions both models localise, and the sign fraction is the fraction of them on which $\psi$ rises. Rows other than SFT use $\beta{=}0.1$ unless they name another value, and brackets give 95\% intervals.}
\label{tab:s2}
\begin{tabular}{@{}>{\raggedright\arraybackslash}p{3.6cm}ccccc@{}}
\toprule
Setting & $\Delta$\Rfive{} & $\Gm$ & $\Gs$ & $n_\cap$ & sign frac. \\
\midrule
\addlinespace[1pt]\multicolumn{6}{@{}l}{\textit{Qwen2.5-VL-7B}} \\
$\alpha{=}0.25$ & +2.1 \ci{0.8}{3.2} & +2.03 \ci{1.76}{2.29} & +1.84 \ci{1.58}{2.10} & 101 & 1.00 \\
$\alpha{=}0.5$ & +4.4 \ci{3.1}{5.7} & +1.63 \ci{1.36}{1.89} & +1.41 \ci{1.18}{1.63} & 112 & 0.973 \\
$\alpha{=}1$ & +6.7 \ci{5.2}{8.2} & +1.03 \ci{0.80}{1.27} & +1.00 \ci{0.79}{1.23} & 102 & 0.804 \\
$\alpha{=}0.5$, $\beta{=}0.03$ & +1.2 \ci{$-$0.6}{2.9} & +1.36 \ci{0.99}{1.72} & +1.16 \ci{0.81}{1.50} & 89 & 0.730 \\
SFT & +9.4 \ci{7.9}{11.0} & $-$0.09 \ci{$-$0.15}{$-$0.02} & $-$0.03 \ci{$-$0.08}{0.01} & 112 & 0.402 \\
$\alpha{=}0.5$, 4,800 pairs & +8.1 \ci{6.7}{9.5} & +2.78 \ci{2.54}{3.03} & +2.58 \ci{2.28}{2.86} & 105 & 1.00 \\
\addlinespace[1pt]\multicolumn{6}{@{}l}{\textit{TimeZero}} \\
$\alpha{=}0.25$ & +0.2$^\dagger$ & +2.00 \ci{1.73}{2.26} & +1.66 \ci{1.40}{1.91} & 133 & 0.992 \\
$\alpha{=}0.5$ & +0.5 \ci{$-$0.7}{1.5} & +1.40 \ci{1.19}{1.60} & +1.31 \ci{1.11}{1.50} & 135 & 0.985 \\
$\alpha{=}1$ & $-$0.6$^\dagger$ & +1.07 \ci{0.89}{1.25} & +0.92 \ci{0.75}{1.08} & 130 & 0.931 \\
$\alpha{=}0.5$, $\beta{=}0.03$ & $-$1.0 \ci{$-$2.1}{0.1} & +0.32 \ci{0.26}{0.39} & +0.23 \ci{0.18}{0.27} & 131 & 0.908 \\
\addlinespace[1pt]\multicolumn{6}{@{}l}{\textit{Time-R1}} \\
$\alpha{=}0.1$ & +0.3$^\dagger$ & +2.40 \ci{2.12}{2.67} & +2.20 \ci{1.88}{2.50} & 130 & 1.00 \\
$\alpha{=}0.25$ & +1.6$^\dagger$ & +1.88 \ci{1.64}{2.09} & +1.66 \ci{1.42}{1.87} & 131 & 1.00 \\
$\alpha{=}0.5$ & +0.9 \ci{$-$0.2}{2.0} & +1.46 \ci{1.23}{1.68} & +1.20 \ci{1.00}{1.38} & 127 & 0.976 \\
$\alpha{=}1$ & $-$8.4$^\dagger$ & +0.95 \ci{0.72}{1.19} & +0.99 \ci{0.75}{1.21} & 102 & 0.863 \\
\addlinespace[1pt]\multicolumn{6}{@{}l}{\textit{VideoChat-R1}} \\
$\alpha{=}0.25$ & $-$3.2 \ci{$-$4.6}{$-$1.8} & +1.63 \ci{1.46}{1.79} & +0.92 \ci{0.75}{1.08} & 136 & 0.993 \\
$\alpha{=}0.5$ & $-$2.7 \ci{$-$4.2}{$-$1.2} & +1.43 \ci{1.27}{1.58} & +0.89 \ci{0.74}{1.04} & 132 & 1.00 \\
$\alpha{=}1$ & $-$2.4 \ci{$-$3.8}{$-$0.9} & +1.20 \ci{1.05}{1.34} & +0.85 \ci{0.72}{1.00} & 129 & 1.00 \\
\addlinespace[1pt]\multicolumn{6}{@{}l}{\textit{TimeLens-8B}} \\
$\alpha{=}0.25$ & +17.9 \ci{15.9}{20.0} & +1.72 \ci{1.60}{1.85} & +1.15 \ci{0.96}{1.32} & 84 & 1.00 \\
$\alpha{=}0.5$ & +20.9 \ci{18.9}{22.8} & +1.36 \ci{1.24}{1.47} & +0.88 \ci{0.76}{1.01} & 85 & 1.00 \\
$\alpha{=}1$ & +23.2 \ci{21.4}{25.1} & +0.92 \ci{0.83}{1.01} & +0.66 \ci{0.57}{0.76} & 87 & 1.00 \\
\bottomrule
\end{tabular}
\end{table}

\paragraph{Budget.} The Qwen2.5-VL-7B model trained on 4{,}800 pairs raises $\psi$ on every question of its $H_\cap$ (sign fraction $1.00$; Table~\ref{tab:s2}). Its placebo drop is $0.11$ nats per token, against $-0.07$ for its base (Table~\ref{tab:s7}).

\begin{figure}[t]
\centering
\includegraphics[width=\linewidth]{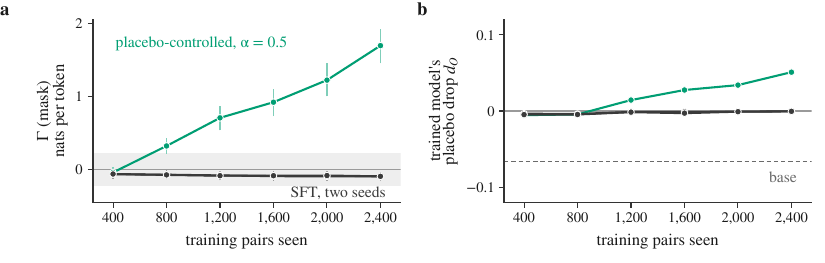}
\caption{\textbf{Evidence dependence grows throughout training; SFT's stays within the audit's resolution from the start.} Checkpoints are taken every 50 steps (400 pairs) of one placebo-controlled run ($\alpha{=}0.5$, $\beta{=}0.1$) and of two SFT runs from Qwen2.5-VL-7B, and each checkpoint is audited against the base. (a) The curves give $\Gamma$ (mask) with 95\% intervals, and the grey band marks the audit's resolution. (b) The curves give the trained model's placebo drop on the same questions, and the dashed line marks the base's.}
\label{fig:dynamics}
\end{figure}

\paragraph{Training dynamics.}
Figure~\ref{fig:dynamics} audits checkpoints of a further run of the placebo-controlled Qwen2.5-VL-7B model at $\alpha{=}0.5$ and of two SFT runs. $\Gamma$ rises throughout the single pass and has not levelled off at its end ($+1.69$ $[1.45,1.92]$ after 2{,}400 pairs), in line with the larger $\Gamma$ after 4{,}800 pairs (Section~\ref{sec:objective}); the trained model's placebo drop stays below $0.06$ nats per token. SFT's $\Gamma$ stays between $-0.09$ and $-0.06$ from the first checkpoint on.

\paragraph{Training frames.} Dependence is learned on the frames the model is audited on. Two models trained in the 16-frame setup (Appendix~\ref{app:prereg}) and audited at 2 frames per second reach a $\Gm$ of $+0.06$ when trained on 2{,}400 pairs and $+0.23$ when trained on all 8{,}434 pairs, the latter with an \Rfive{} $5.7$ points below its base; both are far below the models trained at 2 frames per second (Tables~\ref{tab:s7} and~\ref{tab:s1}).

\paragraph{Two-stage training.} Training Qwen2.5-VL-7B with SFT first and the objective second gives a $\Gm$ of $+2.10$ $[1.77, 2.41]$ at \Rfive{} $0.490$, and with $\beta{=}0.03$ it gives $+2.75$ $[2.29, 3.22]$ at $0.531$. The placebo drops of the two models under mask are $0.04$ and $-0.01$ nats per token.

\section{What the trained models do with the evidence}
\label{app:shift}
\subsection{Unseen edit operators}
\label{app:unseen}
Training uses five edit operators (Appendix~\ref{app:protocols}), among them the audit's mask and swap. Table~\ref{tab:unseen} audits three placebo-controlled Qwen2.5-VL-7B models and two TimeZero models under additive noise, patch scrambling and Gaussian blur, none of which occurs in training. Fourteen of the fifteen intervals exclude zero, and all six values of the Qwen2.5-VL-7B models under noise and Gaussian blur, $+0.07$ to $+0.16$, lie within the mask resolution. Appendix~\ref{app:add-operators} reads noise and blur against their own seed resolutions, measured on 600 further questions.

\begin{table}[!htbp]
\centering\scriptsize\setlength{\tabcolsep}{3pt}\renewcommand{\arraystretch}{1.05}
\caption{\textbf{Under edit operators never used in training, every placebo-controlled model keeps a positive $\Gamma$, large under patch scrambling and small under noise and Gaussian blur.} Each model is audited against its base on the same 200 questions. For each operator, the table gives $\Gamma$ with its 95\% interval on the $n_\cap$ questions both models localise, and the trained model's placebo drop $\dO$, both in nats per token.}
\label{tab:unseen}
\begin{tabular}{@{}lccccccc@{}}
\toprule
 & & \multicolumn{2}{c}{noise} & \multicolumn{2}{c}{patch scrambling} & \multicolumn{2}{c}{Gaussian blur} \\
\cmidrule(lr){3-4}\cmidrule(lr){5-6}\cmidrule(l){7-8}
Model / objective & $n_\cap$ & $\Gamma$ & $\dO$ & $\Gamma$ & $\dO$ & $\Gamma$ & $\dO$ \\
\midrule
\addlinespace[1pt]\multicolumn{8}{@{}l}{\textit{Qwen2.5-VL-7B}} \\
placebo-controlled $\alpha{=}0.5$ & 100 & +0.15 \ci{0.07}{0.26} & $-$0.02 & +1.52 \ci{1.23}{1.82} & 0.09 & +0.16 \ci{0.08}{0.25} & 0.00 \\
placebo-controlled $\alpha{=}0.5$, $\beta{=}0.03$ & 88 & +0.15 \ci{0.02}{0.30} & 0.00 & +1.34 \ci{1.00}{1.71} & 0.03 & +0.07 \ci{0.01}{0.16} & 0.00 \\
placebo-controlled $\alpha{=}1$ & 100 & +0.07 \ci{$-$0.01}{0.14} & $-$0.01 & +1.12 \ci{0.87}{1.38} & 0.04 & +0.08 \ci{0.02}{0.14} & 0.00 \\
\addlinespace[1pt]\multicolumn{8}{@{}l}{\textit{TimeZero}} \\
placebo-controlled $\alpha{=}0.25$ & 134 & +0.28 \ci{0.16}{0.40} & $-$0.03 & +1.79 \ci{1.55}{2.05} & 0.04 & +0.27 \ci{0.16}{0.39} & $-$0.01 \\
placebo-controlled $\alpha{=}0.5$ & 134 & +0.23 \ci{0.13}{0.35} & 0.00 & +1.29 \ci{1.11}{1.48} & 0.04 & +0.23 \ci{0.14}{0.34} & 0.00 \\
\bottomrule
\end{tabular}
\end{table}

\subsection{The partner event's query}
\label{app:qspec}
\paragraph{Definitions.}
Take a question $i$ of the 200 audit questions, with query $q_i$, annotated interval $y^\star_i$ and evidence frames $E_i$ (event A of Section~\ref{sec:readouts}). Its partner $j$ (event B) is another annotation of the same video with a different sentence, whose frames $E_j$ do not overlap $E_i$; among several, it is the one whose frames are centred farthest from $E_i$. Every quantity below scores the same answer $y^\star_i$ by teacher forcing, as the token-mean log-likelihood $\ell(x;q)=\log p(y^\star_i\mid x,q)$ under a query $q$, and uses the mask operator. With $x_{E_i}$ and $x_{E_j}$ the video with all frames of $E_i$ or all frames of $E_j$ masked, the \emph{partner contrast} under query $q$ is
\begin{equation}
\pi(q)=\big[\ell(x;q)-\ell(x_{E_i};q)\big]-\big[\ell(x;q)-\ell(x_{E_j};q)\big]=\ell(x_{E_j};q)-\ell(x_{E_i};q),
\end{equation}
which is positive when masking the question's own event costs more than masking the partner's. We write $\psi_{\mathrm{own}}=\pi(q_i)$ for the question's own query and $\psi_{\mathrm{cross}}=\pi(q_j)$ for the partner's. Both are partner contrasts, not the placebo difference $\psi$ of Section~\ref{sec:audit}. Table~\ref{tab:s11} gives their trained-minus-base means and the difference of the two,
\begin{equation}
\Delta\psi_{\mathrm{own}}=\frac{1}{n}\sum_{i}\big(\psi^{\mathrm{trained}}_{\mathrm{own},i}-\psi^{\mathrm{base}}_{\mathrm{own},i}\big),\qquad
\Gamma_{\mathrm{spec}}=\Delta\psi_{\mathrm{own}}-\Delta\psi_{\mathrm{cross}},
\end{equation}
with $\Delta\psi_{\mathrm{cross}}$ defined in the same way, over the $n$ questions that both models localise (IoU $\ge0.5$ on the clean video) and that have a partner. The sign fraction is the fraction of these questions on which the per-question term of $\Gamma_{\mathrm{spec}}$ is positive. A model that reads the query raises $\psi_{\mathrm{own}}$ and lowers $\psi_{\mathrm{cross}}$, so $\Delta\psi_{\mathrm{cross}}<0$ and $\Gamma_{\mathrm{spec}}>\Delta\psi_{\mathrm{own}}$; a model that responds only to where the edit lands changes both alike, so $\Gamma_{\mathrm{spec}}\approx0$. The right block of the table replaces the partner edit with the placebo. There $\psi(q)=\dE(q)-\dO(q)$ is the difference of Section~\ref{sec:audit} under query $q$, with the placebo frames drawn from outside both $E_i$ and $E_j$ so that the placebo never touches the partner event. $\Gamma$ is the trained-minus-base mean of $\psi(q_i)$, and $\Gamma_{\mathrm{agn}}$ is that of $\psi(q_j)$: the part of $\Gamma$ that remains when the query names another event. The rest, $\Gamma_{\mathrm{sem}}=\Gamma-\Gamma_{\mathrm{agn}}$, is the part that depends on the question's own query. $\Gamma_{\mathrm{spec}}$ and $\Gamma_{\mathrm{sem}}$ thus measure the query-dependent part against two different controls, the partner event's frames and the placebo frames. Figure~\ref{fig:readouts}a plots $\Delta\psi_{\mathrm{own}}$ (horizontal) against $\Delta\psi_{\mathrm{cross}}$ (vertical). Because of its placebo pool, this block leaves out the questions with fewer frames outside $E_i\cup E_j$ than in $E_i$, and its $\Gamma$ differs from the $\Gamma$ of the other tables.

\begin{figure}[t]
\centering
\includegraphics[width=\linewidth]{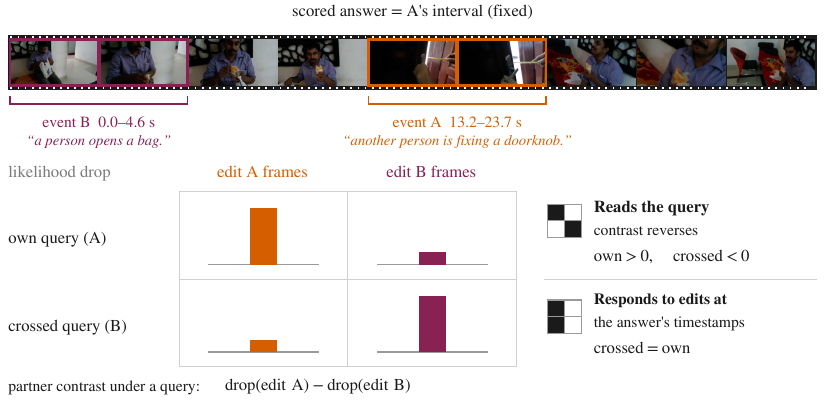}
\caption{\textbf{The query-crossed design scores the same answer under the question's own query and under a partner event's query.} Both conditions score the annotated interval of event A and use the same two edits: masking A's frames, or masking the frames of a partner event B of the same video. Only the query changes, from A's own to B's. A model that reads the query shows a contrast that reverses under B's query; a model that responds to edits at the answer's timestamps shows the same contrast under both queries. The bars are schematic; Figure~\ref{fig:readouts}a and Table~\ref{tab:s11} give the measurements.}
\label{fig:cross_design}
\end{figure}
\begin{table}[!htbp]
\centering\scriptsize
\caption{\textbf{The audited contrast moves with the query: $\Gamma_{\mathrm{spec}}$ is positive in every placebo-controlled model and zero in SFT.} All values use the mask operator and are in nats per token; Appendix~\ref{app:qspec} defines them. The left block uses the $n$ questions that both models localise and that have a partner event, another annotated event of the same video. The partner contrast scores the question's own interval and is the likelihood drop when its event's frames are masked minus the drop when the partner event's frames are masked. The left block gives the trained-minus-base change of this contrast under the question's own query ($\Delta\psi_{\mathrm{own}}$) and under the partner event's query ($\Delta\psi_{\mathrm{cross}}$), and $\Gamma_{\mathrm{spec}}=\Delta\psi_{\mathrm{own}}-\Delta\psi_{\mathrm{cross}}$ with the fraction of questions on which it is positive. The right block draws the placebo from outside both events, which leaves out 8 to 11 of these questions per row, and splits $\Gamma$ into the part that remains under the partner event's query ($\Gamma_{\mathrm{agn}}$) and the rest ($\Gamma_{\mathrm{sem}}=\Gamma-\Gamma_{\mathrm{agn}}$). Brackets give 95\% intervals, and text-first rows put the query before the video.}
\label{tab:s11}
\setlength{\tabcolsep}{2.2pt}\renewcommand{\arraystretch}{1.04}
\begin{tabular}{@{}lrcccc@{\hspace{7pt}}ccc@{}}
\toprule
 & & \multicolumn{4}{c}{Query crossed} & \multicolumn{3}{c@{}}{Placebo pool split} \\
\cmidrule(lr){3-6}\cmidrule(l){7-9}
 & $n$ & $\Delta\psi_{\mathrm{own}}$ & $\Delta\psi_{\mathrm{cross}}$ & $\Gamma_{\mathrm{spec}}$ & \shortstack{sign\\frac.} & $\Gamma$ & $\Gamma_{\mathrm{agn}}$ & $\Gamma_{\mathrm{sem}}$ \\
\midrule
\multicolumn{9}{@{}l}{\emph{Qwen2.5-VL-7B}} \\
\quad placebo-controlled $\alpha{=}0.25$ & 51 & $+1.75$ & $-1.31$ & $+3.07$ \ci{2.30}{3.75} & 1.00 & $+1.52$ & $+0.24$ \ci{0.12}{0.45} & $+1.29$ \\
\quad placebo-controlled $\alpha{=}0.5$ & 46 & $+1.59$ & $-1.27$ & $+2.86$ \ci{1.99}{3.63} & 0.96 & $+1.21$ & $+0.20$ \ci{0.04}{0.41} & $+1.01$ \\
\quad placebo-controlled $\alpha{=}0.5$, $\beta{=}0.03$ & 43 & $+1.34$ & $-1.49$ & $+2.83$ \ci{1.89}{3.70} & 0.91 & $+0.86$ & $0.00$ \ci{$-$0.12}{0.15} & $+0.87$ \\
\quad placebo-controlled $\alpha{=}1$ & 48 & $+1.00$ & $-1.15$ & $+2.15$ \ci{1.40}{2.73} & 0.92 & $+0.65$ & $-0.03$ \ci{$-$0.15}{0.11} & $+0.69$ \\
\quad placebo-controlled $\alpha{=}1$ (second seed) & 48 & $+1.06$ & $-1.04$ & $+2.10$ \ci{1.47}{2.62} & 0.98 & $+0.70$ & $-0.01$ \ci{$-$0.10}{0.12} & $+0.71$ \\
\quad two-stage, $\alpha{=}0.5$ & 42 & $+2.10$ & $-0.88$ & $+2.98$ \ci{2.30}{3.60} & 1.00 & $+1.74$ & $+0.19$ \ci{$-$0.01}{0.44} & $+1.56$ \\
\quad two-stage, $\beta{=}0.03$ & 44 & $+2.64$ & $-1.35$ & $+3.99$ \ci{3.04}{4.90} & 0.98 & $+2.18$ & $+0.28$ \ci{0.03}{0.57} & $+1.90$ \\
\quad evidence margin only & 49 & $+0.99$ & $-1.01$ & $+2.00$ \ci{1.22}{2.61} & 0.94 & $+0.46$ & $-0.35$ \ci{$-$0.76}{0.20} & $+0.81$ \\
\quad SFT & 46 & $-0.11$ & $-0.06$ & $-0.04$ \ci{$-$0.18}{0.09} & 0.59 & $-0.19$ & $-0.11$ \ci{$-$0.21}{$-$0.03} & $-0.07$ \\
\midrule
\multicolumn{9}{@{}l}{\emph{TimeZero}} \\
\quad placebo-controlled $\alpha{=}0.25$ & 59 & $+1.88$ & $-1.27$ & $+3.15$ \ci{2.40}{3.86} & 0.98 & $+1.70$ & $+0.25$ \ci{0.13}{0.41} & $+1.44$ \\
\quad placebo-controlled $\alpha{=}0.5$ & 59 & $+1.37$ & $-0.77$ & $+2.14$ \ci{1.64}{2.64} & 1.00 & $+1.16$ & $+0.13$ \ci{0.02}{0.29} & $+1.04$ \\
\quad placebo-controlled $\alpha{=}0.5$, text-first & 58 & $+0.56$ & $+0.03$ & $+0.54$ \ci{0.33}{0.71} & 0.88 & $+0.56$ & $+0.13$ \ci{0.06}{0.21} & $+0.43$ \\
\midrule
\multicolumn{9}{@{}l}{\emph{Time-R1}} \\
\quad placebo-controlled $\alpha{=}0.25$ & 56 & $+1.64$ & $-1.20$ & $+2.84$ \ci{2.21}{3.41} & 0.98 & $+1.41$ & $+0.20$ \ci{0.07}{0.35} & $+1.21$ \\
\midrule
\multicolumn{9}{@{}l}{\emph{VideoChat-R1}} \\
\quad placebo-controlled $\alpha{=}0.25$ & 58 & $+1.62$ & $+0.69$ & $+0.93$ \ci{0.61}{1.18} & 0.81 & $+1.65$ & $+0.49$ \ci{0.20}{0.77} & $+1.16$ \\
\quad placebo-controlled $\alpha{=}0.5$ & 56 & $+1.40$ & $+0.67$ & $+0.73$ \ci{0.47}{0.95} & 0.84 & $+1.42$ & $+0.49$ \ci{0.20}{0.76} & $+0.93$ \\
\quad placebo-controlled $\alpha{=}0.5$, text-first & 62 & $+0.93$ & $+0.35$ & $+0.58$ \ci{0.34}{0.80} & 0.79 & $+1.03$ & $+0.25$ \ci{0.14}{0.37} & $+0.78$ \\
\bottomrule
\end{tabular}
\end{table}

An edit inside the annotated interval changes two things: the frames the query describes, and the frames at the timestamps the answer names. Figure~\ref{fig:cross_design} and Table~\ref{tab:s11} separate them with the partner contrast, holding the scored answer $y^\star_i$ and the edited frames fixed and changing only the query. Three accounts predict different contrasts under $q_j$. A model that reads the query should now be hurt more by the partner edit, which reverses the contrast. A model that detects damage at the answer's timestamps gives the same contrast under both queries. A low likelihood under $q_j$ shrinks the contrast toward zero, but only reading the query reverses it. For every placebo-controlled Qwen and Time-R1 model with a crossed run, for video-first TimeZero and for the evidence-margin-only model, the crossed contrast reverses. In the three VideoChat-R1 models it shrinks but stays positive, and text-first TimeZero sits near zero. $\Gamma_{\mathrm{spec}}$ is positive in every placebo-controlled model ($+0.5$ to $+4.0$; sign fractions $0.79$--$1.00$) and zero for SFT ($-0.04$ $[-0.18, 0.09]$). The query-agnostic part $\Gamma_{\mathrm{agn}}$ is at most $0.28$ in absolute value for every placebo-controlled Qwen and TimeZero model, and about a quarter to a third of $\Gamma$ for VideoChat-R1, whose models respond to the edit's position as well as to the query.

\paragraph{Reworded queries.}
Re-asking each question through a second human-written description of the same interval retains the contrast ($0.98$ $[0.94, 1.01]$ for mask, $1.02$ $[0.96, 1.09]$ for swap, $n{=}132$; the interval of each original-against-paraphrase difference contains zero), while a different event's query removes it.

\subsection{A second dataset}
\label{app:qvh}
\begin{figure}[t]
\centering
\includegraphics[width=\linewidth]{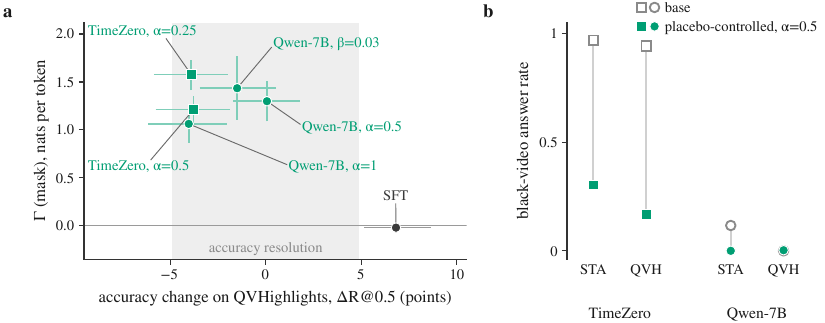}
\caption{\textbf{On QVHighlights, the placebo-controlled models keep a positive $\Gm$ while SFT stays at zero, and the placebo-controlled TimeZero model answers far fewer black videos than its base.} (a) Each Charades-trained model is placed by its paired change in \Rfive{} on QVHighlights (horizontal) and its $\Gm$ (vertical), with 95\% intervals; the grey band marks the out-of-domain accuracy resolution (Appendix~\ref{app:estimands}). \emph{Qwen-7B} is Qwen2.5-VL-7B. Green points are placebo-controlled models, trained at $\beta{=}0.1$ unless the label gives $\beta$; the $\beta{=}0.03$ model uses $\alpha{=}0.5$. (b) For each base model on Charades-STA (STA) and QVHighlights (QVH), the black-video answer rate of the base (open) is set against that of its placebo-controlled model at $\alpha{=}0.5$ (filled); squares are TimeZero and circles Qwen-7B.}
\label{fig:ood}
\end{figure}

\paragraph{QVHighlights.}
Table~\ref{tab:s8} and Figure~\ref{fig:ood} evaluate the Charades-trained configurations on QVHighlights, scored against the TimeLens re-annotation for both accuracy and the audit.

\begin{table}[!htbp]
\centering\scriptsize\setlength{\tabcolsep}{3pt}\renewcommand{\arraystretch}{1.05}
\caption{\textbf{On QVHighlights, every placebo-controlled model keeps a positive $\Gm$ and SFT stays at zero.} All models are trained on Charades-STA and evaluated on QVHighlights with the TimeLens re-annotation (1{,}541 questions). $\Delta$\Rfive{} is the paired change from the model's base in points, and $\Gm$ and $\Gs$ are in nats per token on the $n_\cap$ questions both models localise. The column \emph{ord.} gives the message order, video first (v) or query text first (t); \emph{parse} is the share of outputs that parse to an interval, and \emph{abst.} is the strict abstention rate, the share of outputs that are verbatim \texttt{There are none.}}
\label{tab:s8}
\begin{tabular}{@{}>{\raggedright\arraybackslash}p{4.4cm}cccccccc@{}}
\toprule
Model / objective & ord. & \Rfive & parse & abst. & $\Delta$\Rfive{} & $\Gm$ & $\Gs$ & $n_\cap$ \\
\midrule
\addlinespace[1pt]\multicolumn{9}{@{}l}{\textit{Qwen2.5-VL-7B}} \\
base & v & 0.319 & 0.992 & 0.008 & \NA & \NA & \NA & \NA \\
base, text-first & t & 0.097 & 0.925 & 0.075 & \NA & \NA & \NA & \NA \\
placebo-controlled $\alpha{=}0.5$ & v & 0.319 & 0.988 & 0.012 & +0.1 \ci{$-$1.7}{1.8} & +1.30 \ci{1.09}{1.51} & +1.03 & 43 \\
placebo-controlled $\alpha{=}1$ & v & 0.278 & 0.977 & 0.023 & $-$4.0 \ci{$-$6.1}{$-$2.0} & +1.06 \ci{0.87}{1.25} & +1.08 & 39 \\
placebo-controlled $\alpha{=}0.5$, $\beta{=}0.03$ & v & 0.304 & 0.698 & 0.302 & $-$1.5 \ci{$-$3.5}{0.5} & +1.44 \ci{1.11}{1.77} & +1.28 & 40 \\
SFT & v & 0.387 & 1.00 & 0.000 & +6.8 \ci{5.2}{8.6} & $-$0.02 \ci{$-$0.07}{0.03} & +0.02 & 52 \\
\addlinespace[1pt]\multicolumn{9}{@{}l}{\textit{TimeZero}} \\
base & v & 0.518 & 1.00 & 0.000 & \NA & \NA & \NA & \NA \\
base, text-first & t & 0.379 & 1.00 & 0.000 & \NA & \NA & \NA & \NA \\
placebo-controlled $\alpha{=}0.25$ & v & 0.479 & 1.00 & 0.000 & $-$3.9 \ci{$-$5.9}{$-$2.0} & +1.58 \ci{1.42}{1.73} & +1.52 & 82 \\
placebo-controlled $\alpha{=}0.5$ & v & 0.480 & 0.999 & 0.001 & $-$3.8 \ci{$-$5.8}{$-$1.9} & +1.21 \ci{1.07}{1.35} & +1.10 & 78 \\
\bottomrule
\end{tabular}
\end{table}

\paragraph{A re-split of Charades-STA.}
Charades-CD \citep{yuan2021closer} re-splits the same benchmark so that the test-ood moment distribution no longer matches training; the constant window falls from $0.294$ on test-iid to $0.150$ on test-ood (Table~\ref{tab:s14}). The test sets partly reuse the original training split, on which TimeZero and VideoChat-R1 were trained, so we compare each model's change from iid to ood. In each of the four bases that have an SFT model, SFT loses the most.

\begin{table}[!htbp]
\centering\papertablestyle
\caption{\textbf{Across the shift to the Charades-CD test-ood split, SFT loses more accuracy than the other objectives on Qwen2.5-VL-7B.} The table uses the test-iid (823 questions, 333 videos) and test-ood (3{,}375 questions, 1{,}442 videos) splits of Charades-CD \citep{yuan2021closer}. The re-split shifts the moment distribution, so the constant ten-second window falls from iid to ood; the column \emph{ood$-$iid} is each model's change across that shift, in points. On Qwen2.5-VL-7B, SFT's change is lower than IoU-GRPO's by $4.3$ $[0.8, 7.5]$ points and lower than the placebo-controlled model's by $5.5$ $[1.7, 9.1]$; questions are paired within each split, and the 95\% intervals resample videos. Text-first rows put the query before the video.}
\label{tab:s14}
\begin{tabular}{@{}lccc@{}}
\toprule
Model / objective & \Rfive{} iid & \Rfive{} ood & ood$-$iid \\
\midrule
constant $(0,10)$\,s window (no model) & 0.294 & 0.150 & $-$14.4 \\
\midrule
\multicolumn{4}{@{}l}{\emph{Qwen2.5-VL-7B}} \\
\quad base & 0.480 & 0.449 & $-$3.1 \\
\quad SFT & 0.622 & 0.524 & $-$9.8 \\
\quad IoU-GRPO & 0.635 & 0.580 & $-$5.5 \\
\quad placebo-controlled $\alpha{=}0.5$ & 0.498 & 0.455 & $-$4.3 \\
\addlinespace[2pt]
\multicolumn{4}{@{}l}{\emph{TimeZero}} \\
\quad base & 0.710 & 0.643 & $-$6.7 \\
\quad base, text-first & 0.744 & 0.663 & $-$8.1 \\
\quad SFT, text-first & 0.727 & 0.628 & $-$9.9 \\
\quad placebo-controlled $\alpha{=}0.5$ & 0.705 & 0.626 & $-$7.9 \\
\addlinespace[2pt]
\multicolumn{4}{@{}l}{\emph{Time-R1}} \\
\quad base & 0.612 & 0.591 & $-$2.2 \\
\quad placebo-controlled $\alpha{=}0.5$ & 0.628 & 0.581 & $-$4.7 \\
\quad placebo-controlled $\alpha{=}0.1$ & 0.618 & 0.585 & $-$3.3 \\
\addlinespace[2pt]
\multicolumn{4}{@{}l}{\emph{VideoChat-R1}} \\
\quad base & 0.708 & 0.640 & $-$6.8 \\
\quad base, text-first & 0.761 & 0.684 & $-$7.6 \\
\quad SFT, text-first & 0.759 & 0.659 & $-$10.0 \\
\quad placebo-controlled $\alpha{=}1$ & 0.689 & 0.596 & $-$9.3 \\
\addlinespace[2pt]
\multicolumn{4}{@{}l}{\emph{TimeLens-8B}} \\
\quad base & 0.413 & 0.423 & +1.0 \\
\quad SFT & 0.629 & 0.577 & $-$5.2 \\
\quad placebo-controlled $\alpha{=}0.5$ & 0.617 & 0.596 & $-$2.1 \\
\addlinespace[2pt]
\bottomrule
\end{tabular}
\end{table}

\subsection{Generated answers}
\label{app:regen}
\paragraph{Evidence share and follows.}
The evidence share is computed on the first 1{,}000 test questions. Each model regenerates its answer on the clean video, with the evidence frames edited, and with as many frames outside the annotated interval edited (one placebo draw per question, the same for every model), under mask and under swap. With $a_E$ and $a_O$ the \Rfive{} of the evidence-edited and the placebo-edited runs, the share is $1-a_E/a_O$, a ratio of two accuracies over all 1{,}000 questions rather than a mean over questions. Follows is computed on the likelihood audit's 200 questions. For each question a model localises, the block of evidence frames is exchanged with a block of as many consecutive frames that does not overlap it, placed at random, and the model regenerates its answer on the exchanged video. The answer follows the event if its IoU with the annotated interval, shifted by the offset between the two blocks and clipped to the video, is at least $0.5$; questions with no room for such a block are left out. For both readouts the tables report each trained model's difference from its base in points, the evidence share over all 1{,}000 questions and follows over $H_\cap$, with 95\% intervals from a bootstrap over videos.

\paragraph{Regenerated predictions.}
Table~\ref{tab:s6} gives the full readouts on generated answers behind Section~\ref{sec:readouts}, with the same 1{,}000 questions regenerated under clean, evidence-edited and placebo-edited versions of each operator. With as many frames masked outside the interval, every placebo-controlled model stays at or above its clean accuracy; masking the evidence frames removes most of it.

\begin{table}[!htbp]
\centering\scriptsize\setlength{\tabcolsep}{2.5pt}\renewcommand{\arraystretch}{1.05}
\caption{\textbf{In regenerated answers, every placebo-controlled model loses a larger share of its correct answers to evidence edits than its base does.} Under the mask, SFT and IoU-GRPO lose a smaller share. Each block compares a trained model with its base on the same 1{,}000 questions, edited frames and placebo draws. The first six columns give \Rfive{} on the clean video, with the evidence frames ($E$) or as many frames outside the interval ($O$) masked or swapped, and on all-black versions of these questions. The $\Delta$ columns give trained-minus-base changes in points in two readouts (Appendix~\ref{app:regen}). The evidence share is $1-a_E/a_O$, with $a_E$ and $a_O$ the \Rfive{} under the evidence edit and under the placebo edit, and is given under swap and under mask. \emph{Follows} is the fraction of the audit questions both models localise on which the prediction, regenerated after the event's frames are exchanged with another block, has IoU $\ge0.5$ with the moved interval; Table~\ref{tab:audit} gives it for the Qwen2.5-VL-7B models. Table~\ref{tab:s3} gives answering and refusal on black videos over the full test split. Brackets give 95\% intervals.}
\label{tab:s6}
\begin{tabular}{@{}lccccccccc@{}}
\toprule
 & clean & mask $E$ & mask $O$ & swap $E$ & swap $O$ & black & $\Delta$share swap & $\Delta$share mask & $\Delta$follows \\
\midrule
\multicolumn{10}{@{}l}{\textit{Qwen2.5-VL-7B, placebo-controlled $\alpha{=}0.5$ (video-first)}} \\
\quad base & 0.489 & 0.170 & 0.543 & 0.230 & 0.551 & 0.006 & \NA & \NA & \NA \\
\quad trained & 0.523 & 0.087 & 0.558 & 0.182 & 0.556 & 0.000 & +9.0 \ci{5.2}{12.9} & +15.7 \ci{11.4}{20.3} & \NA \\
\addlinespace[1pt]
\multicolumn{10}{@{}l}{\textit{Qwen2.5-VL-7B, IoU-GRPO (video-first)}} \\
\quad base & 0.489 & 0.170 & 0.543 & 0.230 & 0.551 & 0.006 & \NA & \NA & \NA \\
\quad trained & 0.637 & 0.250 & 0.668 & 0.278 & 0.685 & 0.047 & +1.2 \ci{$-$2.9}{5.5} & $-$6.1 \ci{$-$11.0}{$-$1.2} & \NA \\
\addlinespace[1pt]
\multicolumn{10}{@{}l}{\textit{Qwen2.5-VL-7B, mDPO-style (video-first)}} \\
\quad base & 0.489 & 0.170 & 0.543 & 0.230 & 0.551 & 0.006 & \NA & \NA & \NA \\
\quad trained & 0.596 & 0.190 & 0.630 & 0.217 & 0.629 & 0.005 & +7.2 \ci{3.7}{10.9} & +1.1 \ci{$-$3.3}{5.7} & \NA \\
\addlinespace[1pt]
\multicolumn{10}{@{}l}{\textit{Qwen2.5-VL-7B, SFT (video-first)}} \\
\quad base & 0.489 & 0.170 & 0.543 & 0.230 & 0.551 & 0.006 & \NA & \NA & \NA \\
\quad trained & 0.612 & 0.285 & 0.620 & 0.267 & 0.621 & 0.091 & $-$1.3 \ci{$-$5.3}{3.0} & $-$14.7 \ci{$-$19.8}{$-$9.4} & \NA \\
\addlinespace[1pt]
\multicolumn{10}{@{}l}{\textit{TimeZero, placebo-controlled $\alpha{=}0.5$, trained video-first, evaluated text-first}} \\
\quad base & 0.704 & 0.228 & 0.736 & 0.200 & 0.735 & 0.198 & \NA & \NA & \NA \\
\quad trained & 0.694 & 0.181 & 0.735 & 0.171 & 0.713 & 0.108 & +3.2 \ci{0.8}{5.8} & +6.4 \ci{3.3}{9.5} & +3.1 \ci{$-$3.9}{9.6} \\
\addlinespace[1pt]
\multicolumn{10}{@{}l}{\textit{TimeZero, placebo-controlled $\alpha{=}0.5$ (video-first)}} \\
\quad base & 0.700 & 0.218 & 0.709 & 0.262 & 0.721 & 0.143 & \NA & \NA & \NA \\
\quad trained & 0.690 & 0.145 & 0.726 & 0.213 & 0.724 & 0.049 & +6.9 \ci{3.9}{10.0} & +10.8 \ci{7.5}{14.2} & \NA \\
\addlinespace[1pt]
\multicolumn{10}{@{}l}{\textit{Time-R1, placebo-controlled $\alpha{=}0.25$ (video-first)}} \\
\quad base & 0.616 & 0.232 & 0.650 & 0.287 & 0.648 & 0.093 & \NA & \NA & \NA \\
\quad trained & 0.614 & 0.188 & 0.671 & 0.266 & 0.657 & 0.043 & +3.8 \ci{0.9}{6.8} & +7.7 \ci{4.8}{10.8} & $-$3.2 \ci{$-$10.4}{3.9} \\
\addlinespace[1pt]
\multicolumn{10}{@{}l}{\textit{VideoChat-R1, placebo-controlled $\alpha{=}0.5$, trained video-first, evaluated text-first}} \\
\quad base & 0.714 & 0.178 & 0.739 & 0.197 & 0.723 & 0.222 & \NA & \NA & \NA \\
\quad trained & 0.679 & 0.050 & 0.741 & 0.146 & 0.706 & 0.200 & +6.6 \ci{4.2}{9.1} & +17.3 \ci{13.7}{21.0} & $-$2.3 \ci{$-$10.4}{5.9} \\
\addlinespace[1pt]
\multicolumn{10}{@{}l}{\textit{VideoChat-R1, placebo-controlled $\alpha{=}0.5$ (video-first)}} \\
\quad base & 0.687 & 0.231 & 0.707 & 0.273 & 0.691 & 0.203 & \NA & \NA & \NA \\
\quad trained & 0.661 & 0.026 & 0.730 & 0.154 & 0.681 & 0.116 & +16.9 \ci{13.4}{20.3} & +29.1 \ci{24.8}{33.7} & \NA \\
\addlinespace[1pt]
\bottomrule
\end{tabular}
\end{table}

\paragraph{Public checkpoints.}
We also regenerated TimeZero and Time-R1 with their full reasoning, masking either the evidence or the placebo frames. For both, the difference in evidence share from the Qwen base has an interval that includes zero, while every placebo-controlled model of these bases raises the evidence share over its base (Tables~\ref{tab:audit} and~\ref{tab:s6}).

\subsection{Where the dependence sits}
\label{app:onset}
Table~\ref{tab:split} splits the answers lost to the evidence mask into refusals and wrong intervals, and $\Gamma$ over the onset, timestamp and tag tokens of the scored answer. Among the tagged templates, TimeZero's placebo-controlled model places the most $\Gamma$ on the timestamps, still far less than VideoChat-R1, whose template fixes the tags and leaves only the timestamps free. On the clean video the placebo-controlled models of the tagged templates write the onset token with probability $0.63$ to $0.96$. SFT writes it with probability $1.00$ and places no $\Gamma$ on it.

\begin{table}[tb]
\centering\footnotesize\setlength{\tabcolsep}{4pt}\renewcommand{\arraystretch}{1.1}
\caption{\textbf{Evidence dependence shows in whether a model answers and where it places the interval.} Each entry is a trained model minus its base, with the evidence frames masked. The left block starts from the regenerated answers that are correct under the placebo mask (1{,}000 questions) and gives the change in the share that returns no interval and, among the intervals returned, in the share that is wrong. The right block splits $\Gamma$ over the tokens of the scored answer, using a separate token-level audit of 200 questions; the \emph{onset} is the first token after the opening tag. Bold marks a change whose 95\% interval excludes zero.}
\label{tab:split}
\begin{tabular}{@{}lccccc@{}}
\toprule
 & \multicolumn{2}{c}{Lost answers become (points)} & \multicolumn{3}{c@{}}{Where $\Gamma$ sits (nats per token)} \\
\cmidrule(lr){2-3}\cmidrule(l){4-6}
Model & no interval & wrong interval & onset & timestamps & tags \\
\midrule
\multicolumn{6}{@{}l}{\emph{Trained for accuracy, against the Qwen2.5-VL-7B base}} \\
\quad SFT & \textbf{\boldmath $-$14.9$^\ast$} \ci{$-$18.5}{$-$11.6} & \textbf{\boldmath $-$8.7} \ci{$-$13.2}{$-$4.3} & $-$0.01 & $-$0.01 & $-$0.12 \\
\quad TimeZero (public) & \textbf{\boldmath $-$14.9$^\ast$} \ci{$-$18.5}{$-$11.6} & +4.6 \ci{$-$0.5}{9.7} & $-$0.02 & +0.04 & $-$0.09 \\
\quad Time-R1 (public) & \textbf{\boldmath $-$12.1} \ci{$-$15.3}{$-$9.0} & $-$0.6 \ci{$-$5.6}{4.5} & +0.03 & +0.01 & $-$0.02 \\
\quad VideoChat-R1 (public) & n/a & $-$0.1 \ci{$-$7.2}{7.2} & n/a & +0.06 & n/a \\
\addlinespace[2pt]
\multicolumn{6}{@{}l}{\emph{Placebo-controlled, against the base it was trained from}} \\
\rowcolor{pcrow}\quad Qwen2.5-VL-7B, $\alpha{=}0.5$ & \textbf{+21.5} \ci{16.7}{26.3} & \textbf{+10.4} \ci{5.4}{15.4} & +0.79 & +0.07 & +0.86 \\
\rowcolor{pcrow}\quad TimeZero, $\alpha{=}0.5$$^\ddagger$ & \textbf{+0.8} \ci{0.1}{1.6} & \textbf{+6.6} \ci{3.7}{9.7} & +0.44 & +0.34 & +0.62 \\
\rowcolor{pcrow}\quad Time-R1, $\alpha{=}0.25$ & \textbf{+3.2} \ci{1.4}{5.1} & \textbf{+6.6} \ci{3.7}{9.5} & +1.06 & +0.07 & +0.83 \\
\rowcolor{pcrow}\quad VideoChat-R1, $\alpha{=}0.5$$^\ddagger$ & n/a & \textbf{+15.0} \ci{11.8}{18.4} & n/a & +1.40 & n/a \\
\bottomrule
\end{tabular}
\par\smallskip
{\scriptsize\raggedright $^\ast$Both models return an interval for every question under the mask, so both changes equal minus the base's rate. $^\ddagger$The left block is evaluated with the text before the video, as released, and the right block with the video first. The entries marked n/a belong to VideoChat-R1, whose template fixes its tags and the bracket after the opening tag, leaving only the timestamps free, and has no way to decline. Brackets give 95\% intervals over videos.\par}
\end{table}

\subsection{Queries about another video}
\label{app:mismatch}
Table~\ref{tab:s4} pairs, question by question and for eleven configurations of three bases, abstention on the matched query with abstention on the annotation sentence of a different video. None of the four TimeZero configurations abstains on either query, and each answers mismatched queries at the accuracy of a constant window.

\begin{table}[!htbp]
\centering\scriptsize\setlength{\tabcolsep}{3pt}\renewcommand{\arraystretch}{1.05}
\caption{\textbf{Given the query of another video, the placebo-controlled Qwen2.5-VL-7B and Time-R1 models decline more often than their bases, and no TimeZero configuration declines.} Each question is asked once with its own annotation sentence as the query and once with the sentence of another video, which gives $n$ pairs. \emph{abst.} is the strict abstention rate, the share of outputs that are verbatim \texttt{There are none.}, and \Rfive{} is measured on the same pairs. Each $\Delta$ is the foreign-query value minus the own-query value, in points, with its 95\% interval. Text-first rows put the query before the video.}
\label{tab:s4}
\begin{tabular}{@{}>{\raggedright\arraybackslash}p{4.8cm}cccccc@{}}
\toprule
Model / objective & $n$ & abst.\ own & abst.\ foreign & $\Delta$ abst. & \Rfive{} own & $\Delta$\Rfive{} \\
\midrule
Qwen2.5-VL-7B, base & 1000 & 0.003 & 0.119 & +11.6 \ci{7.8}{15.9} & 0.485 & $-$33.0 \ci{$-$37.8}{$-$28.2} \\
Qwen2.5-VL-7B, placebo-controlled $\alpha{=}0.5$, $\beta{=}0.03$ & 1000 & 0.164 & 0.827 & +66.3 \ci{60.7}{71.4} & 0.528 & $-$47.1 \ci{$-$51.4}{$-$42.7} \\
Qwen2.5-VL-7B, placebo-controlled $\alpha{=}0.5$ & 1000 & 0.052 & 0.210 & +15.8 \ci{11.1}{20.7} & 0.523 & $-$40.5 \ci{$-$44.8}{$-$36.1} \\
Qwen2.5-VL-7B, SFT & 1000 & 0.000 & 0.003 & +0.3 \ci{0.0}{0.8} & 0.599 & $-$32.3 \ci{$-$37.3}{$-$27.7} \\
TimeZero, base & 1000 & 0.000 & 0.000 & +0.0 \ci{0.0}{0.0} & 0.694 & $-$38.2 \ci{$-$43.1}{$-$33.4} \\
TimeZero, placebo-controlled $\alpha{=}0.5$, $\beta{=}0.03$ & 1000 & 0.000 & 0.000 & +0.0 \ci{0.0}{0.0} & 0.687 & $-$38.2 \ci{$-$43.1}{$-$33.6} \\
TimeZero, placebo-controlled $\alpha{=}1$ & 1000 & 0.000 & 0.000 & +0.0 \ci{0.0}{0.0} & 0.688 & $-$38.7 \ci{$-$43.8}{$-$33.7} \\
TimeZero, SFT, text-first & 1000 & 0.000 & 0.000 & +0.0 \ci{0.0}{0.0} & 0.680 & $-$41.1 \ci{$-$46.0}{$-$36.2} \\
Time-R1, base & 1000 & 0.000 & 0.050 & +5.0 \ci{2.5}{8.0} & 0.616 & $-$37.3 \ci{$-$42.2}{$-$32.4} \\
Time-R1, placebo-controlled $\alpha{=}0.5$ & 1000 & 0.018 & 0.404 & +38.6 \ci{32.6}{44.9} & 0.637 & $-$48.4 \ci{$-$52.9}{$-$43.8} \\
Time-R1, placebo-controlled $\alpha{=}1$ & 1000 & 0.194 & 0.857 & +66.3 \ci{60.5}{71.6} & 0.542 & $-$49.2 \ci{$-$53.7}{$-$44.7} \\
\bottomrule
\end{tabular}
\end{table}

\section{Reproduction}
\label{app:repro}
\subsection{Protocols and training setup}
\label{app:protocols}
\paragraph{Inference protocols.}
Every model generates at 2 frames per second with greedy decoding. TimeLens-8B, Qwen3-VL-8B and our adapters of TimeLens-8B generate with TimeLens's released pixel budget of $14{,}336\times32\times32$ and are audited at $3{,}584\times28\times28$; every other model uses $3{,}584\times28\times28$ in both. Against Qwen2.5-VL-7B (Section~\ref{sec:findings}), TimeZero and Time-R1 run with the Qwen base under Time-R1's prompt template with the video before the text, and VideoChat-R1 with the Qwen base under VideoChat-R1's own template. When a public checkpoint is the base of a trained model (Sections~\ref{sec:objective} and~\ref{sec:readouts}), both run under the checkpoint's released prompt text with the video placed first; rows marked text-first use the order of TimeZero's and VideoChat-R1's released code. TimeLens-8B uses its own. The scored answer string follows the answer format of each template: under Time-R1's template it is \texttt{<answer> 20.80 to 30.00 </answer>}, as in our training targets. Appendix~\ref{app:order} evaluates every base under both orders. Abstention is recorded separately from parse failure: the strict criterion counts only the verbatim string \texttt{There are none.}, and the wide criterion counts any refusal wording.

\paragraph{Training.}
All trained models are LoRA adapters \citep{hu2022lora}. All except the 16-frame IoU-reward GRPO have rank 16 (LoRA alpha 32, dropout 0.05) on the attention and MLP projections. All except the IoU-reward GRPO runs are trained with AdamW at a constant learning rate of $5\times10^{-6}$ and eight pairs per update; the settings of the IoU-reward GRPO runs at 2 frames per second are given in Appendix~\ref{app:add-matched}. These models, apart from the budget, frame and two-stage runs of Appendix~\ref{app:budget} and the runs of Appendix~\ref{app:added}, make one pass over the same 2{,}400 pairs from the Charades-STA training split (8{,}434 in the full pool), with the same trainer and schedule. Each pair is edited by two of five operators, chosen among those valid for it, least-used first. Mask replaces the selected frames with black, and swap replaces them with frames from another training video. Pixelation averages each selected frame over a $16\times16$ grid of blocks, and temporal shift replaces each selected frame with the one half the video away. Temporal shuffle permutes the selected frames, never as the identity; it is used only where the query or the pair makes the order of frames meaningful and at least two frames are selected. The operator weights $w_{\mathrm{op}}$ are Group-DRO-style weights \citep{groupdro2020} over nine groups that cross the operator with whether the query makes the order of frames meaningful (a state change or a temporal order, against a generic action), normalised over the pair's two operators and updated with step size $0.05$ on each group's loss relative to its running mean. 

\paragraph{Baselines and cost.}
The SFT model is the same trainer with the evidence and placebo terms disabled ($\beta\to0$, no bound). The IoU-reward GRPO model follows \citet{deepseekmath2024} with a format and an IoU reward. It uses LoRA rank 64 and a KL weight of 0.04 and runs 700 steps of four questions with eight sampled answers each, on 16 frames at 224\,px; the questions are drawn from the 12{,}383 Charades-STA training questions whose annotated window holds at least one of the 16 sampled frames (of the split's 12{,}408). This IoU-GRPO trains on 16 frames per video. The IoU-GRPO runs at the audit's 2 frames per second, with LoRA rank 16 on the same 2{,}400 pairs as our objective, are described in Appendix~\ref{app:add-matched}, and at both frame rates IoU-GRPO leaves $\Gamma$ near zero. The mDPO-style baseline follows the conditional preference of \citet{wang2024mdpo}: the evidence margin of Equation~\ref{eq:objective}, with the edited frames a contiguous quarter of the video placed at random without the annotation, at the same $\alpha{=}0.5$, $\beta{=}0.1$, operators and schedule, with uniform weights and no placebo terms. Each pair of the placebo-controlled objective takes ten forward passes, in which the trained and the frozen model each score the clean video and the evidence and placebo edits of two operators; our SFT, run in the same trainer, scores the clean video and the two evidence edits under both models (six passes). On the same base, a pass of SFT over the 2{,}400 pairs takes $0.66$ to $0.68$ of the time of a pass of the placebo-controlled objective at $\alpha{=}1$ (TimeZero, VideoChat-R1 and TimeLens-8B).

\subsection{Message order}
\label{app:order}
Table~\ref{tab:s13} evaluates every base under both orders, with its answering rate and accuracy on black videos. Message order moves accuracy in different directions for different bases: text-first moves TimeZero and VideoChat-R1 up by $2.8$ points each, Time-R1 down by $6.9$ and Qwen2.5-VL-7B down by $24.1$. Part of TimeZero's $+2.8$ comes from the prior: its black-video \Rfive{} rises from $0.136$ to $0.189$. For a trained adapter the order changes \Rfive{} by $+1.0$ $[-3.9, 5.8]$ points on 200 questions.

\begin{table}[!htbp]
\centering\papertablestyle
\caption{\textbf{Message order moves base models in family-specific directions, and it also moves their accuracy on a black video.} Each base model is evaluated under both message orders, with the video first (video-first) or the query text first (text-first), and everything else identical. $\Delta$\Rfive{} is text-first minus video-first, paired per question, in points, with its 95\% interval; \emph{black ans.} and \emph{black \Rfive{}} are the answering rate and the accuracy on all-black videos under the same order. A $^{\circ}$ marks the order the checkpoint ships with.}
\label{tab:s13}
\begin{tabular}{@{}llcccc@{}}
\toprule
Model & order & \Rfive{} & $\Delta$\Rfive{} & black ans. & black \Rfive{} \\
\midrule
Qwen2.5-VL-7B & video-first$^{\circ}$ & 0.480 & \NA & 0.116 & 0.007 \\
\quad & text-first & 0.238 & $-$24.1 \ci{$-$26.5}{$-$21.9} & 0.578 & 0.092 \\
\addlinespace[2pt]
TimeZero & video-first & 0.677 & \NA & 0.968 & 0.136 \\
\quad & text-first$^{\circ}$ & 0.705 & +2.8 \ci{1.3}{4.1} & 0.940 & 0.189 \\
\addlinespace[2pt]
Time-R1 & video-first$^{\circ}$ & 0.601 & \NA & 0.796 & 0.105 \\
\quad & text-first & 0.533 & $-$6.9 \ci{$-$8.8}{$-$4.8} & 0.762 & 0.131 \\
\addlinespace[2pt]
VideoChat-R1 & video-first & 0.668 & \NA & 1.000 & 0.153 \\
\quad & text-first$^{\circ}$ & 0.697 & +2.8 \ci{1.2}{4.5} & 0.948 & 0.211 \\
\addlinespace[2pt]
\bottomrule
\end{tabular}
\end{table}

\subsection{Release}
\label{app:release}
The code is released at \url{https://github.com/jinzijie-bupt/Accuracy-Is-Not-Evidence}, the question and training-pair identifiers with the per-question outputs at \url{https://huggingface.co/datasets/jinzijie-bupt/Accuracy-Is-Not-Evidence}, and the LoRA adapters of nine trained models at \url{https://huggingface.co/jinzijie-bupt}. The nine are the placebo-controlled model of each of the five base models at $\alpha{=}0.5$, IoU-GRPO with the objective from Qwen2.5-VL-7B and from TimeZero, and the SFT and IoU-GRPO baselines from Qwen2.5-VL-7B. The code contains the trainers, among them IoU-GRPO with and without the placebo-controlled objective, and the evaluation and scoring code; the ids of the 8{,}434-pair pool (the 2{,}400 training pairs are its first 2{,}400), the 200-question audit set, the 600 further test questions of Appendix~\ref{app:add-setup}, the 1{,}000-question generation set, the partner-event pairs, the reworded queries, the mismatched-query pairing and the Charades-CD matches of Appendix~\ref{app:add-cd}; the placebo draws of the behavioural audit; and the scripts that produce every table and figure.

\FloatBarrier
\section{Additional experiments}
\label{app:added}
\subsection{Setup of the additional experiments}
\label{app:add-setup}
Every model added in this appendix is compared with its own base, run again on the same cluster and software stack as the model. These models use more than one such stack, so absolute values can differ between tables and from the main tables, and we compare only paired differences within one stack. On the stack used for most new models, the Qwen2.5-VL-7B base reaches \Rfive{} $0.507$ on the full test split, against $0.480$ in Table~\ref{tab:audit}. A few models are paired with the base of the main results' stack, which reaches $0.480$, and their tables say so. Intervals are bootstrapped over videos unless a caption says otherwise. Besides the first 200 test questions, we audit 600 new test questions drawn at random from the test questions whose video is not among the 79 videos of the first 200. They cover 484 videos, and on that stack the base reaches \Rfive{} $0.507$ on them, as on the full split. Readouts on generated answers use the first 1{,}000 test questions unless a caption says otherwise, as in Appendix~\ref{app:regen}.

\subsection{IoU-GRPO with matched frames and data}
\label{app:add-matched}
\begin{table}[!htbp]
\centering\scriptsize\setlength{\tabcolsep}{3pt}\renewcommand{\arraystretch}{1.05}
\caption{\textbf{With frames and pairs matched, IoU-GRPO adds accuracy but not evidence dependence, and adding the placebo-controlled objective to its loss adds both.} The 16-frame rows repeat the values of the main results (Tables~\ref{tab:audit}, \ref{tab:s1} and~\ref{tab:s7}), with $\Gm$ on the first 200 test questions. All other rows were trained or re-audited at 2 frames per second on the stack of Appendix~\ref{app:add-setup}, where the base reaches \Rfive{} $0.507$, and are paired with the base re-run on that stack. $\Delta$\Rfive{} is the change in points on the full Charades-STA test split (the TimeZero base reaches $0.689$ on this stack), and $\Gm$ is in nats per token on the 600 new questions (Appendix~\ref{app:add-setup}). The QVHighlights columns give the same two quantities on that dataset (Appendix~\ref{app:qvh}; base \Rfive{} $0.407$ on this stack). The SFT row and the first $\alpha{=}0.5$ row are the models of Table~\ref{tab:audit}, and the third seed was trained on this stack. On the stack of the main results, where the base reaches $0.480$ on Charades-STA and $0.319$ on QVHighlights, the same two models gain $+9.4$ and $+4.4$ points on Charades-STA (Table~\ref{tab:audit}) and $+6.8$ and $+0.1$ on QVHighlights (Table~\ref{tab:s8}). Brackets give 95\% intervals over videos, and $^\dagger$ marks a change not paired by question.}
\label{tab:add-matched}
\begin{tabular}{@{}>{\raggedright\arraybackslash}p{5.3cm}cccc@{}}
\toprule
 & \multicolumn{2}{c}{Charades-STA} & \multicolumn{2}{c}{QVHighlights} \\
\cmidrule(lr){2-3}\cmidrule(l){4-5}
Model / objective & $\Delta$\Rfive{} & $\Gm$ & $\Delta$\Rfive{} & $\Gm$ \\
\midrule
\addlinespace[1pt]\multicolumn{5}{@{}l}{\textit{Qwen2.5-VL-7B, trained on 16 frames (main results)}} \\
IoU-GRPO, full training split & +13.6 \ci{11.9}{15.4} & 0.00 & \NA & \NA \\
\rowcolor{pcrow}placebo-controlled $\alpha{=}0.5$, 2{,}400 pairs & +2.5$^\dagger$ & +0.06 & \NA & \NA \\
\rowcolor{pcrow}placebo-controlled $\alpha{=}0.5$, 8{,}434 pairs & $-$5.7$^\dagger$ & +0.23 & \NA & \NA \\
\addlinespace[1pt]\multicolumn{5}{@{}l}{\textit{Qwen2.5-VL-7B, trained at 2 fps on the same 2{,}400 pairs}} \\
SFT & +9.1 \ci{7.5}{10.6} & $-$0.07 \ci{$-$0.10}{$-$0.04} & +2.0 \ci{0.3}{3.8} & +0.02 \ci{$-$0.02}{0.06} \\
IoU-GRPO, first seed & +15.9 \ci{14.1}{17.7} & +0.04 \ci{0.02}{0.07} & +12.2 \ci{10.1}{14.4} & +0.09 \ci{0.05}{0.13} \\
IoU-GRPO, second seed & +15.8 \ci{13.9}{17.7} & $-$0.07 \ci{$-$0.10}{$-$0.05} & \NA & \NA \\
IoU-GRPO, second seed, step 300 of 600 & +14.7 \ci{12.8}{16.6} & $-$0.06 \ci{$-$0.08}{$-$0.03} & \NA & \NA \\
\rowcolor{pcrow}placebo-controlled $\alpha{=}0.5$ & +3.9 \ci{2.6}{5.2} & +1.50 \ci{1.37}{1.64} & +0.4 \ci{$-$1.4}{2.3} & +1.19 \ci{1.00}{1.38} \\
\rowcolor{pcrow}placebo-controlled $\alpha{=}0.5$, third seed & +4.5 \ci{3.1}{6.0} & +1.49 \ci{1.38}{1.61} & \NA & \NA \\
\rowcolor{pcrow}IoU-GRPO with the placebo-controlled objective, first seed & +15.3 \ci{13.5}{17.1} & +0.88 \ci{0.81}{0.94} & +11.2 \ci{9.0}{13.5} & +0.61 \ci{0.52}{0.71} \\
\rowcolor{pcrow}IoU-GRPO with the placebo-controlled objective, second seed & +15.9 \ci{14.0}{17.8} & +0.92 \ci{0.85}{0.99} & \NA & \NA \\
\addlinespace[1pt]\multicolumn{5}{@{}l}{\textit{TimeZero, trained at 2 fps on the same 2{,}400 pairs}} \\
\rowcolor{pcrow}IoU-GRPO with the placebo-controlled objective & +1.0 \ci{0.0}{2.0} & +1.04 \ci{0.96}{1.12} & \NA & \NA \\
\bottomrule
\end{tabular}
\end{table}

Trained at 2 frames per second on the same 2{,}400 pairs as the placebo-controlled objective, IoU-GRPO gains about 16 points of \Rfive{} and leaves $\Gm$ within about $0.07$ of zero at both seeds (Table~\ref{tab:add-matched}). Both objectives start from Qwen2.5-VL-7B and make one pass over the pairs, IoU-GRPO in 600 steps of four questions and the placebo-controlled objective in 300 steps of eight pairs. The first IoU-GRPO seed sees the pairs in the order of the placebo-controlled run, and all runs use LoRA rank 16 (alpha 32, dropout 0.05) and the same software stack. IoU-GRPO here samples eight answers per question at up to 768 frames per video, rewards their format and IoU, and uses a KL weight of $0.04$ to the base and AdamW at a learning rate of $10^{-5}$ after 20 warm-up steps, with the policy loss averaged over all answer tokens of a step. With the 16-frame models of Table~\ref{tab:s7}, these runs cross the training frames with the objective. Only the 2 fps rows are matched in data, since the 16-frame IoU-GRPO trained on the full training split for 700 steps at LoRA rank 64 (Appendix~\ref{app:protocols}). Moving from 16 frames to 2 frames per second raises $\Gm$ under the placebo-controlled objective and leaves IoU-GRPO near zero in both settings, so the frames a model trains on do not account for IoU-GRPO's $\Gamma$.
IoU-GRPO also takes about twice the training time, and most of its accuracy gain is in place within the training time of the placebo-controlled objective. On one A100 GPU, three runs of the placebo-controlled objective on this schedule, which differ from the $\alpha{=}0.5$ model only in the constants of the placebo bound or in the answer tokens the margins score, took 7.4 to 7.6 hours each, and the first IoU-GRPO seed took 15.1 hours. We kept the checkpoint of the second seed after 300 of its 600 steps. At the pace of the first seed this is about 7.5 hours of training, and the checkpoint has seen half of the pairs. Its gain and its $\Gm$ are already close to those of the finished runs (Table~\ref{tab:add-matched}).

Adding the placebo-controlled objective to the IoU-GRPO loss keeps IoU-GRPO's accuracy and raises $\Gm$ to about $+0.9$ at both seeds. Its own placebo drop on the 600 new questions is $-0.02$ nats per token at both seeds, against $-0.08$ for the base, and $-0.01$ against $-0.03$ when trained from TimeZero. The added term has weight $0.01$, and the pairs and schedule are those of IoU-GRPO. On QVHighlights, where none of these models was trained, the combination gains within a point of IoU-GRPO alone and keeps a positive $\Gm$, while the placebo-controlled objective alone shows no accuracy gain there. Started from TimeZero, the same combination raises $\Gm$ to $+1.04$ without lowering in-domain accuracy.

\subsection{An independent sample of test videos}
\label{app:add-sample}

On the 600 new questions (Appendix~\ref{app:add-setup}), no model's $\Gm$ changes sign from the first 200 test questions, and every placebo-controlled model stays far above every accuracy-oriented one (Table~\ref{tab:add-sample}). The new questions are harder for the base than the first 200, on which the Qwen2.5-VL-7B base reaches \Rfive{} $0.600$ on this stack ($0.560$ on the stack of the main results), above its value on the new questions (Appendix~\ref{app:add-setup}). Within Qwen2.5-VL-7B and within TimeLens-8B, the three weights $\alpha$ order $\Gm$ as on the first 200 questions.

\begin{table}[!htbp]
\centering\scriptsize\setlength{\tabcolsep}{3pt}\renewcommand{\arraystretch}{1.05}
\caption{\textbf{On 600 test questions from videos outside the first 200 questions, $\Gm$ keeps its sign, and the placebo-controlled models stay far above the accuracy-oriented ones.} The first column repeats the main results on the first 200 test questions (Tables~\ref{tab:audit}, \ref{tab:families}, \ref{tab:s7} and~\ref{tab:s2}, Appendix~\ref{app:estimands}). The other columns re-audit the models on the 600 new questions, each model paired with its base re-run on the same stack; the last column gives the smallest and the largest value over three disjoint random subsets of 200 of the 600 questions. $\Gm$ is in nats per token, and brackets give 95\% intervals over videos. The third seed was trained on the stack of Appendix~\ref{app:add-setup}. The second seed and the public VideoChat-R1 checkpoint were not re-audited on the new questions. The 16-frame IoU-GRPO, the public TimeZero and Time-R1 checkpoints, $\alpha{=}0.25$, $\alpha{=}1$ and the 4{,}800-pair model are paired with the base of the main results' stack (Appendix~\ref{app:add-setup}). The public TimeZero and Time-R1 checkpoints are audited against Qwen2.5-VL-7B, and every placebo-controlled adapter against the base it was trained from.}
\label{tab:add-sample}
\begin{tabular}{@{}>{\raggedright\arraybackslash}p{5.0cm}ccc@{}}
\toprule
 & first 200 & \multicolumn{2}{c}{600 new questions} \\
\cmidrule(l){3-4}
Model & $\Gm$ & $\Gm$ & range over subsets \\
\midrule
\addlinespace[1pt]\multicolumn{4}{@{}l}{\textit{Trained for accuracy}} \\
SFT & $-$0.09 & $-$0.07 \ci{$-$0.10}{$-$0.04} & $-$0.08 to $-$0.06 \\
mDPO-style & $-$0.03 & $-$0.01 \ci{$-$0.04}{0.02} & $-$0.02 to 0.00 \\
IoU-GRPO, 16 frames & 0.00 & 0.00 \ci{$-$0.02}{0.02} & $-$0.01 to 0.00 \\
TimeZero (public) & $-$0.08 & $-$0.02 \ci{$-$0.05}{0.02} & $-$0.03 to $-$0.01 \\
Time-R1 (public) & +0.02 & +0.05 \ci{0.03}{0.06} & +0.04 to +0.05 \\
TimeLens-8B (public) against Qwen3-VL-8B & $-$0.16 & $-$0.14 \ci{$-$0.18}{$-$0.11} & $-$0.16 to $-$0.12 \\
\addlinespace[1pt]\multicolumn{4}{@{}l}{\textit{Placebo-controlled, Qwen2.5-VL-7B}} \\
\rowcolor{pcrow}$\alpha{=}0.25$ & +2.03 & +1.89 \ci{1.76}{2.02} & +1.83 to +1.96 \\
\rowcolor{pcrow}$\alpha{=}0.5$ & +1.63 & +1.50 \ci{1.37}{1.64} & +1.43 to +1.64 \\
\rowcolor{pcrow}$\alpha{=}0.5$, second seed & +1.48 & \NA & \NA \\
\rowcolor{pcrow}$\alpha{=}0.5$, third seed & \NA & +1.49 \ci{1.38}{1.61} & +1.39 to +1.56 \\
\rowcolor{pcrow}$\alpha{=}1$ & +1.03 & +0.91 \ci{0.79}{1.03} & +0.83 to +0.99 \\
\rowcolor{pcrow}$\alpha{=}0.5$, 4{,}800 pairs & +2.78 & +2.64 \ci{2.48}{2.79} & +2.54 to +2.80 \\
\addlinespace[1pt]\multicolumn{4}{@{}l}{\textit{Placebo-controlled, other bases}} \\
\rowcolor{pcrow}VideoChat-R1, $\alpha{=}0.5$ & +1.43 & +1.45 \ci{1.38}{1.51} & +1.37 to +1.50 \\
\rowcolor{pcrow}TimeZero, $\alpha{=}0.5$ & +1.40 & +1.21 \ci{1.12}{1.30} & +1.10 to +1.33 \\
\rowcolor{pcrow}Time-R1, $\alpha{=}0.5$ & +1.46 & +1.29 \ci{1.19}{1.40} & +1.24 to +1.39 \\
\rowcolor{pcrow}TimeLens-8B, $\alpha{=}0.25$ & +1.72 & +1.80 \ci{1.73}{1.88} & +1.76 to +1.87 \\
\rowcolor{pcrow}TimeLens-8B, $\alpha{=}0.5$ & +1.36 & +1.42 \ci{1.37}{1.48} & +1.42 to +1.43 \\
\rowcolor{pcrow}TimeLens-8B, $\alpha{=}1$ & +0.92 & +0.97 \ci{0.92}{1.01} & +0.95 to +0.98 \\
\bottomrule
\end{tabular}
\end{table}

The result also holds within smaller samples. We split the 600 questions at random into three disjoint subsets of 200. Every model in Table~\ref{tab:add-sample} keeps the sign of its $\Gm$ in each subset, except mDPO-style and the 16-frame IoU-GRPO, whose $\Gm$ stays within $0.02$ of zero in every subset.

\subsection{Noise and blur at three strengths}
\label{app:add-operators}

Under additive noise and Gaussian blur, the $\Gamma$ of the placebo-controlled model grows with the strength of the edit, while SFT's stays at or below zero (Table~\ref{tab:add-operators}). At the settings of Table~\ref{tab:unseen}, noise with a standard deviation of 40 in 8-bit pixel values and blur with a standard deviation of 3 pixels, the edits remove little even from the base. Its $\psi$ is $0.06$ under noise and $0.03$ under blur on the 600 new questions, so there is little for a trained model to depend on. We therefore add two stronger settings of each operator.

\begin{table}[!htbp]
\centering\scriptsize\setlength{\tabcolsep}{3pt}\renewcommand{\arraystretch}{1.05}
\caption{\textbf{Under noise and Gaussian blur, the placebo-controlled model's $\Gamma$ grows with the strength of the edit, while SFT's stays at or below zero.} Qwen2.5-VL-7B models on the 600 new questions, each paired with its base re-run on the same software stack. The noise standard deviation is in 8-bit pixel values and the blur standard deviation in pixels; the first setting of each operator is the one of Table~\ref{tab:unseen}. Base $\psi$ is the base's $\psi$ on the questions that both the base and the $\alpha{=}0.5$ model localise. $\Gamma$ is measured under the operator of its row, in nats per token, and brackets give 95\% intervals over videos. The model trained on 4{,}800 pairs is paired with the base of the main results' stack and was not audited at the middle settings. Two seeds audited against each other at the first setting give an interval of $\Gamma$ that reaches at most $0.030$ from zero under noise and $0.049$ under blur, and the stronger settings were not re-measured.}
\label{tab:add-operators}
\begin{tabular}{@{}lcccc@{}}
\toprule
 & & \multicolumn{3}{c}{$\Gamma$} \\
\cmidrule(l){3-5}
Operator, setting & base $\psi$ & \shortstack{placebo-controlled\\$\alpha{=}0.5$} & SFT & \shortstack{placebo-controlled\\$\alpha{=}0.5$, 4{,}800 pairs} \\
\midrule
noise, 40 & 0.06 & +0.28 \ci{0.21}{0.35} & $-$0.04 \ci{$-$0.06}{$-$0.03} & +0.36 \ci{0.29}{0.44} \\
noise, 80 & 0.11 & +0.58 \ci{0.48}{0.68} & $-$0.08 \ci{$-$0.09}{$-$0.06} & \NA \\
noise, 120 & 0.13 & +0.85 \ci{0.74}{0.97} & $-$0.10 \ci{$-$0.12}{$-$0.08} & +1.47 \ci{1.33}{1.61} \\
\addlinespace[2pt]
Gaussian blur, 3 & 0.03 & +0.25 \ci{0.18}{0.32} & $-$0.01 \ci{$-$0.03}{0.00} & +0.38 \ci{0.32}{0.45} \\
Gaussian blur, 6 & 0.04 & +0.59 \ci{0.49}{0.70} & $-$0.02 \ci{$-$0.04}{0.00} & \NA \\
Gaussian blur, 10 & 0.07 & +0.97 \ci{0.85}{1.09} & $-$0.02 \ci{$-$0.04}{$-$0.00} & +1.78 \ci{1.64}{1.92} \\
\bottomrule
\end{tabular}
\end{table}

Each operator also has its own seed resolution. We measured it with one seed pair, auditing a third seed of the placebo-controlled $\alpha{=}0.5$ model (Table~\ref{tab:add-sample}) against the first on the 600 new questions. The farthest end of the interval of $\Gamma$ lies $0.030$ from zero under noise, $0.049$ under blur and $0.065$ under patch scrambling, against $0.08$ under mask for the same pair on the same questions (Appendix~\ref{app:add-seeds}). On these questions the $\Gamma$ of the $\alpha{=}0.5$ model lies at every setting far beyond the noise and blur resolutions measured at the first setting, and SFT's $\Gamma$ under noise lies more than $0.030$ below zero at every strength (Table~\ref{tab:add-operators}). The six values of the Qwen2.5-VL-7B models under noise and blur in Table~\ref{tab:unseen} also exceed these resolutions, but they were measured on the first 200 questions, whose intervals are wider, and the interval of the $\alpha{=}1$ model under noise includes zero.

\subsection{The annotation prior}
\label{app:add-prior}
The placebo-controlled models keep their $\Gamma$ on the questions that the prior of the training labels cannot answer. We take as the prior the single window, in fractions of the video's length, with the highest mean IoU against the 12{,}408 Charades-STA training labels. This window, $[0, 0.47]$, uses neither the video nor the query and reaches \Rfive{} $0.27$ on the test split. It answers a question when its IoU with the annotated interval is at least $0.5$ and fails it when the IoU is below $0.3$. Table~\ref{tab:add-prior} reads every model on both groups, with $\Gamma$ on the 600 new questions (Appendix~\ref{app:add-setup}) and accuracy on the full test split.

Where the window fails, every model trained with the placebo-controlled objective, alone or added to IoU-GRPO, keeps a $\Gamma$ far beyond the audit's resolution, and every model trained for accuracy alone stays within it. Trained on 2{,}400 pairs, the placebo-controlled objective used alone gains accuracy almost only on these questions at both seeds, while every other model, including the one trained on 4{,}800 pairs, gains on both groups.

The last column of Table~\ref{tab:add-prior} asks whether training moves predictions towards the window, as the change in the mean IoU between a model's predictions and the window. The placebo-controlled models, alone and added to IoU-GRPO, move slightly away from it. SFT at both budgets, mDPO-style and the 16-frame IoU-GRPO move slightly towards it, with intervals that exclude zero, and IoU-GRPO at 2 frames per second does not move. The questions the window fails differ from the others in where their events lie, so we read the comparison between the two groups as directional only.

\begin{table}[!htbp]
\centering\scriptsize\setlength{\tabcolsep}{2.5pt}\renewcommand{\arraystretch}{1.06}
\caption{\textbf{Where the annotation prior fails, the placebo-controlled models keep their $\Gamma$ and the models trained for accuracy alone stay near zero.} The prior is the window $[0, 0.47]$ of each video's length fitted to the training labels; it \emph{fails} a question when its IoU with the annotated interval is below $0.3$ and \emph{answers} it when the IoU is at least $0.5$. $\Gm$ is in nats per token on the 299 and 158 questions of the two groups among the 600 new questions ($n_\cap$ from 109 to 129 and from 89 to 99). $\Delta$\Rfive{} is the paired change in points on the 1{,}885 and 1{,}001 questions of the two groups in the full test split. The last column is the change in the mean IoU between the model's predictions and the window over the full test split, in points. Each model is paired with its base on the same software stack, whose \Rfive{} on the full test split heads each block. Intervals resample videos for $\Gamma$ and questions for the other columns, and the $\Gm$ column for the questions the window answers gives estimates only.}
\label{tab:add-prior}
\begin{tabular}{@{}>{\raggedright\arraybackslash\hangindent=2em\hangafter=1}p{3.7cm}ccccc@{}}
\toprule
 & \multicolumn{2}{c}{$\Gm$} & \multicolumn{2}{c}{$\Delta$\Rfive{} (points)} & $\Delta$IoU with \\
\cmidrule(lr){2-3}\cmidrule(lr){4-5}
Model & window fails & window answers & window fails & window answers & the window \\
\midrule
\multicolumn{6}{@{}l}{\emph{Against the base at \Rfive{} 0.507}} \\
\quad SFT & +0.07 \ci{0.04}{0.09} & $-$0.24 & +8.0 \ci{6.1}{9.9} & +9.0 \ci{6.7}{11.4} & +1.3 \ci{0.8}{1.7} \\
\quad mDPO-style & +0.14 \ci{0.11}{0.17} & $-$0.18 & +7.6 \ci{5.7}{9.4} & +7.8 \ci{5.6}{10.1} & +0.7 \ci{0.3}{1.1} \\
\quad IoU-GRPO, 2 fps & +0.13 \ci{0.10}{0.16} & $-$0.05 & +17.8 \ci{15.7}{19.9} & +11.0 \ci{8.3}{13.7} & +0.1 \ci{$-$0.4}{0.5} \\
\rowcolor{pcrow}\quad IoU-GRPO, 2 fps, with the placebo-controlled objective & +1.02 \ci{0.92}{1.12} & +0.78 & +18.1 \ci{16.0}{20.3} & +9.6 \ci{7.0}{12.2} & $-$0.8 \ci{$-$1.3}{$-$0.3} \\
\rowcolor{pcrow}\quad placebo-controlled $\alpha{=}0.5$ & +1.35 \ci{1.16}{1.53} & +1.83 & +5.1 \ci{3.4}{6.9} & +0.4 \ci{$-$1.7}{2.4} & $-$0.6 \ci{$-$0.9}{$-$0.2} \\
\rowcolor{pcrow}\quad placebo-controlled $\alpha{=}0.5$, third seed & +1.42 \ci{1.25}{1.59} & +1.67 & +5.0 \ci{3.1}{6.7} & +0.9 \ci{$-$1.7}{3.2} & $-$1.5 \ci{$-$2.0}{$-$1.0} \\
\addlinespace[2pt]
\multicolumn{6}{@{}l}{\emph{Against the base at \Rfive{} 0.480}} \\
\quad IoU-GRPO, 16 frames & +0.02 \ci{$-$0.00}{0.05} & $-$0.04 & +15.5 \ci{13.4}{17.6} & +12.0 \ci{9.6}{14.5} & +0.9 \ci{0.4}{1.3} \\
\quad SFT, 4{,}800 pairs & +0.06 \ci{0.03}{0.10} & $-$0.24 & +13.6 \ci{11.5}{15.7} & +8.3 \ci{5.7}{10.9} & +0.8 \ci{0.3}{1.3} \\
\rowcolor{pcrow}\quad placebo-controlled $\alpha{=}0.5$, 4{,}800 pairs & +2.71 \ci{2.52}{2.88} & +2.66 & +9.2 \ci{7.4}{11.1} & +5.1 \ci{3.0}{7.2} & $-$0.7 \ci{$-$1.1}{$-$0.2} \\
\bottomrule
\end{tabular}
\end{table}

\subsection{Frames next to the event}
\label{app:add-flank}
\begin{table}[!htbp]
\centering\papertablestyle
\caption{\textbf{$\Gamma$ appears where the event's frames are edited and not on the frames next to them.} The edited window is the annotated interval (\emph{event}) or that interval moved by one event length onto the frames just before or just after it; in every column the scored answer is the annotated interval. Values are in nats per token on the 600 new questions, with $n_\cap$ from 232 to 282 per entry and 95\% intervals over videos. The base row gives the base's $\psi$ on the questions of the placebo-controlled pair, and the other rows give $\Gamma$ against that base (\Rfive{} 0.507 on the full test split).}
\label{tab:add-flank}
\begin{tabular}{@{}>{\raggedright\arraybackslash\hangindent=2em\hangafter=1}p{4.7cm}ccc@{}}
\toprule
 & before the event & event & after the event \\
\midrule
\multicolumn{4}{@{}l}{\emph{Mask}} \\
\quad base, $\psi$ & +0.01 & +0.18 & $-$0.07 \\
\quad SFT & +0.00 \ci{$-$0.01}{0.01} & $-$0.07 \ci{$-$0.10}{$-$0.04} & +0.05 \ci{0.04}{0.07} \\
\rowcolor{pcrow}\quad placebo-controlled $\alpha{=}0.5$ & +0.03 \ci{$-$0.01}{0.07} & +1.50 \ci{1.37}{1.64} & $-$0.01 \ci{$-$0.08}{0.06} \\
\rowcolor{pcrow}\quad IoU-GRPO, 2 fps, with the placebo-controlled objective & 0.00 \ci{$-$0.03}{0.03} & +0.88 \ci{0.81}{0.94} & $-$0.05 \ci{$-$0.08}{$-$0.02} \\
\addlinespace[2pt]
\multicolumn{4}{@{}l}{\emph{Swap}} \\
\quad base, $\psi$ & 0.00 & +0.14 & $-$0.08 \\
\quad SFT & $-$0.02 \ci{$-$0.03}{$-$0.00} & $-$0.05 \ci{$-$0.08}{$-$0.03} & +0.04 \ci{0.02}{0.06} \\
\rowcolor{pcrow}\quad placebo-controlled $\alpha{=}0.5$ & +0.02 \ci{$-$0.03}{0.06} & +1.30 \ci{1.18}{1.43} & $-$0.01 \ci{$-$0.08}{0.07} \\
\rowcolor{pcrow}\quad IoU-GRPO, 2 fps, with the placebo-controlled objective & $-$0.05 \ci{$-$0.08}{$-$0.02} & +0.72 \ci{0.66}{0.77} & $-$0.07 \ci{$-$0.11}{$-$0.03} \\
\bottomrule
\end{tabular}
\end{table}

The dependence that training adds sits on the frames of the event, with none on the frames next to it. To test this we moved the edited window by one event length, onto the frames just before or just after the annotated interval (clipped at the ends of the video), and kept the scored answer and the size of the placebo unchanged. The base's $\psi$ on these neighbouring frames is close to zero. After training, every $\Gamma$ on them lies within $0.08$ of zero under both operators, while both models trained with the placebo-controlled objective keep a $\Gamma$ on the event itself far beyond the resolution (Table~\ref{tab:add-flank}). Frames farther from the event fall in the placebo's pool, and the placebo-controlled adapters keep their placebo drop within $0.10$ nats per token of their base's (Section~\ref{sec:objective}).
\subsection{Charades-CD and reliance on the training split}
\label{app:add-cd}
Every model we tested keeps a positive accuracy gain on the Charades-CD test-ood questions, and a higher $\Gamma$ does not come with a smaller share of that gain kept out of distribution. Charades-CD \citep{yuan2021closer} re-splits Charades-STA so that the test-ood moment distribution no longer matches training (Table~\ref{tab:s14}). Its full test sets partly reuse the Charades-STA training split. We keep the 742 test-ood and 210 test-iid questions that match a question of the Charades-STA test split in video, interval and sentence, so none of them was seen in training, and Table~\ref{tab:add-cd} gives each model's paired change in \Rfive{} from its base on both sets.

SFT and the placebo-controlled model trained on the same 2{,}400 pairs each keep about a third of their test-iid gain on test-ood. Across all models the share kept runs from $0.35$ to $0.71$, and the placebo-controlled models span that whole range, so a higher $\Gamma$ comes with no greater reliance on the training split. The objective does not transfer better than SFT either. Its test-ood gain is smaller than SFT's at both training budgets, and its larger share at 4{,}800 pairs is a share of a smaller test-iid gain. Table~\ref{tab:s14} compares the two splits over the full Charades-CD test sets, which partly reuse the Charades-STA training split, and there SFT loses the most accuracy from test-iid to test-ood. With 210 test-iid questions from 74 videos, the test-iid intervals are wide.

\begin{table}[!htbp]
\centering\papertablestyle
\caption{\textbf{Every model keeps a positive gain on the Charades-CD test-ood questions, and the share of its test-iid gain kept there is not ordered by $\Gamma$.} The questions are the 210 test-iid (74 videos) and 742 test-ood (286 videos) questions of Charades-CD \citep{yuan2021closer} that are also Charades-STA test questions. Each entry is the paired change in \Rfive{} from the model's base on the same software stack, in points, with 95\% intervals over videos; the block heading gives the base's \Rfive{} on each set. The last column is the test-ood change divided by the test-iid change.}
\label{tab:add-cd}
\begin{tabular}{@{}>{\raggedright\arraybackslash\hangindent=2em\hangafter=1}p{5.2cm}ccc@{}}
\toprule
Model & $\Delta$\Rfive{} test-iid & $\Delta$\Rfive{} test-ood & ood / iid \\
\midrule
\multicolumn{4}{@{}l}{\emph{Against the base at \Rfive{} 0.490 (test-iid) and 0.453 (test-ood)}} \\
\quad SFT & +19.5 \ci{12.2}{27.8} & +7.0 \ci{3.8}{10.2} & 0.36 \\
\quad mDPO-style & +16.7 \ci{9.9}{23.7} & +6.2 \ci{3.4}{8.9} & 0.37 \\
\quad IoU-GRPO, 2 fps & +27.1 \ci{18.2}{36.0} & +13.6 \ci{10.1}{17.3} & 0.50 \\
\rowcolor{pcrow}\quad IoU-GRPO, 2 fps, with the placebo-controlled objective & +26.2 \ci{17.5}{34.8} & +12.1 \ci{8.5}{15.8} & 0.46 \\
\rowcolor{pcrow}\quad placebo-controlled $\alpha{=}0.5$ & +8.1 \ci{2.0}{14.7} & +2.8 \ci{0.3}{5.7} & 0.35 \\
\rowcolor{pcrow}\quad placebo-controlled $\alpha{=}0.5$, third seed & +7.1 \ci{$-$1.4}{15.0} & +3.2 \ci{0.4}{5.9} & 0.45 \\
\addlinespace[2pt]
\multicolumn{4}{@{}l}{\emph{Against the base of the main results' stack, at \Rfive{} 0.495 (test-iid) and 0.431 (test-ood)}} \\
\quad IoU-GRPO, 16 frames & +18.1 \ci{10.6}{25.6} & +12.0 \ci{8.4}{15.7} & 0.66 \\
\quad SFT, 4{,}800 pairs & +20.0 \ci{13.0}{27.2} & +9.2 \ci{5.2}{13.0} & 0.46 \\
\rowcolor{pcrow}\quad placebo-controlled $\alpha{=}0.5$, 4{,}800 pairs & +10.5 \ci{4.9}{16.3} & +7.4 \ci{4.4}{10.6} & 0.71 \\
\bottomrule
\end{tabular}
\end{table}

\subsection{Refusal and dependence}
\label{app:add-refusal}
The rise in $\Gamma$ does not reduce to a learned refusal of masked input. Table~\ref{tab:split} already shows that the answers the evidence mask takes from the placebo-controlled models are not all refusals, since among the intervals they still return the share that is wrong rises as well. Applied to the timestamp tokens alone, which follow the model's commitment to an interval, the two margins give a large $\Gamma$ on those tokens (Appendix~\ref{app:add-tsonly}).

\paragraph{Questions still answered.}
On the questions a placebo-controlled model still answers with an interval under the evidence mask, its $\Gamma$ stays large. We took these questions from the regenerated answers with the evidence frames masked (Appendix~\ref{app:regen}) and restricted the likelihood audit of the first 200 test questions to them, with each model and its base re-run on the same software stack (Table~\ref{tab:add-answered}). For the placebo-controlled model trained on 2{,}400 pairs, 72 of the 108 questions that enter its $\Gamma$ remain, and its $\Gamma$ on them is about two thirds of its $\Gamma$ on all 108. The models trained for accuracy alone answer almost every question under the mask and stay within the audit's resolution on both sets.

\begin{table}[!htbp]
\centering\papertablestyle
\caption{\textbf{On the questions a model still answers under the evidence mask, the placebo-controlled models keep most of their $\Gamma$.} $\Gm$ is in nats per token on the first 200 test questions, over all of $H_\cap$ and over the part of $H_\cap$ on which the trained model still returns an interval when the evidence frames are masked in its regenerated answers, with 95\% intervals over videos. Each model is paired with its base re-run on the same software stack, whose \Rfive{} on the full test split heads each block. In the main results the placebo-controlled model at $\alpha{=}0.5$ has $\Gm=+1.63$ on $n_\cap=112$ (Table~\ref{tab:scale}), and the one trained on 4{,}800 pairs $+2.78$ on $n_\cap=105$ (Table~\ref{tab:s1}).}
\label{tab:add-answered}
\begin{tabular}{@{}>{\raggedright\arraybackslash\hangindent=2em\hangafter=1}p{5.0cm}cccc@{}}
\toprule
 & \multicolumn{2}{c}{all of $H_\cap$} & \multicolumn{2}{c@{}}{still answered under the mask} \\
\cmidrule(lr){2-3}\cmidrule(l){4-5}
Model & $n_\cap$ & $\Gm$ & $n$ & $\Gm$ \\
\midrule
\multicolumn{5}{@{}l}{\emph{Against the base at \Rfive{} 0.507}} \\
\quad IoU-GRPO, 2 fps & 110 & +0.01 \ci{$-$0.06}{0.07} & 108 & +0.01 \ci{$-$0.06}{0.08} \\
\rowcolor{pcrow}\quad IoU-GRPO, 2 fps, with the placebo-controlled objective & 110 & +0.85 \ci{0.71}{1.00} & 108 & +0.83 \ci{0.70}{0.96} \\
\rowcolor{pcrow}\quad placebo-controlled $\alpha{=}0.5$ & 108 & +1.62 \ci{1.35}{1.89} & 72 & +1.02 \ci{0.80}{1.26} \\
\rowcolor{pcrow}\quad placebo-controlled $\alpha{=}0.5$, third seed & 107 & +1.62 \ci{1.38}{1.84} & 54 & +1.01 \ci{0.76}{1.27} \\
\addlinespace[2pt]
\multicolumn{5}{@{}l}{\emph{Against the base at \Rfive{} 0.480}} \\
\quad SFT, 4{,}800 pairs & 105 & $-$0.13 \ci{$-$0.21}{$-$0.06} & 105 & $-$0.13 \ci{$-$0.21}{$-$0.06} \\
\rowcolor{pcrow}\quad placebo-controlled $\alpha{=}0.5$, 4{,}800 pairs & 105 & +2.77 \ci{2.54}{3.03} & 79 & +2.57 \ci{2.30}{2.87} \\
\bottomrule
\end{tabular}
\end{table}

\paragraph{Forced decoding.}
When the model cannot decline, masking the event still costs the placebo-controlled model more accuracy than it costs SFT. We regenerated the first 1{,}000 test questions with decoding that starts with the answer tag and is constrained to the interval grammar, so every output is an interval. Relative to the base, the evidence mask then lowers the \Rfive{} of the placebo-controlled model at $\alpha{=}0.5$ by $15$ $[12, 18]$ points more than the placebo mask does, and that of SFT by $5$ $[2, 8]$ points more.

\paragraph{IoU-GRPO with the objective.}
IoU-GRPO with the placebo-controlled objective raises $\Gamma$ while almost never declining. With the evidence frames masked it returns no interval on $1.4\%$ of the first 1{,}000 test questions ($3.0\%$ for a second seed), against $11.0\%$ for its base, and the correct answers it loses under the mask become wrong intervals.

\paragraph{Equal accuracy.}
At the accuracy gain of the placebo-controlled model trained on 4{,}800 pairs, SFT has a negative $\Gamma$. The checkpoint of an SFT run on 4{,}800 pairs taken after its first 800 pairs comes closest to that model in accuracy gain, and its $\Gamma$ is below zero (Table~\ref{tab:add-matched-acc}). At the later checkpoints of the same run, after 1{,}600 to 4{,}800 pairs, SFT's gain rises to $13.0$ points and its $\Gamma$ on the first 200 test questions stays between $-0.15$ and $-0.13$.

\subsection{Occlusion and boundary jitter}
\label{app:add-occlusion}
\begin{table}[!htbp]
\centering\papertablestyle
\caption{\textbf{With the event frozen or covered by a grey box, and under a 10\% jitter of the boundaries, the placebo-controlled models stay above the audit's resolution and SFT stays at or below zero.} $\Gm$ in nats per token on the first 200 test questions, each model against its base on the $n_\cap$ questions both localise (108 for the Qwen2.5-VL-7B pairs, 132 for the TimeZero re-runs). Jitter moves the boundaries of the edited window by 10\% of the event's length. The TimeZero entry for the annotated interval is that of Table~\ref{tab:families}; every other entry is a re-run, paired with its base re-run on the same software stack. Brackets give 95\% intervals over videos.}
\label{tab:add-occlusion}
\begin{tabular}{@{}lccc@{}}
\toprule
 & \multicolumn{2}{c}{placebo-controlled, $\alpha{=}0.5$} & SFT \\
\cmidrule(lr){2-3}\cmidrule(l){4-4}
Edited frames & Qwen2.5-VL-7B & TimeZero & Qwen2.5-VL-7B \\
\midrule
annotated interval & +1.62 \ci{1.35}{1.89} & +1.40 \ci{1.19}{1.60} & $-$0.09 \ci{$-$0.16}{$-$0.03} \\
each boundary moved inward & +1.10 \ci{0.87}{1.37} & +1.06 \ci{0.87}{1.27} & $-$0.05 \ci{$-$0.09}{$-$0.01} \\
each boundary moved outward & +1.74 \ci{1.47}{2.00} & +1.53 \ci{1.33}{1.73} & $-$0.10 \ci{$-$0.16}{$-$0.03} \\
window shifted left & +1.52 \ci{1.26}{1.76} & +1.27 \ci{1.07}{1.47} & $-$0.12 \ci{$-$0.18}{$-$0.06} \\
window shifted right & +1.28 \ci{1.04}{1.55} & +1.27 \ci{1.08}{1.49} & $-$0.04 \ci{$-$0.09}{0.01} \\
event frozen on the frame before it & +1.19 \ci{0.93}{1.46} & +1.08 \ci{0.89}{1.25} & $-$0.03 \ci{$-$0.08}{0.01} \\
grey box over the central 60\% & +0.60 \ci{0.36}{0.87} & +0.57 \ci{0.39}{0.77} & $-$0.03 \ci{$-$0.05}{0.00} \\
\bottomrule
\end{tabular}
\end{table}

Freezing the event, covering its frames with a grey box and moving its boundaries by 10\% leave the three audited models in their class (Table~\ref{tab:add-occlusion}). With the models of the main results and the resolution of Section~\ref{sec:audit} held fixed, the placebo-controlled models of Qwen2.5-VL-7B and TimeZero keep a $\Gm$ above $0.23$ under every condition, and SFT stays at or below zero. Each condition changes which frames are edited or how they are edited, while the scored answer stays the annotated interval and the placebo edit covers as many frames outside the edited window. Moving each boundary inward or outward by 10\% of the event's length narrows or widens the masked window, and shifting both boundaries by the same amount moves the window partly off the event. Freezing replaces each edited frame with the nearest earlier frame that is left unedited, so that the video holds still on the frame before the event. The grey box covers the central 60\% of the height and width of each edited frame and leaves its border visible.

The grey box lowers $\Gamma$ the most, to less than half of its value with the annotated boundaries for both placebo-controlled models, and their intervals stay above the resolution. For these models, widening the masked window raises $\Gamma$ and narrowing it lowers it. We did not re-measure the seed resolution under these edits, so each class is read against the resolution measured with the full mask.
\subsection{More seeds}
\label{app:add-seeds}
Under the trained prompt, every new pair of seeds lies inside the resolution of Section~\ref{sec:audit} (Table~\ref{tab:add-seeds}), and Appendix~\ref{app:add-prompts} gives seed pairs under other prompts. We trained a third seed of the placebo-controlled $\alpha{=}0.5$ Qwen2.5-VL-7B model, on the stack of Appendix~\ref{app:add-setup}, and a second seed of the placebo-controlled TimeZero model, and audited each against its base and against the first seed on the 600 new questions (Appendix~\ref{app:add-setup}). Against its base, the third Qwen2.5-VL-7B seed matches the first (Table~\ref{tab:add-sample}). Audited directly against each other, the first and third seeds reach at most $0.08$ from zero under mask and $0.28$ under swap, inside the resolutions of $0.23$ and $0.31$ that the two pairs of Section~\ref{sec:audit} set. The TimeZero pair differs by $0.04$ under mask with an interval that excludes zero, an estimate far inside the resolution.

The same pair of Qwen2.5-VL-7B seeds gives finer resolutions under the three edit operators never used in training (Appendix~\ref{app:add-operators}).

\begin{table}[!htbp]
\centering\papertablestyle
\caption{\textbf{Two seeds of each placebo-controlled model agree within the audit's resolution.} All models are placebo-controlled at $\alpha{=}0.5$ and audited on the 600 new questions. A row marked \emph{against} audits the newer seed with the first seed as its base; the first seeds against their bases are in Table~\ref{tab:add-sample}. $\Gamma$ in nats per token on the $n_\cap$ questions both models localise; brackets give 95\% intervals over videos.}
\label{tab:add-seeds}
\begin{tabular}{@{}llccc@{}}
\toprule
Base model & Comparison & $n_\cap$ & $\Gm$ & $\Gs$ \\
\midrule
Qwen2.5-VL-7B & third seed against first & 297 & $-$0.02 \ci{$-$0.08}{0.03} & +0.22 \ci{0.16}{0.28} \\
\addlinespace[2pt]
TimeZero & second seed against base & 372 & +1.23 \ci{1.14}{1.32} & +1.21 \ci{1.12}{1.31} \\
 & second seed against first & 391 & +0.04 \ci{0.02}{0.06} & +0.05 \ci{0.03}{0.07} \\
\bottomrule
\end{tabular}
\end{table}

\subsection{Other prompts}
\label{app:add-prompts}
SFT and IoU-GRPO trained at 2 frames per second (Appendix~\ref{app:add-matched}) stay within $0.1$ of zero under every prompt we tried, while the placebo-controlled objective on its own varies with the prompt more than its seed resolution under the trained prompt suggests (Table~\ref{tab:add-prompts}). We audited the models on the first 200 test questions under two rewordings of the instruction that keep the Time-R1 answer format, and under VideoChat-R1's answer format, which fixes the tags and leaves only the timestamps free, with the trained instruction. Each model is paired with the base under the same prompt. A resolution measured under one prompt does not carry over to another, so each $\Gamma$ in the table is read against a seed pair audited under the same prompt. Under rewording~1 the first and third seeds of the placebo-controlled Qwen2.5-VL-7B model differ by $0.78$, with an interval reaching $0.89$, and against that resolution the third seed's $\Gamma$ does not count as an effect. IoU-GRPO with the placebo-controlled objective is steadier. Its two seeds differ by about $0.05$ or less under each new prompt, and both keep about a third of their $\Gamma$ under VideoChat-R1's format.

Under VideoChat-R1's format the placebo-controlled model keeps little likelihood $\Gamma$, while its generated answers still depend on the event (Table~\ref{tab:add-prompts-behav}). That format fixes the answer tags, on which the objective alone places most of its $\Gamma$ (Section~\ref{sec:readouts}), and the model trained with the margins on the timestamps alone keeps $+0.60$ under it (Appendix~\ref{app:add-tsonly}). Regenerated in that format, masking the event costs it 15 to 17 points more \Rfive{} than the placebo mask does, relative to the base, and costs SFT less than one point. On the first 200 questions, rewording~2 is the worst of eight instruction wordings for this model. Its third seed gives no interval on 65\% of clean questions (the first seed on 18\%, the base on 0.5\%), so only 44 questions enter its $\Gamma$. In free generation it declines on most clean and edited videos, which leaves no measurable behavioural difference. Forced to give an interval, by decoding that opens the answer and is constrained to the interval's syntax, it localises as well as SFT, and masking the event costs it 22 points more than the placebo mask does. IoU-GRPO with the placebo-controlled objective gives an interval on at least 99\% of clean questions under each of the eight wordings, and under rewording~2 it keeps both its $\Gamma$ and the behavioural difference.

Two further models, trained on mixed prompts, repair one prompt at a cost elsewhere. A placebo-controlled model trained with VideoChat-R1's answer format on about half of its 2{,}400 pairs and the Time-R1 format on the rest (run with the first seed) keeps its $\Gamma$ under VideoChat-R1's format. Under TimeLens's answer format, held out from its training, its $\Gamma$ is $+0.07$, as small as that of the model trained on one format ($+0.06$), and it gives no interval on up to 25.5\% of clean questions across the eight wordings. A model trained with the instruction in five wordings mixed per pair (run with the third seed) declines less than the third seed under rewording~2 (no interval on 26\% of clean questions) and more under the trained instruction (25.5\% against 2.5\%). On the 600 new questions the two mixed models localise less well than the models trained on one prompt (\Rfive{} $0.45$ and $0.44$, against $0.54$ and $0.55$ for the first and third seeds and $0.51$ for the base). Their $\Gm$ there is $+1.20$ $[1.07,1.32]$ and $+0.97$ $[0.84,1.10]$, and audited directly against the third seed the mixed-wording model gives $-0.52$ $[-0.57,-0.47]$.

\begin{table}[!htbp]
\centering\papertablestyle
\caption{\textbf{SFT and IoU-GRPO stay near zero under every prompt, and the placebo-controlled objective on its own is less stable than when added to IoU-GRPO.} $\Gm$ in nats per token on the first 200 test questions, each model against the base under the same prompt; the rewordings change the instruction and keep the Time-R1 answer format. A row marked \emph{against} audits the newer seed with the first as its base, giving the seed resolution under that prompt, with 95\% intervals over videos. The last two models were trained with the Time-R1 and VideoChat-R1 answer formats mixed, or with five instruction wordings mixed, and were run with the first and the third seed. Under the trained instruction the placebo-controlled seeds are compared in Table~\ref{tab:add-seeds}, and the two seeds of IoU-GRPO with the placebo-controlled objective were not audited against each other under it. $n_\cap$ lies between 87 and 140 except where marked. IoU-GRPO is trained at 2 frames per second.}
\label{tab:add-prompts}
\setlength{\tabcolsep}{2pt}
\begin{tabular}{@{}lcccc@{}}
\toprule
Model & \shortstack{trained\\instruction} & \shortstack{rewording\\1} & \shortstack{rewording\\2} & \shortstack{VideoChat-R1\\format} \\
\midrule
SFT & $-$0.09 & $-$0.04 & $-$0.08 & $-$0.04 \\
IoU-GRPO & +0.01 & +0.05 & $-$0.09 & +0.08 \\
\addlinespace[2pt]
\addlinespace[1pt]\multicolumn{5}{@{}l}{\textit{Placebo-controlled objective}} \\
\rowcolor{pcrow}\quad first seed & +1.62 & +1.24 & +1.15 & +0.10 \\
\rowcolor{pcrow}\quad third seed & +1.62 & +0.45 & +0.67$^a$ & +0.20 \\
\quad third against first & --- & $-$0.78 \ci{$-$0.89}{$-$0.65} & $-$0.38$^b$ \ci{$-$0.49}{$-$0.29} & +0.10 \ci{0.08}{0.12} \\
\addlinespace[1pt]\multicolumn{5}{@{}l}{\textit{IoU-GRPO with the placebo-controlled objective}} \\
\rowcolor{pcrow}\quad first seed & +0.85 & +0.82 & +0.81 & +0.31 \\
\rowcolor{pcrow}\quad second seed & +0.96 & +0.92 & +0.91 & +0.31 \\
\quad second against first & --- & +0.05 \ci{$-$0.03}{0.14} & +0.04 \ci{$-$0.04}{0.12} & $-$0.01 \ci{$-$0.04}{0.01} \\
\addlinespace[1pt]\multicolumn{5}{@{}l}{\textit{Placebo-controlled objective, trained on mixed prompts}} \\
\rowcolor{pcrow}\quad formats mixed & --- & +0.82 & +1.01 & +1.11 \\
\rowcolor{pcrow}\quad wordings mixed & --- & +1.00 & +1.15 & --- \\
\bottomrule
\end{tabular}
\par\smallskip
{\scriptsize\raggedright $^a$$n_\cap{=}44$, since this model gives no interval on 65\% of clean questions under rewording~2. $^b$$n_\cap{=}48$.\par}
\end{table}

\begin{table}[!htbp]
\centering\papertablestyle
\caption{\textbf{Under another answer format and another wording, masking the event still costs the placebo-controlled models more correct answers than the placebo mask does.} Answers regenerated on the first 500 test questions, clean and with the evidence or as many other frames masked. Each entry is the drop in \Rfive{} from the placebo mask to the evidence mask, minus the base's drop under the same prompt, in points with 95\% intervals over videos. In the forced columns decoding opens the answer and is constrained to one interval; clean \Rfive{} is that of the forced answers, $0.50$ for the base.}
\label{tab:add-prompts-behav}
\begin{tabular}{@{}lcccc@{}}
\toprule
 & VideoChat-R1 format & rewording 2 & \multicolumn{2}{c}{rewording 2, interval forced} \\
\cmidrule(lr){2-2}\cmidrule(lr){3-3}\cmidrule(l){4-5}
Model & mask $-$ placebo & mask $-$ placebo & mask $-$ placebo & clean \Rfive{} \\
\midrule
SFT & +0.6 \ci{$-$4.4}{5.6} & --- & +1.8 \ci{$-$2.6}{6.6} & 0.60 \\
\rowcolor{pcrow}placebo-controlled, first seed & +15.2 \ci{10.2}{20.2} & +13.4 \ci{7.9}{19.1} & +12.2 \ci{7.7}{16.6} & 0.58 \\
\rowcolor{pcrow}placebo-controlled, third seed & +16.8 \ci{11.6}{22.1} & $-$2.0 \ci{$-$8.6}{4.7} & +22.4 \ci{17.0}{27.9} & 0.59 \\
\rowcolor{pcrow}\shortstack[l]{IoU-GRPO with the\\placebo-controlled objective} & --- & +20.2 \ci{13.6}{27.0} & --- & --- \\
\bottomrule
\end{tabular}
\end{table}

\subsection{Constants of the placebo bound}
\label{app:add-bound}
The bound's constants were set in the 16-frame setup (Appendix~\ref{app:prereg}), and the bound keeps its effect at the audit's 2 frames per second and $3{,}584\times28\times28$ pixels. In that setting, removing the bound leaves the trained model with a placebo drop of $0.49$ nats per token and the full objective brings it to $0.01$ (Tables~\ref{tab:objectives} and~\ref{tab:audit}), and each of the five adapters of Table~\ref{tab:families} keeps its placebo drop within $0.10$ of its base's. Two further settings of the bound's ratio and slack, one tighter and one looser than the $(1.25, 0.10)$ of Equation~\ref{eq:objective}, change $\Gm$ on the 600 new questions by at most $0.1$ and the accuracy gain by less than $0.2$ points (Table~\ref{tab:add-variants}). The looser setting lets the trained model's placebo drop rise to $0.07$ nats per token, with the base at $-0.08$.

\begin{table}[!htbp]
\centering\papertablestyle
\caption{\textbf{Other bound constants and a margin on the timestamps alone leave $\Gamma$ and the accuracy gain where the setting of Equation~\ref{eq:objective} puts them.} Placebo-controlled Qwen2.5-VL-7B models at $\alpha{=}0.5$, each run with the first seed and trained on the same 2{,}400 pairs in the same order. $\Gm$ and the trained model's placebo drop $\dO$ are in nats per token on the 600 new questions ($n_\cap$ from 239 to 271; the base's $\dO$ is $-0.08$ in every pair). $\Delta$\Rfive{} is the change from the base on the full 3{,}720-question split in points; the first row is paired by question with its 95\% interval, and the others are differences of the two accuracies on the same questions.}
\label{tab:add-variants}
\begin{tabular}{@{}lccc@{}}
\toprule
Bound (ratio, slack) and margin & $\Gm$ & $\Delta$\Rfive{} & $\dO$ \\
\midrule
\rowcolor{pcrow}$(1.25, 0.10)$, whole answer & +1.50 \ci{1.37}{1.64} & +3.9 \ci{2.6}{5.2} & 0.00 \\
\rowcolor{pcrow}$(1.0, 0.05)$, whole answer & +1.57 \ci{1.42}{1.70} & +3.8 & 0.00 \\
\rowcolor{pcrow}$(1.5, 0.20)$, whole answer & +1.60 \ci{1.47}{1.72} & +4.0 & 0.07 \\
\rowcolor{pcrow}$(1.25, 0.10)$, timestamp tokens only & +1.52 \ci{1.43}{1.61} & +3.8 & $-$0.01 \\
\bottomrule
\end{tabular}
\end{table}

\subsection{A margin on the timestamps alone}
\label{app:add-tsonly}
Applying the two margins to the timestamp tokens alone moves the dependence onto the timestamps and leaves $\Gamma$ over the whole answer and the accuracy gain unchanged (Table~\ref{tab:add-variants}). On the first 200 test questions this model's $\Gm$ is $+3.17$ $[2.86,3.45]$ on the timestamp tokens and $+1.52$ $[1.35,1.70]$ over the whole answer ($n_\cap{=}94$), and the other tokens of the answer carry none of it ($-0.12$ $[-0.25,0.00]$). VideoChat-R1's template produces the same split through its answer format. It fixes the tags and leaves only the timestamps free, and the placebo-controlled VideoChat-R1 model places all of its $\Gamma$ on them ($+1.40$; Table~\ref{tab:split}). With tags, the placebo-controlled TimeZero model has $\Gamma$ $+0.34$ on the timestamps.

The timestamp-only model also carries more of its $\Gamma$ to other answer formats. Under VideoChat-R1's and TimeLens's formats it keeps $+0.60$ and $+0.44$ on the first 200 questions, where the whole-answer model keeps little (Appendix~\ref{app:add-prompts}). Across the eight instruction wordings it gives no interval on up to 24\% of clean questions.

\subsection{Refusal-aware fine-tuning: RA-RFT}
\label{app:add-rarft}

RA-RFT \citep{lee2026rarft} is trained to decline queries that do not match the video, and it leaves $\Gamma$ where its base has it (Table~\ref{tab:add-rarft}). It is a generative video model trained from Time-R1 \citep{wang2025timer1}, so we audit the two with the protocol of the main text, both under Time-R1's prompt template. On the 600 new questions (Appendix~\ref{app:add-setup}) the intervals of both $\Gm$ and $\Gs$ contain zero and lie within $0.03$ nats per token of it. When decoding is forced to give one interval (the output starts with the answer tag and must continue with an interval), Time-R1 and RA-RFT each lose 41 points more \Rfive{} to the evidence mask than to the placebo mask, and the difference between them is $0.0$ points.

Under its released prompt, RA-RFT gives no interval for 84\% of mismatched queries, against 72\% for Time-R1 under the same prompt. This rate counts any output without an interval, unlike the verbatim abstention of Table~\ref{tab:s4}, and RA-RFT also gives no interval for more of the matched queries than Time-R1 does, so its rise on mismatched queries is largely matched by a rise on the question's own query. Our mismatched queries are annotation sentences of other videos, which Time-R1 already leaves without an interval most of the time. IoU-GRPO with the placebo-controlled objective shows the reverse, with $\Gamma$ near $+0.9$ and fewer mismatched queries left without an interval than its base (Table~\ref{tab:add-matched-acc}). Declining mismatched queries and $\Gamma$ can therefore come apart.

\begin{table}[!htbp]
\centering\scriptsize\setlength{\tabcolsep}{4pt}\renewcommand{\arraystretch}{1.05}
\caption{\textbf{RA-RFT leaves $\Gamma$ unchanged, and like its base it gives no interval for most mismatched queries.} Time-R1 is RA-RFT's base. $\Gm$ and $\Gs$ are in nats per token on the 600 new questions ($n_\cap{=}340$), under Time-R1's template. \emph{Forced} is the change from the base, in points, of the \Rfive{} lost to the evidence mask beyond the placebo mask, with decoding forced to give one interval. The last two columns give the share of outputs without an interval for the question's own query and for the query of another video, under RA-RFT's released prompt; they count any output without an interval. Both behavioural readouts use the first 1{,}000 test questions. Brackets give 95\% intervals.}
\label{tab:add-rarft}
\begin{tabular}{@{}lccccc@{}}
\toprule
Model & $\Gm$ & $\Gs$ & forced & no interval, own query & no interval, mismatched \\
\midrule
Time-R1 & \NA & \NA & \NA & 0.113 & 0.719 \\
RA-RFT & $-$0.003 \ci{$-$0.022}{0.016} & $-$0.014 \ci{$-$0.030}{0.002} & 0.0 \ci{$-$1.7}{1.7} & 0.198 & 0.841 \\
\bottomrule
\end{tabular}
\end{table}

\subsection{Negative-aware moment retrieval: NA-VMR}
\label{app:add-navmr}

On the questions both models answer correctly, the set on which $\Gamma$ is defined (Section~\ref{sec:audit}), the released negative-aware model of NA-VMR \citep{flanagan2025navmr} depends more on the event's clips than UniVTG, the model it extends, both when they are swapped for clips of another video and when they are masked. Over all audited questions it does so under swap only (Table~\ref{tab:add-navmr}). NA-VMR trains UniVTG \citep{lin2023univtg} also on irrelevant queries and adds a rejection head, which declines a query when its output falls below a threshold ($0.5$ in the release). Both models read precomputed SlowFast and CLIP features of one-second clips, so we port the audit to these features. The event clips are those that overlap the annotated interval, and the placebo edits as many clips outside it, with three draws per question. Swap replaces the features of the edited clips with those of clips from one other video, the same donor clips for the evidence and the placebo edit, and mask sets them to zero. The clip-position features are left unchanged. Here $\psi$ is the drop of the mean log foreground probability over the event clips under the evidence edit minus its mean drop under the placebo edits, and $\Gamma$ is the change in $\psi$ from UniVTG to UniVTG-NA, in nats. We read the released Charades-STA checkpoints of both models, for which our feature pipeline reproduces the reported \Rfive{} within $0.4$ points. Of the 3{,}720 test questions we leave out the 95 whose event covers more than half of the clips, which leaves no room for an equally large placebo.

Under swap, $\Gamma$ is positive on the 1{,}883 questions both checkpoints answer correctly and on all 3{,}625 audited questions, and the rejection head fires more often when the event clips are replaced than when as many other clips are. Under the mask, $\Gamma$ is smaller on the questions both answer correctly and zero over all audited questions, while the rejection head and the predicted intervals still respond more to the evidence edit. The two released checkpoints also differ in learning rate and starting point, so we attribute these differences to the released model and make no claim about negative training alone.

\begin{table}[!htbp]
\centering\scriptsize\setlength{\tabcolsep}{4pt}\renewcommand{\arraystretch}{1.05}
\caption{\textbf{On the questions both models answer correctly, the released UniVTG-NA depends more on the event clips than UniVTG under swap and under the mask, and over all questions under swap only.} The audit edits clip features on Charades-STA. $\Gamma$ is in nats, from the mean log foreground probability over the event clips. The rejection head exists only in UniVTG-NA, and its rows give how much more often, in points, it fires under the evidence edit than under the placebo edit. The last block gives the change from UniVTG to UniVTG-NA, in points, of the \Rfive{} lost to the evidence edit beyond the placebo edit, with the rejection head ignored. \emph{Both correct} is the 1{,}883 questions both checkpoints answer correctly on the clean video, and \emph{all} the 3{,}625 questions whose event covers at most half of the clips. Brackets give 95\% intervals.}
\label{tab:add-navmr}
\begin{tabular}{@{}lcc@{}}
\toprule
 & swap & mask \\
\midrule
\addlinespace[1pt]\multicolumn{3}{@{}l}{\textit{$\Gamma$, nats}} \\
both correct & +0.55 \ci{0.47}{0.63} & +0.28 \ci{0.14}{0.43} \\
all & +0.35 \ci{0.30}{0.40} & $-$0.01 \ci{$-$0.10}{0.09} \\
\addlinespace[1pt]\multicolumn{3}{@{}l}{\textit{rejection head, evidence edit minus placebo edit, points}} \\
both correct & +27.8 \ci{25.4}{30.2} & +42.5 \ci{39.6}{45.3} \\
all & +15.4 \ci{13.9}{17.0} & +25.4 \ci{23.4}{27.4} \\
\addlinespace[1pt]\multicolumn{3}{@{}l}{\textit{\Rfive{} lost to the evidence edit beyond the placebo edit, change, points}} \\
both correct & +10.2 \ci{7.7}{12.7} & +13.6 \ci{11.1}{16.3} \\
all & +11.0 \ci{9.1}{12.8} & +14.2 \ci{12.1}{16.2} \\
\bottomrule
\end{tabular}
\end{table}

\subsection{Likelihood relative to fixed competing intervals}
\label{app:add-competing}

Scored against a fixed competing interval, the placebo-controlled models separate little from SFT when the competitor is the base model's own prediction, and clearly when it is the annotated interval shifted by its own length, a comparison in which IoU-GRPO also moves (Table~\ref{tab:add-competing}). For each question the competing answer texts are fixed and identical for a model and its base. With $m(x)$ the log-likelihood of the annotated answer minus that of the competitor on video $x$, the readout is $m(x_O)-m(x_E)$, how much more the annotated interval's advantage shrinks when the event frames are masked than when as many other frames are, with $m(x_O)$ averaged over three placebo draws. The table gives the mean change of this readout from the base on the questions among the first 200 test questions that the model and its base both answer correctly. On these questions the base's own prediction is itself correct, so that competitor lies close to the annotated interval. The values are summed over the answer tokens, in nats, and are not on the per-token scale of $\Gamma$.

Against the base's prediction, the 4{,}800-pair model lies above SFT trained on the same 4{,}800 pairs, though their intervals overlap slightly, and the interval of the $\alpha{=}0.5$ model overlaps SFT's. Against the shifted interval the placebo-controlled models lie well above SFT, but IoU-GRPO alone also reaches $+1.87$, so this comparison partly reflects accuracy training. Adding the placebo-controlled objective to IoU-GRPO raises it well beyond that at the same accuracy.

\begin{table}[!htbp]
\centering\scriptsize\setlength{\tabcolsep}{4pt}\renewcommand{\arraystretch}{1.05}
\caption{\textbf{Against the base's own prediction the placebo-controlled models separate little from SFT; against a shifted interval they separate clearly, and IoU-GRPO moves as well.} Each value is the mean trained-minus-base change in how much more the annotated interval's log-likelihood advantage over the competitor shrinks under the evidence mask than under the placebo mask, in nats summed over the answer tokens. $n$ is the number of the first 200 test questions that the model and its base both answer correctly. IoU-GRPO is trained at 2 frames per second. The two 4{,}800-pair models are paired with the base of the main results' stack (\Rfive{} $0.480$), the others with the base at $0.507$ (Appendix~\ref{app:add-setup}). Brackets give 95\% intervals.}
\label{tab:add-competing}
\begin{tabular}{@{}lccc@{}}
\toprule
Model & $n$ & base model's own prediction & annotated interval shifted by its length \\
\midrule
IoU-GRPO & 101 & +0.09 \ci{$-$0.96}{1.04} & +1.87 \ci{0.21}{3.27} \\
\rowcolor{pcrow} IoU-GRPO with the placebo-controlled objective & 96 & +1.42 \ci{0.11}{2.64} & +10.70 \ci{8.12}{12.92} \\
SFT, 4{,}800 pairs & 100 & +0.75 \ci{0.04}{1.37} & $-$0.47 \ci{$-$1.15}{0.19} \\
\rowcolor{pcrow} placebo-controlled $\alpha{=}0.5$ & 100 & +1.25 \ci{0.16}{2.26} & +4.33 \ci{3.50}{5.14} \\
\rowcolor{pcrow} placebo-controlled $\alpha{=}0.5$, 4{,}800 pairs & 94 & +2.42 \ci{1.17}{3.69} & +4.07 \ci{3.10}{4.99} \\
\bottomrule
\end{tabular}
\end{table}

\subsection{What a higher \texorpdfstring{$\Gamma$}{Gamma} adds at matched accuracy}
\label{app:add-matched-acc}

At the same gain in accuracy, the model with the higher $\Gamma$ keeps fewer of its correct answers once the event frames are masked (Table~\ref{tab:add-matched-acc}). We compare two pairs of models whose gains in \Rfive{} lie within about one point of each other. In the first, IoU-GRPO trained at 2 frames per second on the 2{,}400 pairs of our objective is set against the same training with the placebo-controlled objective added to its loss, for two seeds (Appendix~\ref{app:add-matched}). In the second, the SFT run on 4{,}800 pairs, taken at its checkpoint after 800 pairs, is set against the placebo-controlled model trained on the same 4{,}800 pairs. We take the questions that a model and its base both answer correctly and ask how many of them each still answers correctly with the event frames masked. In both pairs the higher-$\Gamma$ model keeps fewer of them than its base does. The SFT checkpoint keeps more of them than its base, and IoU-GRPO about as many. The evidence share over all 1{,}000 questions, which the placebo mask controls, gives the same order: it rises over the base for the higher-$\Gamma$ models, falls for the SFT checkpoint, and for IoU-GRPO changes by an amount whose interval includes zero. Giving no interval for a mismatched query rises with the objective against the SFT checkpoint and barely against IoU-GRPO, so we do not attribute it to $\Gamma$.

\begin{table}[!htbp]
\centering\scriptsize\setlength{\tabcolsep}{3pt}\renewcommand{\arraystretch}{1.05}
\caption{\textbf{At equal accuracy gain, the model with the higher $\Gamma$ keeps fewer correct answers once the event frames are masked.} Each model is compared with its base re-run on the same software stack. The second block runs on a stack where the base reaches \Rfive{} $0.480$, as in the main results, and where the 4{,}800-pair model's $\Gm$ is $+2.78$ in Table~\ref{tab:s1}. $\Delta$\Rfive{} is in points on the full test split, and $\Gm$ in nats per token. \emph{Correct under the mask} is, among the questions that both the model and its base answer correctly, the share still answered correctly with the event frames masked, for the base and for the model. $\Delta$\emph{share} is the change from the base, in points, of the evidence share $1-a_E/a_O$ under the mask over all 1{,}000 questions, which the placebo mask controls. \emph{Mismatched} is the share of outputs without an interval when the query is the annotation sentence of another video; the two bases give $0.288$ and $0.275$. The three behavioural columns use the first 1{,}000 test questions, and the second seed has no mismatched-query run. Brackets give 95\% intervals.}
\label{tab:add-matched-acc}
\begin{tabular}{@{}>{\raggedright\arraybackslash}p{4.1cm}ccccc@{}}
\toprule
Model & $\Delta$\Rfive{} & $\Gm$ & \shortstack{correct under the mask\\base $\to$ model} & $\Delta$share & mismatched \\
\midrule
\addlinespace[1pt]\multicolumn{6}{@{}l}{\textit{2{,}400 pairs at 2 frames per second; $\Gamma$ on the 600 new questions}} \\
IoU-GRPO & +15.9 & +0.04 \ci{0.02}{0.07} & 0.29 $\to$ 0.32 & +3.4 \ci{$-$1.8}{8.7} & 0.046 \\
\rowcolor{pcrow} IoU-GRPO with the placebo-controlled objective, first seed & +15.3 & +0.88 \ci{0.81}{0.94} & 0.29 $\to$ 0.23 & +12.8 \ci{7.9}{17.8} & 0.057 \\
\rowcolor{pcrow} IoU-GRPO with the placebo-controlled objective, second seed & +15.9 & +0.92 \ci{0.85}{0.99} & 0.29 $\to$ 0.18 & +15.8 \ci{10.9}{20.9} & \NA \\
\addlinespace[1pt]\multicolumn{6}{@{}l}{\textit{4{,}800 pairs; $\Gamma$ on the first 200 test questions}} \\
SFT, checkpoint after 800 pairs & +7.1 & $-$0.13 \ci{$-$0.20}{$-$0.06} & 0.27 $\to$ 0.39 & $-$7.5 \ci{$-$11.9}{$-$2.9} & 0.092 \\
\rowcolor{pcrow} placebo-controlled $\alpha{=}0.5$, 4{,}800 pairs & +8.1 & +2.77 \ci{2.54}{3.03} & 0.27 $\to$ 0.20 & +10.5 \ci{6.0}{15.2} & 0.373 \\
\bottomrule
\end{tabular}
\end{table}

The objective also adds dependence to a base that already uses the event. On TimeZero, the placebo-controlled objective at $\alpha{=}0.5$ leaves accuracy and coverage where they were (Table~\ref{tab:s1}) and raises $\Gamma$ (Table~\ref{tab:families}), the share of correct answers lost to the evidence mask (Table~\ref{tab:s6}) and the share of all-black videos left unanswered (Table~\ref{tab:s3}). IoU-GRPO with the placebo-controlled objective, trained from TimeZero, gains about a point of \Rfive{} over TimeZero and raises $\Gamma$ to about $+1$ (Table~\ref{tab:add-matched}). In its answers regenerated on the first 1{,}000 test questions, masking the event lowers \Rfive{} by $4$ $[1,7]$ points more than the placebo mask does, relative to TimeZero.

\end{document}